\PassOptionsToPackage{table,dvipsnames}{xcolor}

\documentclass{maskwam}

\usepackage{amsmath}
\usepackage{amssymb}
\usepackage{amsfonts}
\usepackage{mathtools}
\usepackage{enumitem}
\usepackage{algorithm}
\usepackage{algorithmic}
\usepackage{wrapfig}
\usepackage{adjustbox}
\usepackage{tabularx}
\usepackage{makecell}
\usepackage{siunitx}
\usepackage{float}
\usepackage{diagbox}
\usepackage{dsfont}
\usepackage{needspace}

\usepackage{amsmath,amsfonts,bm}

\def\eqref#1{equation~\ref{#1}}

\def\1{\bm{1}}

\DeclareMathAlphabet{\mathsfit}{\encodingdefault}{\sfdefault}{m}{sl}
\SetMathAlphabet{\mathsfit}{bold}{\encodingdefault}{\sfdefault}{bx}{n}

\usepackage{amssymb}
\usepackage{mathtools}
\usepackage{graphicx}
\usepackage{booktabs}
\usepackage{longtable}
\usepackage{nicefrac}
\usepackage{microtype}
\usepackage{enumitem}
\usepackage{algorithm}
\usepackage{algorithmic}
\usepackage{wrapfig}
\usepackage{adjustbox}
\usepackage{tabularx}
\usepackage{makecell}
\usepackage{siunitx}
\usepackage{float}
\usepackage{diagbox}
\usepackage{dsfont}
\usepackage{multirow}
\usepackage{subcaption}
\usepackage{colortbl}
\usepackage{pifont}
\usepackage{marvosym}
\usepackage{tcolorbox}
\tcbuselibrary{listings, skins, breakable}

\definecolor{promptbg}{HTML}{F2F6FA}
\definecolor{promptframe}{HTML}{C5D4E3}

\newtcblisting{promptbox}{
    listing only,
    breakable,
    colback=promptbg,
    colframe=promptframe,
    coltext=black!85,
    boxrule=0.8pt,
    arc=4pt,
    left=6pt,
    right=6pt,
    top=6pt,
    bottom=6pt,
    listing options={
        basicstyle=\ttfamily\scriptsize,
        breaklines=true,
        breakatwhitespace=true,
    }
}

\definecolor{bestgreen}{RGB}{224,232,240}
\definecolor{secondgreen}{RGB}{240,245,249}
\definecolor{baselinecolor}{gray}{.9}
\definecolor{drp-blue}{HTML}{1f77b4}
\definecolor{pretty-blue}{RGB}{0, 113, 188}
\definecolor{kaiming-green}{RGB}{57,181,74}
\definecolor{mypurple}{RGB}{55,0,168}
\definecolor{icmlblue}{rgb}{0,0.08,0.45}
\definecolor{mygreen}{HTML}{4FC978}
\definecolor{linecolor1}{RGB}{246, 248, 239}
\definecolor{linecolor2}{RGB}{230, 234, 217}
\definecolor{linecolor3}{RGB}{211, 222, 190}
\definecolor{tableprofile}{HTML}{FAF1D6}
\definecolor{tablesetting}{HTML}{EAF2F8}
\definecolor{tableattention}{HTML}{EAF5EA}
\definecolor{tableintention}{HTML}{E8F2F8}
\definecolor{tablereaction}{HTML}{FAF0DA}
\definecolor{tableinteraction}{HTML}{F0E9F7}
\definecolor{reconcolor}{HTML}{412F8A}
\definecolor{runpei-orange}{HTML}{F35F27}
\definecolor{runpei_blue}{HTML}{14294B}
\definecolor{datacolor}{HTML}{0009BF}
\definecolor{vitcolor}{HTML}{fc8e62}
\definecolor{cvprblue}{rgb}{0.21,0.49,0.74}
\definecolor{myblue}{rgb}{.39,.58,.93}

\newcommand{\cmark}{\ding{51}}
\newcommand{\pcmark}{\textcolor{gray}{\ding{51}}}
\newcommand{\xmark}{\ding{55}}
\newcolumntype{x}[1]{>{\centering\arraybackslash}p{#1pt}}

\newcounter{question}
\newcommand{\airislogo}{%
  \raisebox{-0.65em}{%
    \includegraphics[height=2.75em]{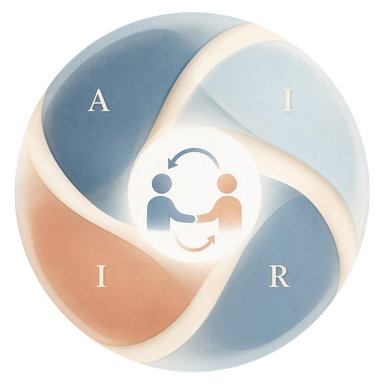}}%
}


\title{\airislogo\hspace{0.50em}\parbox[b]{0.80\linewidth}{\raggedright
AIRIS: Benchmarking Social Embodied Intelligence in Egocentric Streaming Videos}}

\author{Yuchen Zhou}
\author{Hongsheng Wu}
\author{Jiayu Tang}
\author{Yunxuan Mao}
\author{Hanjing Ye}
\author{Shenghao Zhang}
\author{Jing Tang}
\author{Chun-Kai Fan}
\author[*]{Yufei Huang}
\author[*]{Yonggen Ling}
\author{He Zhang}

\affiliation{Tencent Robotics X Lab}
\contribution[*]{Corresponding authors}

\abstract{
Embodied agents operating around people must not only recognize social cues, but also decide when the available evidence is sufficient, whether to intervene, and how to respond without violating social norms. Existing benchmarks typically evaluate these abilities in isolation or with access to completed videos, obscuring where an online social decision fails. We introduce AIRIS, a causal streaming benchmark that formulates social embodied intelligence as a time-anchored chain of Attention, Intention, Reaction, and Interaction (A-I-R-I). AIRIS contains 8,000 expert-curated questions grounded in 357 egocentric clips across 11 context categories and 36 capability dimensions. At each query, models receive only the observation prefix available at that time; decision-aware metrics reward calibrated waiting and complete reactions while penalizing premature commitment. Evaluating 45 proprietary and open-source general-purpose, embodied, streaming-video, and omni-modal models reveals fragmented and temporally miscalibrated competence. General-purpose VLMs excel at passive cue grounding and intent recognition, whereas embodied specialists lead on isolated reaction sub-decisions; yet across all families, models frequently guess intents before sufficient evidence emerges and fail to jointly align intervention timing, action domain, and concrete execution within the same episode. AIRIS thus provides a diagnostic foundation for bridging the gap between passive social perception and reliable embodied interaction.
}
\metadata[Project page]{\url{https://airis-bench.github.io}}
\metadata[GitHub]{\url{https://github.com/AIRIS-bench/AIRIS}}

\begin{document}
\maketitle
\vspace{-3mm}

\section{Introduction}

An embodied agent sharing space with people must do more than recognize objects and execute commands.
When a person glances toward an object and begins a gesture, the agent must determine whether the cue concerns it, infer what the person intends, and decide whether to act immediately, wait for more evidence, or refrain from acting.
Even a physically feasible response can be socially mistimed, intrusive, or incomplete.
Such decisions arise continually in collaborative homes, public spaces, and assistive settings, where successful behavior depends on both task competence and sensitivity to people \citep{puig2024habitat3,ramnauth2024shouldihelp,zhou2023pit}.
We conceptualize the ability to perceive, interpret, and act appropriately within these evolving interpersonal situations as \emph{social embodied intelligence} \citep{fong2003survey,dautenhahn2007socially,lee2024towards}.

\begin{figure}[t]
  \centering
  \includegraphics[width=\columnwidth,trim={17.5bp 1bp 17.25bp 3bp},clip]{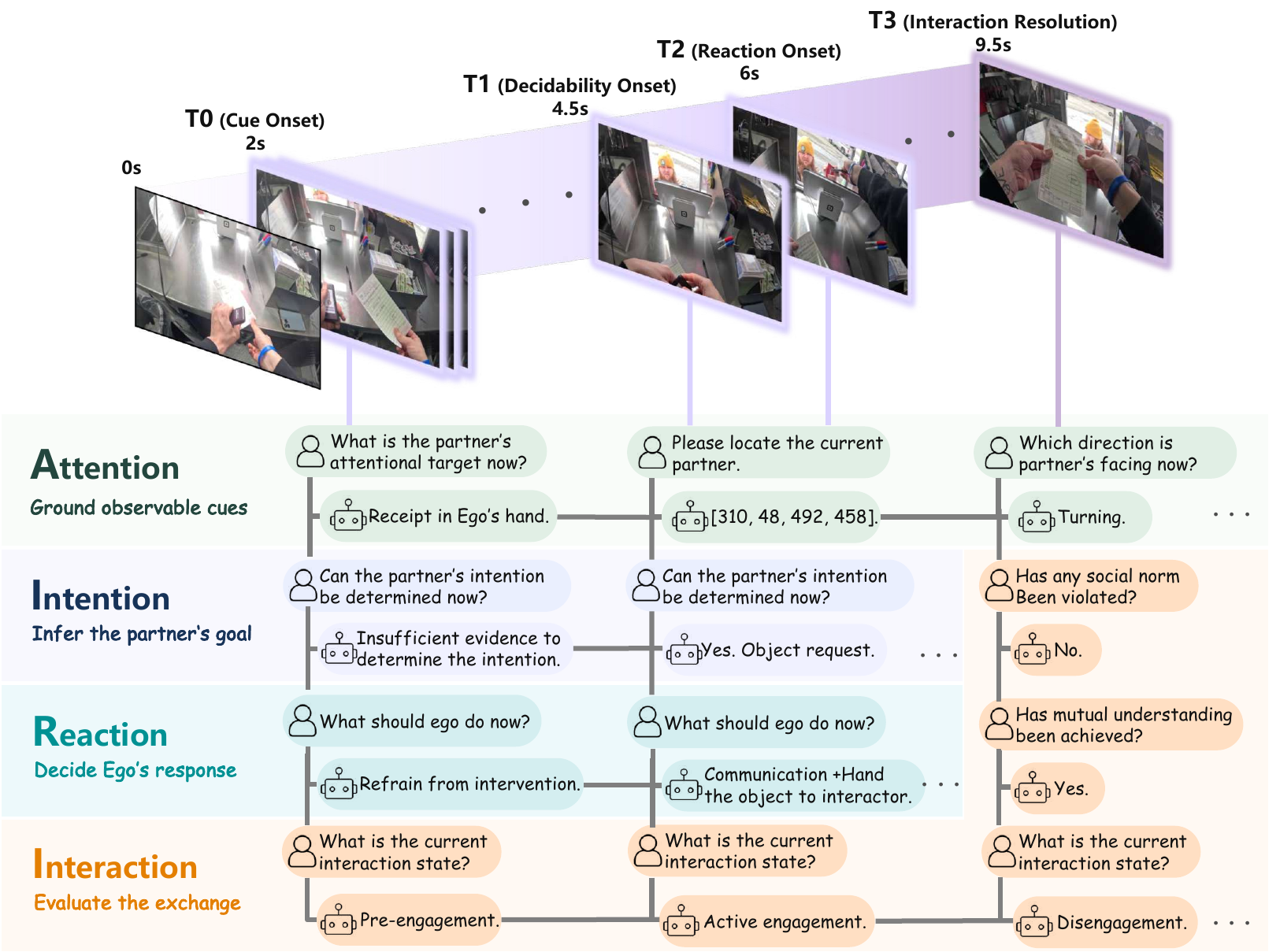}
  \caption{\textbf{AIRIS: evaluating social embodied intelligence over time.} An egocentric episode illustrates four linked capabilities: grounding social cues (Attention), inferring the partner's goal (Intention), deciding Ego's response (Reaction), and evaluating the exchange (Interaction). Each query uses only observations up to the query time, without access to future evidence.}
  \label{fig:teaser}
\end{figure}

To date, the community has successfully operationalized distinct facets of this intelligence through specialized, domain-specific tasks, as shown in Table \ref{tab1}. 
For instance, research in streaming and proactive video evaluates models under continuous input, emphasizing strict observation prefixes and timely intervention \citep{lin2026streamingbench,niu2025ovobench,lei2026egosat,zhang2025proassist}. 
Meanwhile, multimodal social benchmarks dive deep into interpreting gaze, gesture, intent, and latent mental states in natural human interactions \citep{zadeh2019socialiq,kong2026sivbench,lee2024modeling,shi2025mumatom}. 
In parallel, norm-aware embodied evaluations assess whether proposed actions are permissible and compatible with hidden constraints during ongoing human activities \citep{rezaei2025egonormia,fang2026roboteq,normact2026,assistance2026nonintrusive}.
However, existing evaluations often assess social cues, intent, response appropriateness, and interaction outcomes separately. This separation makes it difficult to characterize these capabilities within the same unfolding interaction. A unified evaluation with shared episodes and temporal anchors would enable systematic comparison of stage-specific strengths and weaknesses.

To address this mismatch, we introduce \textbf{AIRIS}, a streaming benchmark that evaluates social embodied intelligence as a multi-stage decision process spanning \textbf{A}ttention, \textbf{I}ntention, \textbf{R}eaction, and \textbf{I}nteraction, as shown in Figure \ref{fig:teaser}.
Specifically, \emph{Attention} evaluates whether the model can localize the interaction partner from the first-person stream \citep{ye2025tpt,ye2026refertrack} and integrate explicit multimodal cues, including gaze, gestures, body posture, emotional states, and speech, into an observable focus.
Building on this evidence, \emph{Intention} infers the partner's underlying goal and identifies the earliest moment this goal becomes decidable.
\emph{Reaction} then shifts to the ego-agent, assessing its ability to select well-timed navigation, manipulation, communication, or justified inaction in response to the inferred intent to advance the exchange.
Finally, \emph{Interaction} tracks the evolving episode to evaluate the overall outcome against task completion, mutual understanding, and social norms.
By aligning these capabilities within the same episodes, AIRIS assesses social cues, intent inference, response selection, and interaction outcomes under a shared temporal framework. This enables stage-specific diagnosis of model strengths and weaknesses.

To operationalize this evaluation, AIRIS aligns every social episode 
along four precise temporal anchors: $T_0$ (social cue onset), $T_1$ (when intent becomes inferable), $T_2$ (ego-agent response onset), and $T_3$ (interaction resolution). 
By querying models at specific moments $\tau$ using only the causal observation 
prefix up to that point, we prevent future evidence from leaking into online 
decisions.
To capture the complexity and diversity of real-world social environments, AIRIS encompasses 8,000 questions across 357 egocentric video clips, spanning 11 diverse interaction contexts ranging from retail and market commerce to everyday social encounters and collaborative professional services.
These questions probe 36 fine-grained capability dimensions systematically structured across the four A-I-R-I interaction stages.
We benchmark 45 representative proprietary and open-source general-purpose, embodied, streaming video, and omni-modal Vision-Language Models (VLMs), including Gemini, GPT, Doubao, the Qwen and InternVL series, as well as specialized models such as HY-Embodied, RoboBrain, RynnBrain, and JoyAI.

Our evaluation reveals that current social embodied competence is both fragmented across model families and miscalibrated over time. Leading general-purpose VLMs excel at perceptual cue grounding (up to 93.6\% on Attention) and intent recognition, whereas embodied specialists lead on isolated response sub-decisions (up to 68.8\% on Reaction). Yet no model closes the full online decision loop. Prior to intent decidability ($\tau < T_1$), model families prematurely guess a specific intent on 26\%-51\% of queries instead of waiting for sufficient evidence. Furthermore, requiring simultaneous correctness across intervention timing, action domain, and concrete execution causes full-reaction accuracy to collapse below 14.7\% across all 45 models. Current VLMs thus remain capable observers and post-hoc judges, rather than temporally calibrated social agents.
Our contributions are threefold:

\begin{itemize}[leftmargin=1.5em,itemsep=2pt,topsep=2pt]
  \item \textbf{Formalizing the Social Interaction Loop.}
        We break down multi-party social interaction into an interdependent four-stage decision process (A-I-R-I), introducing a time-anchored framework that bridges passive multimodal perception with proactive embodied action and social norm compliance.
  \item \textbf{Ego-Centric Causal Streaming Benchmark.}
        We present AIRIS, a benchmark of 8,000 QA pairs across 11 diverse social contexts and 36 capability dimensions, evaluated under a strict causal streaming protocol that mirrors ego-centric sensory constraints and eliminates future-information leakage.
  \item \textbf{Exposing the Perception-Decision Gap.}
  Across 45 models spanning general-purpose, embodied, streaming-video, and omni-modal families, AIRIS reveals that relatively strong social-cue perception and post-hoc assessment do not translate into evidence-calibrated anticipation or complete embodied reactions. Temporal calibration and end-to-end decision composition emerge as two major unresolved bottlenecks to reliable social embodied intelligence.
\end{itemize}

\begin{table*}[t]
  \centering
  \caption{Comparison with representative online-video, social-reasoning, and embodied-decision benchmarks. QA counts cover evaluation sets only. The grouped columns instantiate A-I-R-I as interactant/cue grounding, latent-goal/decidability, ego action/non-intervention, and completion/norm compliance. \cmark/\xmark indicate explicit/no explicit coverage; \pcmark/- indicate partial/unreported human QA review. AIRIS alone spans all columns under both causal-prefix and full-video settings.}
  \label{tab1}
  \setlength{\tabcolsep}{2.1pt}
  \renewcommand{\arraystretch}{1.08}
  \resizebox{\textwidth}{!}{%
  \begin{tabular}{lcccccccccccccc}
    \toprule
    \multirow{2}{*}{\textbf{Benchmark}} &
    \multicolumn{4}{c}{\cellcolor{tableprofile}\textbf{Benchmark profile}} &
    \multicolumn{2}{c}{\cellcolor{tablesetting}\textbf{Input setting}} &
    \multicolumn{2}{c}{\cellcolor{tableattention}\textbf{Attention}} &
    \multicolumn{2}{c}{\cellcolor{tableintention}\textbf{Intention}} &
    \multicolumn{2}{c}{\cellcolor{tablereaction}\textbf{Reaction}} &
    \multicolumn{2}{c}{\cellcolor{tableinteraction}\textbf{Interaction}} \\
    \cmidrule(lr){2-5}\cmidrule(lr){6-7}\cmidrule(lr){8-9}\cmidrule(lr){10-11}\cmidrule(lr){12-13}\cmidrule(lr){14-15}
    & \cellcolor{tableprofile}\makecell{Ego\\view} & \cellcolor{tableprofile}\makecell{QA\\count} & \cellcolor{tableprofile}\makecell{QA\\dimensions} & \cellcolor{tableprofile}\makecell{Human\\review}
    & \cellcolor{tablesetting}\makecell{Causal\\streaming} & \cellcolor{tablesetting}\makecell{Full\\video}
    & \cellcolor{tableattention}\makecell{Interactant\\grounding} & \cellcolor{tableattention}\makecell{Cue-target\\grounding}
    & \cellcolor{tableintention}\makecell{Latent\\goal} & \cellcolor{tableintention}\makecell{Intent\\Decidability}
    & \cellcolor{tablereaction}\makecell{Ego-action\\decision} & \cellcolor{tablereaction}\makecell{Non-\\intervention}
    & \cellcolor{tableinteraction}\makecell{Completion /\\understanding} & \cellcolor{tableinteraction}\makecell{Social-norm\\compliance} \\
    \midrule
    StreamingBench~\citep{lin2026streamingbench}        & \xmark & 4,500        & 18    & \cmark     & \cmark & \xmark & \xmark & \xmark & \xmark & \cmark & \xmark & \cmark & \xmark & \xmark \\
    OVO-Bench~\citep{niu2025ovobench}                   & \cmark & 2,814        & 12    & \cmark     & \cmark & \xmark & \xmark & \xmark & \xmark & \cmark & \xmark & \cmark & \xmark & \xmark \\
    RTV-Bench~\citep{xun2025rtvbench}                   & \cmark & 4,631        & 8     & \cmark     & \cmark & \xmark & \xmark & \xmark & \xmark & \xmark & \xmark & \xmark & \xmark & \xmark \\
    EgoSAT~\citep{lei2026egosat}                        & \cmark & 4,800 & 6     & -          & \cmark & \xmark & \xmark & \xmark & \xmark & \xmark & \xmark & \xmark & \xmark & \xmark \\
    ESTP-Bench~\citep{zhang2025eyeswideopen}            & \cmark & 2,264        & 14    & \cmark     & \cmark & \cmark & \xmark & \xmark & \xmark & \cmark & \cmark & \cmark & \xmark & \xmark \\
    ProactiveVideoQA~\citep{wang2025proactivevideoqa}   & \cmark & 1,427        & 4     & -          & \cmark & \xmark & \xmark & \xmark & \xmark & \cmark & \cmark & \cmark & \xmark & \xmark \\
    IPIBench~\citep{li2026ipibench}                     & \cmark & 3,738        & 9     & -          & \cmark & \xmark & \xmark & \xmark & \xmark & \xmark & \cmark & \cmark & \xmark & \xmark \\
    EgoSocial~\citep{wang2025egosocial} & \cmark & 13,500 & 9 & \cmark & \xmark & \cmark & \xmark & \xmark & \xmark & \xmark & \xmark & \cmark & \xmark & \xmark \\
    Human-MME~\citep{liu2025humanmme}                   & \xmark & 19,945       & 21 & -         & \xmark & \xmark & \xmark & \xmark & \cmark & \xmark & \xmark & \xmark & \xmark & \xmark \\
    HumanPCR~\citep{li2026humanpcr}                     & \xmark & 6,618        & 34 & \cmark     & \xmark & \cmark & \xmark & \xmark & \cmark & \xmark & \xmark & \xmark & \xmark & \xmark \\
    SIV-Bench~\citep{kong2026sivbench}                  & \cmark & 5,455        & 10    & \pcmark & \xmark & \cmark & \xmark & \xmark & \cmark & \xmark & \xmark & \xmark & \cmark & \xmark \\
    R$^3$-VQA~\citep{niu2026readtheroom}                & \xmark & 5,156        & 4     & \cmark     & \xmark & \cmark & \xmark & \xmark & \cmark & \xmark & \xmark & \xmark & \xmark & \xmark \\
    GRASP-Bench~\citep{kim2026grasp}                    & \xmark & 1,000        & 16    & \pcmark & \xmark & \xmark & \cmark & \cmark & \xmark & \xmark & \xmark & \xmark & \xmark & \xmark \\
    EgoNormia~\citep{rezaei2025egonormia}               & \cmark & 5,559        & 7 & \cmark     & \xmark & \xmark & \xmark & \xmark & \xmark & \xmark & \cmark & \xmark & \xmark & \cmark \\
    RobotEQ~\citep{fang2026roboteq}                     & \cmark & 6,639        & 8 & \cmark     & \xmark & \xmark & \cmark & \xmark & \xmark & \xmark & \cmark & \cmark & \xmark & \cmark \\
    \midrule
    \rowcolor{linecolor1}\textbf{AIRIS (Ours)}          & \cmark & \textbf{8,000} & \textbf{36} & \cmark & \cmark & \cmark & \cmark & \cmark & \cmark & \cmark & \cmark & \cmark & \cmark & \cmark \\
    \bottomrule
  \end{tabular}%
  }
\end{table*}

\section{The AIRIS Benchmark}

\subsection{Benchmark Formulation}
\textbf{Problem Formulation.}
We evaluate social embodied intelligence as an online decision process over a first-person stream, spanning Attention, Intention, Reaction, and Interaction.
A social episode is a time-bounded interpersonal exchange observed as $S = \{o_t\}_{t=0}^{T_{\max}}$, where each $o_t = (v_t, a_t)$ contains only first-person RGB frames $v_t$ and acoustic signals $a_t$ (raw audio for omni-modal models, or time-aligned ASR transcripts for standard VLMs).
At query time $\tau$, a model may access only the causal prefix $O_{\le \tau} = \{o_t \mid t \le \tau\}$.
Each question is evaluated from its designated observation prefix. The four task groups share episode annotations and temporal anchors, while their predictions are scored separately.

\textbf{Temporal Anchors.} 
To track the progression of a social exchange, we map the causal chain of each episode to four precise temporal anchors:
\begin{itemize}[leftmargin=1.5em,itemsep=2pt,topsep=2pt]
    \item $T_0$ (\textit{Cue Onset}): the first manifestation of a social cue (e.g., a gaze shift or gesture onset), marking the start of the \emph{Attention} window.
    \item $T_1$ (\textit{Decidability Onset}): the earliest moment at which human experts can unambiguously determine the partner's intent.
    \item $T_2$ (\textit{Reaction Onset}): the onset of the ego-agent's substantive physical or verbal response.
    \item $T_3$ (\textit{Interaction Resolution}): the conclusion of the immediate exchange.
\end{itemize}

\begin{figure*}[t]
    \centering
    \captionsetup{skip=3pt}
    \includegraphics[width=\textwidth]{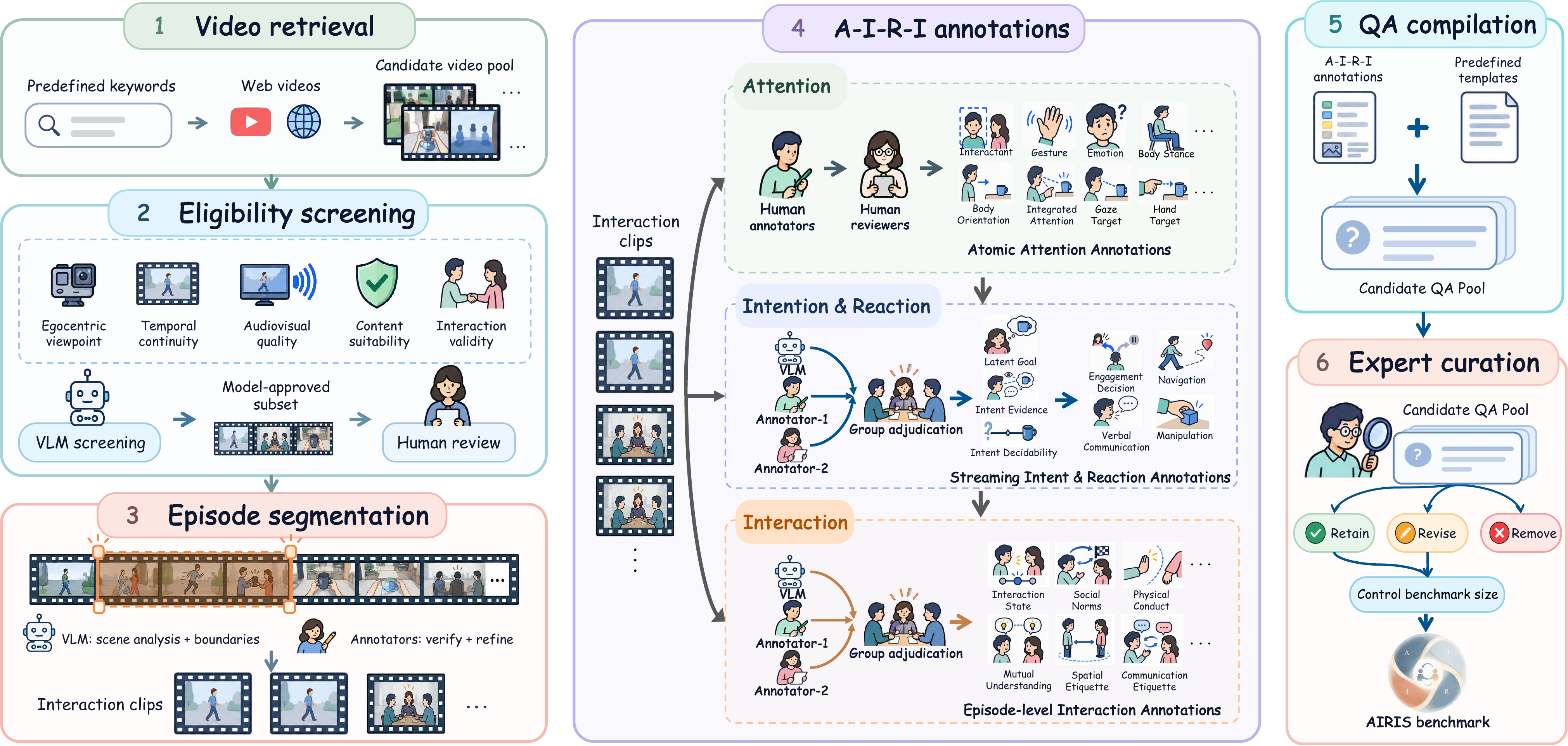}
    \caption{\textbf{AIRIS construction pipeline.} Web videos are screened, human-verified, and segmented into episodes. Clips then receive fully manual Attention labels and model-assisted, multi-annotator Intention, Reaction, and Interaction labels with group adjudication. Verified A-I-R-I annotations are templated into a candidate question pool and expert-curated into the final benchmark.}
    \label{fig:construction}
\end{figure*}

\textbf{The A-I-R-I Task Suite.} 
To operationalize this causal timeline into measurable capabilities, AIRIS structures the evaluation into four interdependent task layers, encompassing 36 fine-grained capability dimensions, as shown in Figure~\ref{fig:teaser}. 
Full capability definitions appear in \textit{Supplementary Section C.1}, with example questions and reviewed answers in \textit{Supplementary Sections E.1-E.4}.
All tasks are formulated as closed-set multiple-choice to ensure objective and reproducible evaluation. 
Specifically, the suite unfolds along the interaction timeline as follows:
\begin{itemize}[leftmargin=1.5em,itemsep=4pt,topsep=4pt]
    \item \emph{Attention (Understanding observable social cues):} This layer models attention as a sequential, two-step grounding process. First, the model must allocate the ego-agent's focus to localize the potential interaction partner within a complex environment. Second, it must analyze the partner's own attention and observable state, including gaze and pointing targets, as well as gestures, poses, and emotions. By evaluating interactant identification and target anchoring strictly from the $O_{\le \tau}$ stream, this layer ensures the model correctly isolates the relevant person and their focus before inferring deeper intent.
    
    \item \emph{Intention (Inferring the partner's latent goals):} Given the grounded evidence, this layer asks what the partner is attempting to achieve. Intention is therefore a timed inference of a latent goal. First, the model must decide whether the goal is already supportable from $O_{\le \tau}$. Second, only when $\tau \ge T_1$ must it recover that goal and select the visual, vocal, or verbal evidence that justifies the inference. For prefixes with $\tau < T_1$, ``Insufficient evidence'' is the explicitly correct answer. This strictly separates calibrated waiting from premature guessing.
    
    \item \emph{Reaction (Deciding ego-agent decisions):} 
    This layer shifts from interpreting the partner to deciding what the ego-agent should do now, given the inferred intent and the evidence available in $O_{\le \tau}$. First, the model must make an engagement decision: whether to act, communicate, keep observing, or refrain. Second, only if engagement is warranted must it specify the action domain, and the concrete sub-action. Waiting and non-intervention remain valid targets when $\tau < T_1$ or when an action would interrupt an ongoing activity.
        
    \item \emph{Interaction (Evaluate completeness and norms):} Evaluated post-hoc at $\tau \ge T_3$, this layer assesses the episode's overall state and outcome. First, the model must track the episode state and detect whether a social violation occurs, and if so of what type. Second, at $\tau \ge T_3$ it must judge the completed exchange along task completeness, mutual understanding, spatial etiquette, physical demeanor, communication etiquette, and overall interaction category. 
\end{itemize}

\subsection{Dataset Construction}
\label{sec:dataset}
To construct AIRIS under rigorous quality control, we employ a six-layered human-model collaborative pipeline spanning video curation, structured annotation, and benchmark assembly (Figure~\ref{fig:construction}). Quality assurance is incorporated throughout the pipeline: candidate videos undergo VLM screening and human verification, temporal proposals are manually refined, structured labels are reviewed and adjudicated, and every generated question receives expert inspection. 
This layered design prevents unverified model outputs from propagating directly into the benchmark and preserves a traceable connection from each released question to its source clip, temporal context, and reviewed annotation. More details in \textit{Supplementary Sections C.3 and C.4}.

\textbf{Keyword-guided video retrieval.} We first define a keyword inventory covering the target interaction contexts and use it to retrieve candidate egocentric videos from publicly accessible web platforms. At this stage, retrieval serves only to construct a broad candidate pool; the presence of a first-person viewpoint or an interpersonal exchange does not by itself qualify a video for inclusion.

\textbf{Model-assisted eligibility screening.}
A proprietary VLM screens candidate videos against five criteria: egocentric viewpoint, temporal continuity, audiovisual quality, content suitability, and interaction validity. These criteria distinguish usable first-person social exchanges from footage with disrupted event sequences, insufficient perceptual evidence, or no meaningful interpersonal interaction. Human reviewers reassess the model-approved subset using the same criteria; only videos passing both stages proceed to episode segmentation.

\textbf{Human-refined episode segmentation.}
For eligible videos, a proprietary VLM analyzes the scene and proposes temporal boundaries for interaction episodes. Human reviewers verify the proposed episodes and refine their boundaries under the same annotation guidelines, preserving the context needed to interpret the exchange. The approved intervals are extracted as egocentric interaction clips.

\textbf{Layer-specific A-I-R-I annotation.}
We tailor the annotation protocol for each clip to the distinct evidential demands of the A-I-R-I layers. 
The Attention layer focuses on perceptually subtle yet directly observable social cues. To minimize missed cues and spatial-grounding errors from automated pre-annotation, we annotate this layer entirely by hand. One annotator identifies the interactant, observable social cues, their targets, and the corresponding spatial grounding, while a second human reviewer verifies the resulting labels. 
For streaming Intention and Reaction annotations, whose temporal boundaries and semantic interpretations can be ambiguous, a strong proprietary VLM and two human annotators separately propose candidate temporal anchors, partner intentions, and Ego responses based on the admissible causal prefixes. Interaction annotations follow the same model-assisted, multi-annotator protocol, covering both the evolving interaction state and resolved outcomes such as completeness, mutual understanding, and norm compliance. An expert panel reviews the candidate annotations, resolves disagreements against the video evidence, and determines the final labels. Model outputs therefore serve only as annotation proposals and are never accepted as ground truth without human verification.

\textbf{Template-based question compilation.} The verified annotations are converted into a candidate question pool using predefined templates associated with the 36 capability dimensions. Each generated item retains a traceable link to its source clip, supporting annotation, admissible temporal context, and answer key. Streaming items expose only the causal observation prefix $O_{\leq\tau}$. In particular, Intention questions queried before $T_1$ assign ``Insufficient evidence'' as the correct response rather than requiring a premature intent prediction. Interaction-state and online-compliance questions likewise use the prefix available at their query time, whereas resolved outcome questions use the episode through $T_3$. This construction prevents future evidence from entering online decisions while preserving complete episodes for retrospective outcome assessment. 

\textbf{Expert question review and benchmark finalization.} Finally, expert reviewers inspect each candidate item together with its video evidence, temporal scope, question, answer options, and proposed answer key. Reviewers retain valid items, revise correctable items, and remove questions with insufficient evidence, ambiguous or redundant options, non-unique answers, inconsistent temporal context, or future-information leakage. This question-level review is conducted separately from annotation adjudication because a verified structured label does not necessarily yield an unambiguous evaluation item. The resulting review decisions control benchmark size while ensuring that every released question is supported by its designated visual, acoustic, and temporal evidence.

\begin{figure}[t]
    \centering
    \includegraphics[width=0.5\linewidth,trim=14pt 5pt 14pt 6pt,clip]{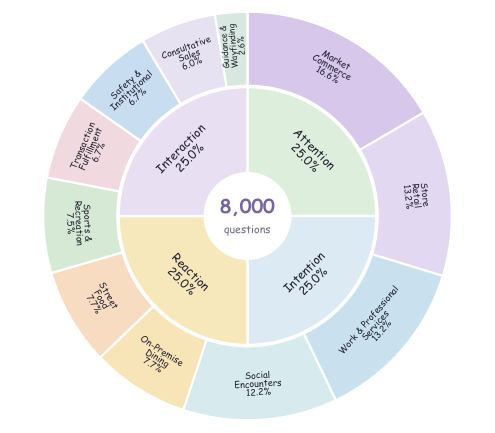}
    \caption{\textbf{AIRIS composition.} The inner ring shows the balanced A-I-R-I hierarchy; the outer ring illustrates the distribution across 11 interaction contexts.}
    \label{fig:benchmark-statistics}
\end{figure}

\subsection{Evaluation Protocol and Metrics}
\label{sec:metrics}
We evaluate AIRIS with a balanced top-level score complemented by stage-specific diagnostic probes that 
penalize ungrounded, collapsed, or premature predictions. Individual items are scored in $[0,1]$ by exact 
match (single-choice, yes/no, and bounding-box selection) or set F1 (multi-choice), with invalid or 
unparseable outputs receiving zero credit. Aggregating item scores within each A-I-R-I stage yields four 
question-micro layer scores, namely $S_{\mathrm{Att}}$, $S_{\mathrm{Int}}$, $S_{\mathrm{Rea}}$, and $S_{\mathrm{Inter}}$, whose unweighted mean defines the headline metric:
\begin{equation}
    \mathrm{Overall} = \frac{1}{4}\left(S_{\mathrm{Att}} + S_{\mathrm{Int}} + S_{\mathrm{Rea}} + S_{\mathrm{Inter}}\right).
    \label{eq:overall}
\end{equation}

Beyond aggregate stage scores, we isolate the primary reasoning bottlenecks within each layer. For \textbf{Attention}, we report \emph{Semantic target} (identifying the attended entity) and \emph{Grounding} (localizing its bounding box) separately, as naming a referent without spatial localization fails to support downstream embodied decisions. For \textbf{Intention}, because ``insufficient evidence'' is the ground-truth label prior to $T_1$, plain accuracy cannot distinguish calibrated waiting from lucky guessing. We therefore introduce the \emph{Intent Anticipation Score} (IAS) to evaluate streamed intent recovery. For each event, let $t_f$ be the first query timestamp satisfying $t_f \ge T_1$, and let $p \ge T_1$ be the earliest timestamp from which the model maintains stable correctness. The normalized anticipation margin is defined as
\begin{equation}
    m =
    \begin{cases}
        \operatorname{clip}\!\left(\dfrac{T_2 - p}{T_2 - t_f},\, 0,\, 1\right), & \text{if } p \text{ exists and } T_2 > t_f, \\[4pt]
        0, & \text{otherwise},
    \end{cases}
    \label{eq:anticipation-margin}
\end{equation}
which rewards earlier post-$T_1$ stabilization ahead of the reaction onset $T_2$. Discounting this margin by premature guessing yields the event-level score:
\begin{equation}
    \mathrm{IAS}_{\mathrm{event}} = m \cdot (1 - \mathrm{FCR}),
    \label{eq:ias-event}
\end{equation}
where the \emph{False Commitment Rate} ($\mathrm{FCR}$) is the fraction of pre-$T_1$ queries that prematurely commit to a specific intent rather than abstaining. Macro-averaging $\mathrm{IAS}_{\mathrm{event}}$ across evaluable events gives the final $\mathrm{IAS}$.

For \textbf{Reaction}, we report \emph{Engage} as the macro-F1 between active intervention and non-intervention classes to penalize collapsed policies, alongside a strict \emph{Joint-EM} diagnostic requiring simultaneous exact matches across engagement, action domain, subaction, and applicable speech function. Finally, \textbf{Interaction} evaluates \emph{Compliance} across spatial, physical, and communication etiquette at episode resolution, as well as \emph{Violation}, which treats ``violation present'' as the positive class to prevent exploitation of the heavily skewed negative baseline. Additional metric definitions and full per-stage diagnostics are provided in \textit{Supplementary Sections C.2 and D.2}, respectively.

\subsection{Benchmark Statistics}
\label{sec:benchmark-statistics}

\begin{table*}[h]
  \centering
  \caption{\textbf{Main results on AIRIS benchmark.} Models are grouped by type and sorted by Overall within each group. Scores are in $[0,1]$ (higher is better); \textbf{bold}/\underline{underline} mark the best/second-best.}
  \label{tab:main-results}
  \begingroup
  \scriptsize
  \newcommand{\modelref}[1]{\mbox{\fontsize{6}{7}\selectfont\citeyearpar{#1}}}
  \setlength{\tabcolsep}{0.8pt}
  \renewcommand{\arraystretch}{1.02}
  \begin{tabular*}{\textwidth}{@{\extracolsep{\fill}}>{\raggedright\arraybackslash}p{0.30\textwidth} ccc ccc ccc ccc >{\columncolor{tablesetting}}c@{}}
    \toprule
    \multirow{2}{*}{\textbf{Model}} &
    \multicolumn{3}{c}{\cellcolor{tableattention}\textbf{Attention}} &
    \multicolumn{3}{c}{\cellcolor{tableintention}\textbf{Intention}} &
    \multicolumn{3}{c}{\cellcolor{tablereaction}\textbf{Reaction}} &
    \multicolumn{3}{c}{\cellcolor{tableinteraction}\textbf{Interaction}} &
    \multirow{2}{*}{\textbf{Overall}} \\
    \cmidrule(lr){2-4}\cmidrule(lr){5-7}\cmidrule(lr){8-10}\cmidrule(lr){11-13}
    & \textbf{All} & Target & Ground & \textbf{All} & Online & IAS & \textbf{All} & Engage & Joint & \textbf{All} & Comp. & Viol. & \\
    \midrule
    \multicolumn{14}{l}{\cellcolor{black!4}\textit{Closed-source models}} \\
    Gemini-3.7-Flash \modelref{googledeepmind2026gemini37flash} & \textbf{0.936} & \textbf{0.951} & \textbf{0.948} & \textbf{0.569} & \textbf{0.569} & \textbf{0.237} & 0.638 & 0.546 & 0.083 & \textbf{0.760} & 0.903 & 0.525 & \textbf{0.726} \\
    Gemini-3.1-Pro \modelref{googledeepmind2026gemini31pro} & 0.851 & 0.897 & 0.820 & 0.496 & \underline{0.523} & 0.126 & 0.620 & 0.530 & 0.111 & \textbf{0.760} & 0.902 & 0.495 & \underline{0.682} \\
    GPT-5.6-Sol \modelref{openai2026gpt56sol} & \underline{0.906} & \underline{0.925} & \underline{0.884} & 0.501 & 0.457 & 0.154 & 0.489 & 0.261 & 0.128 & 0.729 & 0.906 & 0.494 & 0.656 \\
    Doubao-seed-2.0-lite \modelref{bytedanceseed2026seed20lite} & 0.759 & 0.859 & 0.742 & 0.418 & 0.363 & 0.066 & 0.619 & 0.564 & 0.092 & 0.669 & 0.726 & 0.572 & 0.616 \\
    \midrule
    \multicolumn{14}{l}{\cellcolor{black!4}\textit{Open-source general-purpose models}} \\
    Qwen3.6-35B-A3B \modelref{qwenteam2026qwen3635ba3b} & 0.723 & 0.891 & 0.646 & 0.485 & 0.439 & 0.122 & \underline{0.655} & \underline{0.605} & 0.064 & \underline{0.738} & 0.881 & 0.577 & 0.650 \\
    GLM-5.3-Flash \modelref{zai2026glm53flash} & 0.801 & 0.882 & 0.802 & 0.501 & 0.455 & 0.128 & 0.590 & 0.531 & 0.067 & 0.638 & 0.706 & 0.470 & 0.633 \\
    Gemma-4-31B-IT \modelref{gemmateam2026gemma4} & 0.733 & 0.899 & 0.619 & 0.502 & 0.482 & \underline{0.167} & 0.535 & 0.394 & 0.123 & 0.728 & 0.907 & 0.495 & 0.624 \\
    DeepSeek-V4.1-Flash \modelref{deepseekai2026deepseekv41flash} & 0.707 & 0.829 & 0.627 & \underline{0.509} & 0.451 & 0.121 & 0.548 & 0.456 & 0.108 & 0.657 & 0.737 & 0.508 & 0.605 \\
    Qwen3.6-27B \modelref{qwenteam2026qwen3627b} & 0.649 & 0.836 & 0.528 & 0.480 & 0.483 & 0.119 & 0.545 & 0.383 & 0.128 & 0.733 & 0.905 & 0.501 & 0.602 \\
    Qwen3-VL-32B-Instruct \modelref{bai2025qwen3vl} & 0.682 & 0.886 & 0.598 & 0.472 & 0.481 & 0.133 & 0.479 & 0.217 & 0.144 & 0.723 & \textbf{0.923} & \textbf{0.589} & 0.589 \\
    Qwen3.8-27B \modelref{qwenteam2026qwen3827b} & 0.685 & 0.835 & 0.616 & 0.395 & 0.374 & 0.052 & 0.475 & 0.251 & 0.137 & 0.632 & 0.717 & 0.341 & 0.547 \\
    InternVL3-8B \modelref{zhu2025internvl3} & 0.583 & 0.778 & 0.513 & 0.417 & 0.375 & 0.082 & 0.518 & 0.271 & 0.123 & 0.603 & 0.724 & 0.318 & 0.530 \\
    InternVL3.5-30B-A3B \modelref{wang2025internvl35} & 0.598 & 0.727 & 0.564 & 0.426 & 0.458 & 0.095 & 0.442 & 0.134 & 0.144 & 0.639 & 0.754 & 0.511 & 0.526 \\
    InternVL3.5-8B \modelref{wang2025internvl35} & 0.574 & 0.768 & 0.490 & 0.406 & 0.374 & 0.094 & 0.458 & 0.309 & 0.119 & 0.562 & 0.689 & 0.351 & 0.500 \\
    LLaVA-OneVision-7B \modelref{li2025llavaonevision} & 0.500 & 0.757 & 0.346 & 0.387 & 0.380 & 0.078 & 0.379 & 0.132 & 0.144 & 0.602 & 0.657 & 0.453 & 0.467 \\
    Qwen3.5-4B \modelref{qwenteam2026qwen35} & 0.534 & 0.715 & 0.433 & 0.389 & 0.442 & 0.123 & 0.481 & 0.453 & 0.129 & 0.425 & 0.523 & 0.322 & 0.457 \\
    Llama-3.2-11B-Vision-Instruct \modelref{meta2024llama32vision} & 0.391 & 0.620 & 0.261 & 0.384 & 0.422 & 0.011 & 0.463 & 0.371 & 0.128 & 0.455 & 0.498 & 0.381 & 0.423 \\
    Qwen3.5-0.8B \modelref{qwenteam2026qwen35} & 0.381 & 0.537 & 0.278 & 0.299 & 0.276 & 0.036 & 0.506 & 0.517 & 0.044 & 0.398 & 0.353 & 0.329 & 0.396 \\
    \midrule
    \multicolumn{14}{l}{\cellcolor{black!4}\textit{Open-source embodied models}} \\
    HY-Embodied-VLM-1.0 \modelref{tencent2026hyembodiedvlm10} & 0.596 & 0.834 & 0.492 & 0.396 & 0.419 & 0.105 & \textbf{0.688} & 0.485 & 0.028 & 0.716 & 0.873 & 0.481 & 0.599 \\
    RynnBrain1.1-9B \modelref{li2026rynnbrain11} & 0.593 & 0.819 & 0.470 & 0.417 & 0.468 & 0.123 & 0.557 & \underline{0.605} & 0.082 & 0.736 & \underline{0.918} & \underline{0.583} & 0.576 \\
    RynnBrain-30B-A3B \modelref{dang2026rynnbrain} & 0.656 & 0.777 & 0.596 & 0.435 & 0.471 & 0.091 & 0.410 & 0.245 & \textbf{0.147} & 0.674 & 0.723 & 0.508 & 0.544 \\
    VeBrain-7B \modelref{luo2025vebrain} & 0.546 & 0.801 & 0.440 & 0.449 & 0.494 & 0.111 & 0.470 & 0.154 & 0.141 & 0.701 & 0.828 & 0.502 & 0.541 \\
    Pelican1.0-VL-7B \modelref{zhang2025pelicanvl} & 0.540 & 0.724 & 0.425 & 0.462 & 0.503 & 0.099 & 0.440 & 0.128 & 0.144 & 0.700 & 0.875 & 0.470 & 0.536 \\
    MiMo-Embodied-7B \modelref{hao2025mimoembodied} & 0.533 & 0.761 & 0.391 & 0.440 & 0.407 & 0.114 & 0.643 & 0.558 & 0.056 & 0.487 & 0.669 & 0.466 & 0.526 \\
    RoboBrain2.0-7B \modelref{baaiteam2025robobrain2} & 0.511 & 0.809 & 0.349 & 0.439 & 0.447 & 0.054 & 0.471 & 0.184 & \underline{0.146} & 0.597 & 0.696 & 0.379 & 0.505 \\
    Embodied-R1-3B-V1 \modelref{yuan2026embodiedr1} & 0.475 & 0.703 & 0.375 & 0.408 & 0.376 & 0.073 & 0.468 & 0.126 & 0.144 & 0.569 & 0.635 & 0.412 & 0.480 \\
    HY-Embodied-0.5 \modelref{tencent2026hyembodied05} & 0.482 & 0.694 & 0.464 & 0.368 & 0.388 & 0.022 & 0.563 & \textbf{0.606} & 0.065 & 0.429 & 0.353 & 0.318 & 0.461 \\
    \midrule
    \multicolumn{14}{l}{\cellcolor{black!4}\textit{Open-source streaming video models}} \\
    JoyAI-VL-Interaction-Preview \modelref{yao2026joyaivlinteraction} & 0.639 & 0.761 & 0.628 & 0.436 & 0.408 & 0.113 & 0.610 & 0.508 & 0.029 & 0.586 & 0.635 & 0.417 & 0.568 \\
    AURA-8B \modelref{lu2026aura} & 0.624 & 0.834 & 0.574 & 0.472 & 0.490 & 0.113 & 0.434 & 0.146 & 0.144 & 0.601 & 0.766 & 0.510 & 0.533 \\
    StreamingVLM \modelref{xu2026streamingvlm} & 0.456 & 0.753 & 0.292 & 0.400 & 0.449 & 0.029 & 0.536 & 0.302 & 0.141 & 0.655 & 0.785 & 0.413 & 0.512 \\
    Flash-VStream-Qwen-7B \modelref{zhang2025flashvstream} & 0.470 & 0.675 & 0.367 & 0.396 & 0.398 & 0.065 & 0.376 & 0.131 & 0.142 & 0.631 & 0.682 & 0.436 & 0.468 \\
    VITA-1.5 \modelref{fu2025vita15} & 0.460 & 0.638 & 0.412 & 0.388 & 0.435 & 0.091 & 0.420 & 0.344 & 0.103 & 0.477 & 0.523 & 0.468 & 0.436 \\
    \midrule
    \multicolumn{14}{l}{\cellcolor{black!4}\textit{Open-source omni-modal models}} \\
    Qwen2.5-Omni-7B \modelref{xu2025qwen25omni} & 0.507 & 0.745 & 0.375 & 0.453 & 0.485 & 0.070 & 0.449 & 0.126 & 0.144 & 0.705 & 0.855 & 0.510 & 0.529 \\
    Qwen3-Omni-30B-A3B-Instruct \modelref{xu2025qwen3omni} & 0.646 & 0.814 & 0.590 & 0.419 & 0.498 & 0.153 & 0.390 & 0.138 & 0.141 & 0.647 & 0.746 & 0.415 & 0.526 \\
    OmniVinci \modelref{ye2026omnivinci} & 0.546 & 0.733 & 0.456 & 0.401 & 0.410 & 0.116 & 0.340 & 0.126 & 0.144 & 0.651 & 0.740 & 0.454 & 0.484 \\
    Qwen2.5-Omni-3B \modelref{xu2025qwen25omni} & 0.477 & 0.762 & 0.338 & 0.433 & 0.466 & 0.065 & 0.467 & 0.126 & 0.144 & 0.520 & 0.534 & 0.200 & 0.474 \\
    Phi-4-Multimodal-Instruct \modelref{microsoft2025phi4multimodal} & 0.435 & 0.541 & 0.368 & 0.398 & 0.426 & 0.000 & 0.435 & 0.133 & 0.137 & 0.356 & 0.172 & 0.301 & 0.406 \\
    \bottomrule
  \end{tabular*}
  \par\vspace{3pt}
  \parbox{\textwidth}{\textit{Abbreviations:} All = stage Overall; Target = semantic target; Ground = grounding; IAS = Intent Anticipation Score; Engage = engagement Macro-F1; Joint = full Reaction-chain correctness; Comp. = compliance; Viol. = violation.}
  \endgroup
\end{table*}

The AIRIS benchmark comprises 8,000 questions grounded in 357 egocentric clips and 276 source videos, totaling 10.91 hours of footage. Clip durations range from 3.0 to 767.3 seconds with a mean of 110.0 seconds, capturing both brief encounters and extended interactions. The dataset averages 22.4 questions per clip and 11.46 sampled frames per question, comprehensively covering 36 capabilities, 11 contexts, and 230 fine-grained subscenes.

These clips span diverse real-world settings as shown in Figure~\ref{fig:benchmark-statistics}. The predominant categories are Market Commerce with 1,324 questions, Store Retail with 1,055, Work \& Professional Services with 1,052, and Social Encounters with 976, alongside environments like dining, recreation, and wayfinding. Through rigorous expert curation, AIRIS maintains a perfectly balanced top-level hierarchy that allocates exactly 2,000 questions to each A-I-R-I layer.

To probe different reasoning facets, the benchmark employs four response formats: single-choice at 63.1\%, multi-choice at 25.6\%, bounding-box selection at 6.4\%, and yes/no at 5.0\%. Furthermore, 87.7\% of questions feature streaming triggers and 41.1\% incorporate speech context. To account for heterogeneous question density across clips, we report both per-capability and clip-macro metrics.

\section{Experiments}

\begin{figure*}[t]
  \centering
  \includegraphics[width=\textwidth,trim={9bp 6bp 7bp 6bp},clip]{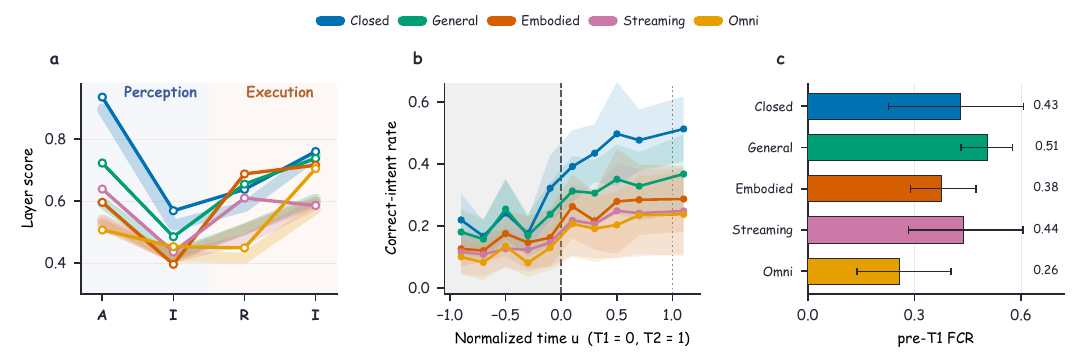}
  \caption{\textbf{Uneven performance across A-I-R-I capabilities.} (a) A-I-R-I performance: category averages (thick, translucent) and best models (thin, markers). (b) Correct-intent rate vs.\ normalized interaction time.
  (c) Mean False Commitment Rate (FCR) before $T_1$.
  }
  \label{fig:perception_anticipation}
  \vspace{-0.5em}
\end{figure*}

\subsection{Experimental Setup}

\textbf{Evaluated Models.} We evaluate 45 models in total to provide a broad landscape of social embodied capabilities. Due to space constraints, the main text focuses on 37 representative state-of-the-art models spanning five categories: closed-source models, open-source general-purpose VLMs, open-source embodied models, open-source streaming video models, and open-source omni-modal models. Results for these 37 models are reported in Table~\ref{tab:main-results}. Supplementary Sections D provide the full model list and results, respectively.

\textbf{Evaluation Harness.} Models are evaluated zero-shot on the 8,000-question AIRIS benchmark. For each query at timestamp $\tau$, models receive only the causal video prefix $O_{\le \tau}$. Acoustic context is supplied as raw audio for omni-modal models and as time-aligned ASR transcripts for standard VLMs. Questions are formatted as multiple-choice, yes/no, or bounding-box selection tasks.

\subsection{Experimental Results and Insights}

Table~\ref{tab:main-results} presents the evaluation results of 37 representative models on the AIRIS benchmark. Closed-source models, led by Gemini-3.7-Flash, establish a dominant baseline in perceptual grounding and intent recognition. Gemini-3.7-Flash achieves the highest overall score of 0.726 and significantly outperforms open-source counterparts in Attention at 0.936 and Intention at 0.569. 
Conversely, open-source models, particularly those fine-tuned for embodied tasks, demonstrate highly competitive or superior decision-making capabilities. HY-Embodied-VLM-1.0 achieves the highest overall Reaction score of 0.688, while Qwen3-VL-32B-Instruct leads in Interaction compliance at 0.923 and violation detection at 0.589. More experimental results are provided in Supplementary Section D.

\textbf{The Perception-Decision Gap.} We observe a pronounced capability divide across the A-I-R-I causal chain (Figure~\ref{fig:perception_anticipation}a). While general-purpose VLMs excel at passive observation and semantic reasoning (Attention and Intention), their advantage vanishes in the downstream Reaction and Interaction layers. Instead, embodied-specific models take the lead in generating socially appropriate, executable responses. This divergence reveals that ``understanding'' does not naturally translate to ``doing''; generating compliant embodied actions in dynamic social scenarios requires specialized action-oriented alignment beyond standard semantic pre-training.

\textbf{Miscalibrated Causal Anticipation.} Streaming intent anticipation remains a critical bottleneck across all models. Even the leading Gemini-3.7-Flash reaches an Intent Anticipation Score (IAS) of only 0.237, and most models score near 0.1. The anticipation curves in Figure~\ref{fig:perception_anticipation}b make the temporal structure explicit. The correct intent rate stays low before the decidability onset ($u<0$), climbs only after $T_1$, and remains modest even at the reaction onset $T_2$ (at most $\sim$0.5 for closed source models). Crucially, this early deficit is \emph{not} calibrated waiting. As Figure~\ref{fig:perception_anticipation}c shows, the False Commitment Rate (FCR) before $T_1$ remains substantial across categories (0.26 to 0.51). This fraction measures queries that commit to a specific intent when ``insufficient evidence'' is the correct answer. Individual models range from near abstention (Phi-4-Multimodal, 0.08) to aggressive guessing (GLM-5.3-Flash, 0.64; Qwen3.6-35B-A3B, 0.63; GPT-5.6-Sol, 0.57). Even Gemini-3.7-Flash commits in 42\% of early queries. Current models thus fail anticipation in two opposite ways: overconfident premature commitment or blind passivity, rather than exercising the evidence gated inference the task rewards.

\textbf{No model closes the full reaction decision chain.} Our reaction evaluation reports two complementary views: an aggregate score (R-All) crediting each sub-decision (engagement, action domain, and the concrete subaction or utterance) and a stricter Joint-EM requiring all of them to be correct simultaneously for the same event. Each sub-decision is individually non-trivial, yet requiring them to hold jointly makes performance collapse multiplicatively. No model exceeds a Joint-EM of 0.147, which is far below what the component scores alone would predict. Deciding \emph{whether} to engage, or planning a plausible subaction, is therefore not enough on its own; only their correct, end-to-end composition yields a trustworthy social response. The strict lens accordingly re-orders the leaderboard. HY-Embodied-VLM-1.0, Qwen3.6-35B-A3B, and MiMo-Embodied-7B lead on aggregate Reaction (0.688, 0.655, 0.643) but fall to 0.028, 0.064, and 0.056 under the joint constraint. This integrative competence, rather than isolated perception, intent, or action, is the true prerequisite for trustworthy social embodied intelligence, and it remains largely unmet.

\textbf{Unrealized Architectural Dividends in Social Reasoning.} Models specifically designed for native streaming video (e.g., JoyAI-VL, StreamingVLM) or omni-modal audio-visual processing (e.g., Qwen3-Omni) do not exhibit significant advantages over traditional frame-sampled VLMs (e.g., Qwen3.6-35B) in overall social embodied intelligence. The architectural benefits of continuous stream processing appear confined to low-level perception or speech dialogue. For tasks requiring deep Theory of Mind and adherence to complex social norms, these architectural shifts have not yet compensated for the lack of high-level social reasoning capabilities.

\textbf{Explicit reasoning does not reliably help streaming interaction.} Allocating more test-time computation to explicit chain-of-thought is a natural way to attack our causal streaming tasks, yet under identical streaming input, scored with a thinking-aware re-parse that corrects a systematic answer-parsing artifact, it brings no overall benefit: Qwen3.6-27B ties its no-think baseline (0.1 points) and Qwen3.6-35B drops (6.4 points), while Qwen3-Omni gains only marginally (2.9 points).

\clearpage
\begin{wrapfigure}[34]{l}{0.54\textwidth}
  \centering
  \captionsetup{skip=3pt,font=footnotesize}
  \includegraphics[width=0.96\linewidth,trim={13.5bp 0bp 1bp 15bp},clip]{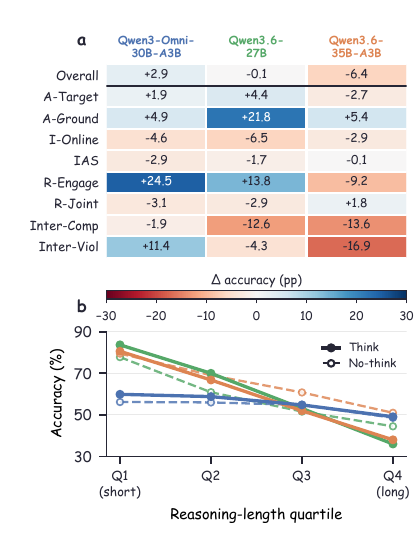}
  \caption{\textbf{Think versus no-think.} (a) Accuracy change across AIRIS columns. (b) Accuracy by reasoning-length quartile (solid, think; dashed, no-think). Results use thinking-aware re-parsing.}
  \label{fig:think}
  \vspace{-0.8em}
\end{wrapfigure}
\noindent The effect is structured along the A-I-R-I chain (Figure~\ref{fig:think}a): reasoning tends to help multi-step spatial or decision sub-tasks, such as spatial grounding (A-Ground, +21.8 points for Qwen3.6-27B), but consistently hurts the columns that reward restraint under partial evidence, namely online intent (I-Online) and interaction compliance, across all models. The penalty is legible at the trace level (Figure~\ref{fig:think}b): accuracy falls with trace length, and although harder items independently elicit longer reasoning (a selection effect), thinking still widens the gap on the hardest Qwen3.6 items, whereas the restrained Qwen3-Omni is nearly unchanged. Traces carrying cognitive-revision markers (``however'', ``wait'', ``reconsider''), present in about 77\% of Qwen3.6 traces, lose roughly 27 to 30 points of accuracy: when perception is insufficient, extended deliberation fills the visual gap with language and overturns a correct first instinct. Chain-of-thought optimized for static text problems is thus not yet adapted to the real-time, causally restrained demands of streaming social interaction.
\par
\WFclear
\vspace{4\baselineskip}

\section{Conclusion}
We introduced AIRIS, a first-person causal streaming benchmark evaluating social embodied intelligence as a time-anchored progression across Attention, Intention, Reaction, and Interaction. Across 45 models, strong cue perception and intent recognition do not reliably translate to calibrated anticipation or complete embodied reactions. AIRIS thus shifts the evaluation focus from static social recognition to the dynamic timing of evidence sufficiency, action necessity, and response composition.

These findings highlight key priorities for future models. First, models must preserve calibrated uncertainty before an intent becomes decidable, committing only when supported by observable evidence. Second, perception, inference, and action specification must be optimized as a coupled decision process to ensure isolated strengths produce coherent reactions. Finally, test-time computation alone is insufficient; streaming reasoning must remain grounded in current sensory evidence and adapt its depth to uncertainty, avoiding the replacement of missing observations with unsupported verbal inference.

By making timing, restraint, and decision composition measurable under causal observation, AIRIS establishes a rigorous foundation for evaluating social embodied intelligence. Future work will extend this framework to interactive robot-in-the-loop settings, broader social contexts, and open-ended generation. Ultimately, this progress fosters human-centered embodied agents: ones that not only understand what people are doing, but also know when to wait, whether to act, and how to respond in ways that align with human social norms and facilitate seamless human-robot coordination.

\clearpage
\setlength{\bibsep}{5pt}
\bibliography{main}
\bibliographystyle{unsrtnat}

\clearpage
\appendix
\section*{Supplementary Material}
\renewcommand{\thesubsection}{\Alph{subsection}}
\renewcommand{\thesubsubsection}{\thesubsection.\arabic{subsubsection}}
\phantomsection
\refstepcounter{subsection}
\label{app:overview}

\begingroup
\setlength{\parindent}{0pt}
\setlength{\parskip}{3pt}
\newcommand{\airiscontentsentry}[3]{%
  \noindent\hspace*{#1}%
  \hyperref[#2]{\ref*{#2}\quad #3}%
  \nobreak\dotfill\nobreak\hyperref[#2]{\pageref*{#2}}\par
}
\hfill\textbf{Page}\par

\airiscontentsentry{0pt}{app:overview}{\textbf{Contents}}
\medskip
\airiscontentsentry{0pt}{app:related-work}{\textbf{Related Work}}
\airiscontentsentry{1.5em}{app:related-streaming}{Streaming and Proactive Video Understanding}
\airiscontentsentry{1.5em}{app:related-social}{Multimodal Social Interaction Understanding}
\airiscontentsentry{1.5em}{app:related-norms}{Social Norms and Embodied Decision Making}
\medskip
\airiscontentsentry{0pt}{app:benchmark-details}{\textbf{Supplementary Benchmark Details}}
\airiscontentsentry{1.5em}{app:capability-ontology}{A-I-R-I capability ontology}
\airiscontentsentry{1.5em}{app:supp-metrics}{Supplementary Evaluation Protocol and Metrics}
\airiscontentsentry{1.5em}{app:annotation-interface}{Expert Annotation Interfaces}
\airiscontentsentry{1.5em}{app:annotation-process}{Expert Annotation and Audit Process}
\medskip
\airiscontentsentry{0pt}{app:experimental-details}{\textbf{Supplementary Experiments and Results}}
\airiscontentsentry{1.5em}{app:evaluated-models}{Model evaluation pool}
\airiscontentsentry{1.5em}{app:extended-results}{Complete quantitative results}
\medskip
\airiscontentsentry{0pt}{app:qualitative}{\textbf{Benchmark Visualizations and Qualitative Analysis}}
\airiscontentsentry{1.5em}{app:attention-cases}{Attention cases}
\airiscontentsentry{1.5em}{app:intention-cases}{Intention cases}
\airiscontentsentry{1.5em}{app:reaction-cases}{Reaction cases}
\airiscontentsentry{1.5em}{app:interaction-cases}{Interaction cases}
\endgroup
\clearpage

\subsection{Related Work}
\label{app:related-work}
\subsubsection{Streaming and Proactive Video Understanding}
\label{app:related-streaming}
Most video-language benchmarks \citep{li2024mvbench, fu2025videomme, wu2024longvideobench} assume that a model can inspect a complete, temporally bounded clip before answering, whereas deployed assistants receive observations sequentially.
On the model side, VideoLLM-online \citep{chen2024videollmonline} introduced temporally aligned dialogue over continuous video streams and an asynchronous inference pipeline for real-time processing.
Subsequent systems target complementary bottlenecks: Flash-VStream \citep{zhang2025flashvstream} compresses long-running visual context into context and augmentation memories for low-latency query answering, whereas Dispider \citep{qian2025dispider} separates perception, decision, and reaction into asynchronous modules so that response generation does not suspend stream processing.
More recently, Streamo \citep{xia2026streaming} learns explicit silence, standby, and response states within a single autoregressive sequence, coupling response timing with content generation across multiple streaming tasks.
On the evaluation side, StreamingBench \citep{lin2026streamingbench} distributes timestamped questions throughout a stream, while OVO-Bench \citep{niu2025ovobench} organizes online understanding into backward tracing, real-time perception, and forward active responding, including waiting until sufficient evidence becomes available.
OmniMMI and SocialOmni \citep{wang2025omnimmi,xie2026socialomni} broaden this setting to multimodal interaction, proactive alerting, and turn taking.
EgoSAT and StreamGaze \citep{lei2026egosat,lee2025streamgaze} evaluate temporal reasoning from strict egocentric prefixes, with the latter incorporating gaze.
Complementary robot-egocentric tracking research evaluates persistent target-person grounding over long horizons, from large-scale benchmarking to referring-then-tracking models \citep{ye2025tpt,ye2026refertrack}.

Recent work has also shifted from answering predefined queries to deciding when and how an assistant should respond.
ViSpeak \citep{fu2025vispeak} extracts visual instructions from gestures and body language in streaming video; ESTP-Bench \citep{zhang2025eyeswideopen} evaluates proactive coherence and just-in-time responses; and ProAssist and EDITH \citep{zhang2025proassist,lee2026edith} use egocentric streams for proactive guidance and robot assistance.
IPIBench \citep{li2026ipibench} further evaluates proactive monitoring, task modification and cancellation, and interleaved reactive-proactive requests.
EgoServe \citep{gong2026vinci2} frames proactive assistance as a context-dependent intervention problem over multiple memory horizons.
Existing streaming and proactive benchmarks predominantly focus on object-centric task execution or user-directed assistance, treating humans primarily as task performers or commanders. Building on their real-time protocols, AIRIS shifts the paradigm to social embodiment. It evaluates the ego-agent as a social participant, diagnosing its ability to navigate interpersonal dynamics, anticipate social intents, and deliver norm-appropriate responses in shared environments.

\subsubsection{Multimodal Social Interaction Understanding}
\label{app:related-social}
A foundational requirement for social embodied intelligence is the ability to perceive and interpret human behavior from heterogeneous social signals.
Social-IQ \citep{zadeh2019socialiq} frames this ability as multimodal question answering over natural social situations, while Human-MME \citep{liu2025humanmme} and HumanPCR \citep{li2026humanpcr} provide broader evaluations spanning fine-grained human perception, comprehension, and reasoning.
SIV-Bench \citep{kong2026sivbench} further structures social video understanding into social scene understanding, state reasoning, and dynamics prediction.
Moving beyond answer accuracy, Social Genome \citep{mathur2025socialgenome} evaluates whether social inferences are explicitly grounded in visual, verbal, and vocal evidence as well as relevant external knowledge.
A complementary line grounds interaction evidence to relevant regions and participants \citep{zhou2024learning,ouyang2026multispeaker}.
Recently, attention-prediction research further links gaze to human-object interactions and behavior-aware context, while explainable formulations connect where attention falls to what is attended and why \citep{zhou2025behavior,zhou2025towards}.
Robot-centric work models individual attributes, group context, intent, and social actions from onboard observations \citep{jahangard2024jrdbsocial,bian2025robusthri}.
SocialGesture \citep{cao2025socialgesture} targets the recognition, temporal localization, and question answering of natural gestures in multi-person interactions.
The MultiMediate challenge series \citep{muller2022multimediate,muller2023multimediate} evaluates conversational behaviors such as eye contact, backchannels, agreement, engagement, and bodily behavior in group discussions.
The MMSI lineage \citep{lee2024modeling,li2025onlinemmsi,li2026omnimmsi} instead emphasizes identity-attributed social cues, including speaking targets and pronoun references, while GRASP \citep{kim2026grasp} connects high-level social reasoning to participant-specific gaze and gesture evidence.
At a higher reasoning level, MuMA-ToM \citep{shi2025mumatom} evaluates goals, beliefs, and nested beliefs in embodied multi-agent interactions.
In egocentric settings, EgoGazeVQA \citep{peng2025egogazevqa} studies gaze as an indicator of user intent, whereas R$^3$-VQA \citep{niu2026readtheroom} represents belief, intent, desire, and emotion through mental-physical causal chains.
Collectively, these benchmarks substantially advance the perception and interpretation of other people's attention, references, intentions, and social states.
AIRIS evaluates these capabilities alongside response selection, justified inaction, and assessment of observed interaction outcomes within the same time-anchored episodes.

\subsubsection{Social Norms and Embodied Decision Making}
\label{app:related-norms}
Understanding others is insufficient for social embodiment: an agent must also determine what it should do, when it should act, and which otherwise feasible actions are socially inappropriate.
EgoNormia \citep{rezaei2025egonormia} grounds this problem in egocentric video by evaluating socially appropriate next actions, their justifications, and acceptable alternatives under potentially competing physical-social norms.
RobotEQ \citep{fang2026roboteq} similarly evaluates action permissibility and spatial grounding from robot-view scenes, with an emphasis on deciding how, when, and whether to act without explicit instructions.
NormAct \citep{normact2026} extends norm evaluation to embodied planning, testing whether models can infer hidden social constraints and incorporate them into action sequences while still achieving an explicit task goal.
The SHREC benchmark \citep{lee2025hsri} evaluates whether foundation models can identify a robot's social errors and competencies, reason about their effects on interaction progression, and propose corrections.
Should I Help? \citep{ramnauth2024shouldihelp} challenges the assumption that assistance is always desirable and formalizes whether help is appropriate in terms of human and robot skills together with the social benefits and costs to the recipient.
NIABench \citep{assistance2026nonintrusive} further formulates non-intrusive assistance as a joint decision over when to act and what action to take while preserving the human's ongoing task flow.
Embodied robotics has also operationalized particular dimensions of socially appropriate behavior.
Habitat 3.0 \citep{puig2024habitat3} introduces human-in-the-loop social navigation and collaborative rearrangement tasks in which robots must coordinate with people, maintain distance, and yield shared space.
VLM-Social-Nav and Social-LLaVA \citep{song2025vlmsocialnav,payandeh2025socialllava} use vision-language models for socially aware navigation.
SONAR \citep{sonar2024} instead provides an adaptive norm-aware control architecture for social robots, combining explicit agent reasoning with learned contextual norms in human-robot conversations.
These studies address complementary aspects of norm recognition, embodied planning, assistance, navigation, and robot-behavior assessment, but generally evaluate them as separate decision problems.
AIRIS links norm-aware responses and justified inaction to the social evidence and inferred intent that precede them, enabling process-level diagnosis of both Reaction and Interaction within a single time-anchored streaming episode.

\subsection{Supplementary Benchmark Details}
\label{app:benchmark-details}

This section provides the granular details of the AIRIS benchmark construction and evaluation. It is organized into four parts: the formal capability ontology (Section~\ref{app:capability-ontology}), the supplementary evaluation protocol and metrics (Section~\ref{app:supp-metrics}), the expert annotation interfaces (Section~\ref{app:annotation-interface}), and the expert annotation and audit process (Section~\ref{app:annotation-process}).

\subsubsection{A-I-R-I capability ontology}
\label{app:capability-ontology}

AIRIS operationalizes social embodied intelligence as 36 separately scored
capabilities: 19 in Attention, three in Intention, four in Reaction, and ten in
Interaction (Tables~\ref{tab:attention-ontology}-\ref{tab:interaction-ontology}).
The decomposition is
intentionally diagnostic.  Attention asks what can be directly grounded in the
observed scene; Intention asks what latent goal is supported by that evidence;
Reaction asks what Ego should do at the current moment; and Interaction judges
how the exchange actually unfolded.  Thus, Interaction does not repeat the
partner-intent label.  Its labels describe episode phase, completion, mutual
understanding, and Ego's norm compliance.  These labels are adjudicated
post hoc using the complete episode, although every model query is still
restricted to the released causal prefix.  In particular, state and violation
polls use $O_{\leq\tau}$ at their query time, whereas resolved episode-level
judgments use $O_{\leq T_3}$.

Tables~\ref{tab:attention-ontology}-\ref{tab:interaction-ontology}
organize the ontology into the four consecutive stages of the A-I-R-I process.
For every capability, we state its operational meaning, the observation prefix
available to the model, and the concrete answer space.  

\begingroup
\small
\setlength{\tabcolsep}{3pt}
\renewcommand{\arraystretch}{1.16}

\paragraph{Stage 1: Attention.}
This stage establishes whom Ego should attend to and what directly observable
social cues that person exhibits.  It progresses from interactant localization
to cue-specific and integrated attentional grounding.

\begin{longtable}{@{}p{0.04\textwidth}p{0.20\textwidth}p{0.34\textwidth}p{0.12\textwidth}p{0.25\textwidth}@{}}
\caption{Attention capabilities In AIRIS.}\label{tab:attention-ontology}\\
\toprule
\textbf{ID} & \textbf{Capability} & \textbf{Meaning} & \textbf{Input prefix} & \textbf{Answer space} \\
\midrule
\endfirsthead
\multicolumn{5}{l}{\small\itshape Table~\ref{tab:attention-ontology} continued.}\\
\toprule
\textbf{ID} & \textbf{Capability} & \textbf{Meaning} & \textbf{Input prefix} & \textbf{Answer space} \\
\midrule
\endhead
\midrule
\multicolumn{5}{r}{\small\itshape Continued on next page.}\\
\endfoot
\bottomrule
\endlastfoot
A1 & Interactant Localization & Localizes the person who is socially interacting with Ego. This tests whether the model can isolate the relevant participant before reasoning about finer cues. & $O_{\leq\tau_A}$ & Single selection among candidate bounding boxes normalized to $[0,1000]$. \\
A2 & Interactant Identification and Localization & Jointly identifies the interactant by a short visual description and localizes that person. Both semantic identity and spatial grounding must be correct. & $O_{\leq\tau_A}$ & Single choice among person-description and bounding-box pairs. \\
A3 & Gaze Target Recognition & Identifies the object, person, or region indicated by the interactant's gaze or head orientation. It measures semantic gaze grounding. & $O_{\leq\tau_A}$ & Single choice among scene-conditioned target descriptions. \\
A4 & Gaze Target Localization & Localizes the target of the interactant's gaze on the current frame. This separates spatial grounding from target naming. & $O_{\leq\tau_A}$ & Single selection among candidate normalized bounding boxes. \\
A5 & Gaze Target Identification and Localization & Jointly identifies and localizes the gaze target. The selected option must contain both the correct target description and its location. & $O_{\leq\tau_A}$ & Single choice among target-description and bounding-box pairs. \\
A6 & Hand Gesture Recognition & Classifies the interactant's visible hand gesture or hand action. The label captures the form of the cue rather than its target. & $O_{\leq\tau_A}$ & Single choice from the hand-gesture ontology, such as pointing, waving, reaching, holding, or no salient gesture. \\
A7 & Hand Target Recognition & Identifies the object or person indicated, touched, or otherwise acted upon by the hand. It tests semantic grounding of manual cues. & $O_{\leq\tau_A}$ & Single choice among scene-conditioned target descriptions. \\
A8 & Hand Target Localization & Localizes the target of the observed hand gesture or action. The target is selected on the last visible frame. & $O_{\leq\tau_A}$ & Single selection among candidate normalized bounding boxes. \\
A9 & Hand Target Identification and Localization & Jointly identifies and localizes the target of the hand cue. It requires consistency between the semantic label and spatial region. & $O_{\leq\tau_A}$ & Single choice among target-description and bounding-box pairs. \\
A10 & Body Pose Recognition & Classifies the interactant's body orientation relative to Ego. This captures whether the person is facing, leaving, or turning toward the interaction. & $O_{\leq\tau_A}$ & Single choice among facing Ego, facing away, turning, and side-on orientation. \\
A11 & Body Stance Recognition & Classifies the interactant's visible body configuration. The label summarizes posture independently of the object toward which the body is oriented. & $O_{\leq\tau_A}$ & Single choice from the stance ontology, such as standing, seated, leaning, crouching, or arms crossed. \\
A12 & Body-orientation Target Recognition & Identifies what the interactant's torso or body is primarily oriented toward. It grounds body direction in a semantic target. & $O_{\leq\tau_A}$ & Single choice among scene-conditioned target descriptions. \\
A13 & Body-orientation Target Localization & Localizes the target toward which the interactant's body is oriented. It evaluates spatial rather than semantic target recovery. & $O_{\leq\tau_A}$ & Single selection among candidate normalized bounding boxes. \\
A14 & Body-orientation Target Identification and Localization & Jointly identifies and localizes the body-orientation target. The task requires agreement between target semantics and position. & $O_{\leq\tau_A}$ & Single choice among target-description and bounding-box pairs. \\
A15 & Emotion Recognition & Recognizes the affective state most visibly expressed by the interactant. Only states supported by observable facial or bodily evidence are considered. & $O_{\leq\tau_A}$ & Single choice from the affect ontology, such as neutral, happy, confused, concerned, or frustrated. \\
A16 & Overall Attention Target Recognition & Infers the interactant's dominant attentional target by integrating gaze, hand, body, and speech cues. It tests cross-cue attention resolution. & $O_{\leq\tau_A}$ & Single choice among scene-conditioned target descriptions. \\
A17 & Overall Attention Target Localization & Localizes the dominant attentional target after integrating all available cues. This is the spatial counterpart of A16. & $O_{\leq\tau_A}$ & Single selection among candidate normalized bounding boxes. \\
A18 & Overall Attention Target Identification and Localization & Jointly identifies and localizes the integrated attentional target. Both the semantic and spatial components must agree. & $O_{\leq\tau_A}$ & Single choice among target-description and bounding-box pairs. \\
A19 & Dominant Attention Modalities & Identifies which observable modalities provide the primary evidence for the current attentional focus. Multiple cues may jointly support the judgment. & $O_{\leq\tau_A}$ & Multiple selection over gaze, hand movement, body posture, and speech. \\
\end{longtable}

\paragraph{Stage 2: Intention.}
This stage moves from observable evidence to the partner's latent goal.  Its
temporal questions explicitly distinguish justified inference after $T_1$ from
premature commitment before sufficient evidence is available.

\begin{longtable}{@{}p{0.04\textwidth}p{0.20\textwidth}p{0.34\textwidth}p{0.12\textwidth}p{0.25\textwidth}@{}}
\caption{Intention capabilities in AIRIS.}\label{tab:intention-ontology}\\
\toprule
\textbf{ID} & \textbf{Capability} & \textbf{Meaning} & \textbf{Input prefix} & \textbf{Answer space} \\
\midrule
\endfirsthead
\multicolumn{5}{l}{\small\itshape Table~\ref{tab:intention-ontology} continued.}\\
\toprule
\textbf{ID} & \textbf{Capability} & \textbf{Meaning} & \textbf{Input prefix} & \textbf{Answer space} \\
\midrule
\endhead
\bottomrule
\endlastfoot
I1 & Key-moment Intention Recognition & Recovers the partner's latent social goal at an annotated key moment. The prediction concerns what the partner seeks to achieve, not merely the visible action. & $O_{\leq\tau_k}$ & Single choice from the intent ontology, including information seeking, guidance request, object request, action request, social conversation, and other goals. \\
I2 & Temporal Intention Recognition & Determines whether the partner's intent is already supportable at each sampled time and, when it is, recovers that intent. Before $T_1$, calibrated waiting is represented explicitly. & $O_{\leq\tau}$ sampled around $T_1$ & Single choice from the intent ontology plus ``Insufficient evidence to determine the intention.'' \\
I3 & Intention Evidence Grounding & Identifies the observable modalities that justify the stated intent inference. This tests whether a correct intent is supported by the appropriate evidence. & $O_{\leq\tau_k}$ with $\tau_k\geq T_1$ & Multiple selection over gaze, hand movement, body orientation, and speech. \\
\end{longtable}

\paragraph{Stage 3: Reaction.}
This stage converts the inferred social situation into an Ego-centered decision.
It first asks whether intervention is warranted and then specifies the response
at increasingly concrete levels.

\begin{longtable}{@{}p{0.04\textwidth}p{0.20\textwidth}p{0.34\textwidth}p{0.12\textwidth}p{0.25\textwidth}@{}}
\caption{Reaction capabilities in AIRIS.}\label{tab:reaction-ontology}\\
\toprule
\textbf{ID} & \textbf{Capability} & \textbf{Meaning} & \textbf{Input prefix} & \textbf{Answer space} \\
\midrule
\endfirsthead
\multicolumn{5}{l}{\small\itshape Table~\ref{tab:reaction-ontology} continued.}\\
\toprule
\textbf{ID} & \textbf{Capability} & \textbf{Meaning} & \textbf{Input prefix} & \textbf{Answer space} \\
\midrule
\endhead
\bottomrule
\endlastfoot
R1 & Engagement Decision & Decides whether Ego should actively respond or refrain from intervention at the current moment. It directly tests justified action versus justified inaction. & $O_{\leq\tau_k}$ & Single choice between active response and refrain from intervention. \\
R2 & Action Domain Selection & Selects the broad response modalities that Ego should employ. More than one domain may be required for a coordinated response. & $O_{\leq\tau_k}$ & Multiple selection over navigation, object manipulation, verbal communication, and idle/non-intervention. \\
R3 & Sub-action Selection & Selects the concrete actions that operationalize Ego's response. The labels distinguish, for example, approaching, yielding, grasping, speaking, observing, and taking no action. & $O_{\leq\tau_k}$ & Multiple selection from the embodied sub-action ontology, with an additional ``Other action'' option. \\
R4 & Speech Function Selection & Determines the communicative function of Ego's response when speech is warranted. It evaluates pragmatic response selection rather than free-form generation. & $O_{\leq\tau_k}$ & Single choice among greeting, confirmation, clarification question, information response, recommendation, warning, farewell, and empathic response. \\
\end{longtable}

\paragraph{Stage 4: Interaction.}
This stage evaluates the progression and consequences of the exchange.  Online
state and violation probes use the current prefix, while resolved episode-level
attributes assess completion, mutual understanding, and Ego's social compliance
from the prefix ending at $T_3$.

\begin{longtable}{@{}p{0.04\textwidth}p{0.20\textwidth}p{0.34\textwidth}p{0.12\textwidth}p{0.25\textwidth}@{}}
\caption{Interaction capabilities in AIRIS.}\label{tab:interaction-ontology}\\
\toprule
\textbf{ID} & \textbf{Capability} & \textbf{Meaning} & \textbf{Input prefix} & \textbf{Answer space} \\
\midrule
\endfirsthead
\multicolumn{5}{l}{\small\itshape Table~\ref{tab:interaction-ontology} continued.}\\
\toprule
\textbf{ID} & \textbf{Capability} & \textbf{Meaning} & \textbf{Input prefix} & \textbf{Answer space} \\
\midrule
\endhead
\midrule
\multicolumn{5}{r}{\small\itshape Continued on next page.}\\
\endfoot
\bottomrule
\endlastfoot
N1 & Interaction State Recognition & Tracks the current phase of the exchange rather than inferring the partner's goal. It indicates whether social engagement is beginning, active, ending, or absent. & $O_{\leq\tau}$ & Single choice among pre-engagement, active engagement, disengagement, and no social engagement. \\
N2 & Online Violation Detection & Detects whether Ego is exhibiting a social violation at the current time. This provides an online binary compliance signal before episode resolution. & $O_{\leq\tau}$ & Binary choice between violation present and no violation. \\
N3 & Online Violation Type Recognition & Identifies the type or types of Ego violation visible at the current time. Multiple simultaneous violations can be selected. & $O_{\leq\tau}$ & Multiple selection from the spatial, navigation, manipulation, physical-contact, and communication violation ontology. \\
N4 & Episode-level Violation Recognition & Summarizes all Ego violation types that occurred anywhere in the resolved episode. It aggregates event-level compliance evidence without re-inferring partner intent. & $O_{\leq T_3}$ & Multiple selection from the complete violation ontology, including ``No violation.'' \\
N5 & Interaction Completeness & Judges whether the exchange formed a complete social and task-level loop. Incomplete episodes distinguish external interruption from unilateral abandonment. & $O_{\leq T_3}$ & Single choice among completed, interrupted, and abandoned. \\
N6 & Mutual Understanding & Judges whether Ego and the interactant reached a shared understanding of the request, response, or closure. Completion alone does not guarantee mutual understanding. & $O_{\leq T_3}$ & Binary choice between mutual understanding achieved and not achieved. \\
N7 & Spatial Compliance & Assesses whether Ego's approach, stopping distance, route, and use of shared space were socially compliant. If not, it identifies the principal spatial violation. & $O_{\leq T_3}$ & Single choice between compliant behavior and one applicable spatial or navigation violation category. \\
N8 & Physical Compliance & Assesses whether Ego's manipulation and physical-contact behavior was safe, controlled, and respectful of bodily boundaries. If not, it identifies the principal physical violation. & $O_{\leq T_3}$ & Single choice between compliant behavior and one applicable manipulation or physical-contact violation category. \\
N9 & Communication Compliance & Assesses conversational and non-verbal etiquette across the exchange. It can capture several communication violations when they co-occur. & $O_{\leq T_3}$ & Multiple selection from the communication-violation ontology, including interrupting, prolonged staring, impolite tone, ignored queries, and missing acknowledgement. \\
N10 & Overall Compliance Severity & Assigns the overall severity of Ego's social compliance after considering the complete exchange. This is an outcome-level judgment rather than an intent label. & $O_{\leq T_3}$ & Single choice among fully compliant, partial violation, and serious violation. \\
\end{longtable}
\endgroup

For evaluation, single-choice, binary, and candidate-box questions use
option-key accuracy.  Multi-select questions use set F1 as the primary score
and additionally report exact match.  For Temporal Intention Recognition, the
event-level Intent Anticipation Score rewards the earliest stable correct prediction after
$T_1$ and discounts it by the false-commitment rate before $T_1$.  For the six
resolved Interaction attributes N5-N10, we additionally report the mean
per-segment component score and joint exact match when all six labels are
available.

\subsubsection{Supplementary Evaluation Protocol and Metrics}
\label{app:supp-metrics}

Building upon the core evaluation protocol in Section 2.3 of the main paper, this section 
defines the complete set of metrics used to isolate fine-grained reasoning bottlenecks across the 36 capabilities of AIRIS. These metrics are reported in the detailed evaluation tables (Section~\ref{app:extended-results}) to prevent surface-level accuracy from masking deeper failure modes.

\textbf{Scoring and aggregation.}
Base items are scored by exact match (single-choice, yes/no, bounding-box) or set F1 (multi-choice). Empty, unparseable, and out-of-vocabulary responses score 0. The headline metric is the unweighted mean of the four question-micro layer scores: $\mathrm{Overall} = \tfrac{1}{4}(S_{\mathrm{Att}}+S_{\mathrm{Int}}+S_{\mathrm{Rea}}+S_{\mathrm{Inter}})$.

\textbf{Attention Metrics:}
\begin{itemize}[leftmargin=1.5em,topsep=2pt,itemsep=2pt]
    \item \emph{Semantic Target \& Spatial Localization:} \emph{Semantic target} assesses which entity a social cue (e.g., gaze, pointing) refers to. \emph{Spatial localization} evaluates its grounding over candidate boxes. 
    \item \emph{Joint Identity-Localization:} A strict metric requiring both the semantic class and the spatial bounding box to match the reference simultaneously. If either fails, the joint score is 0. This penalizes models that name a referent without localizing it, yielding a non-executable percept.
    \item \emph{Nonverbal States:} Accuracy on classifying body pose, physical stance, hand gesture, and observable emotion.
    \item \emph{Multimodal Attribution:} Set F1 score for identifying which observable modalities (gaze, hand movement, body posture, speech) provide the primary evidence for the current attentional focus.
\end{itemize}

\textbf{Intention Metrics:}
\begin{itemize}[leftmargin=1.5em,topsep=2pt,itemsep=2pt]
    \item \emph{Intent Anticipation Score (IAS):} Evaluates streamed intent recovery. Before the earliest supportable moment $T_1$, ``insufficient evidence'' is the correct label. Let $\mathrm{FCR}$ be the False Commitment Rate (the fraction of pre-$T_1$ queries prematurely committing to a specific intent). For models reaching stable correctness at timestamp $p \ge T_1$, we define an anticipation margin $m = \mathrm{clip}((T_2-p)/(T_2-t_f), 0, 1)$, where $t_f$ is the first query at or after $T_1$ (with $m=0$ if stability is never reached). The event-level score is $\mathrm{IAS}_{\mathrm{event}} = m \cdot (1 - \mathrm{FCR})$, and IAS is its macro average.
    \item \emph{Key-Moment \& Online Accuracy:} \emph{Key-moment} evaluates accuracy at fixed expert-annotated query timestamps. \emph{Online} evaluates accuracy across sampled temporal prefixes.
    \item \emph{Evidence Grounding:} Set F1 score for identifying which specific observable cues (visual, vocal, or verbal) justify the inferred intent.
\end{itemize}

\textbf{Reaction Metrics:}
\begin{itemize}[leftmargin=1.5em,topsep=2pt,itemsep=2pt]
    \item \emph{Engage:} Macro-F1 between active intervention and idle/waiting classes. Chosen over plain accuracy to penalize collapsed policies on heavily imbalanced classes.
    \item \emph{Action Domain \& Subaction:} \emph{Action domain} and \emph{Subaction} use multi-label F1 to evaluate the broad response modalities and concrete actions Ego should employ. \emph{Speak subtype} uses accuracy for communicative functions.
    \item \emph{Joint-EM:} A strict diagnostic for responses requiring the engagement, action domain, subaction, and any annotated speech subtype to match the reference simultaneously.
    \item \emph{Domain-Specific Diagnostics:} Breaks down the Subaction + Speak scores by the required action domain (Communication, Navigation, Manipulation, Refrain/Idle) to isolate modality-specific performance.
\end{itemize}

\textbf{Interaction Metrics:}
\begin{itemize}[leftmargin=1.5em,topsep=2pt,itemsep=2pt]
    \item \emph{State \& Task Outcome:} \emph{State} tracks the current phase of the exchange (e.g., active engagement, disengagement). \emph{Task outcome} judges whether the underlying social exchange succeeded, failed, or was aborted, and whether mutual understanding was achieved.
    \item \emph{Compliance:} Equally averages spatial etiquette, physical demeanor, communication etiquette, and overall compliance category ($\tau \ge T_3$).
    \item \emph{Violation:} Equally averages online violation-presence F1, online violation type, and segment-level violation types. ``Violation present'' is the positive class to prevent exploitation of the skewed negative baseline.
\end{itemize}

\subsubsection{Expert Annotation Interfaces}
\label{app:annotation-interface}

\begin{figure}[!ht]
    \centering
    \begin{subfigure}[b]{0.48\textwidth}
        \centering
        \includegraphics[width=\textwidth]{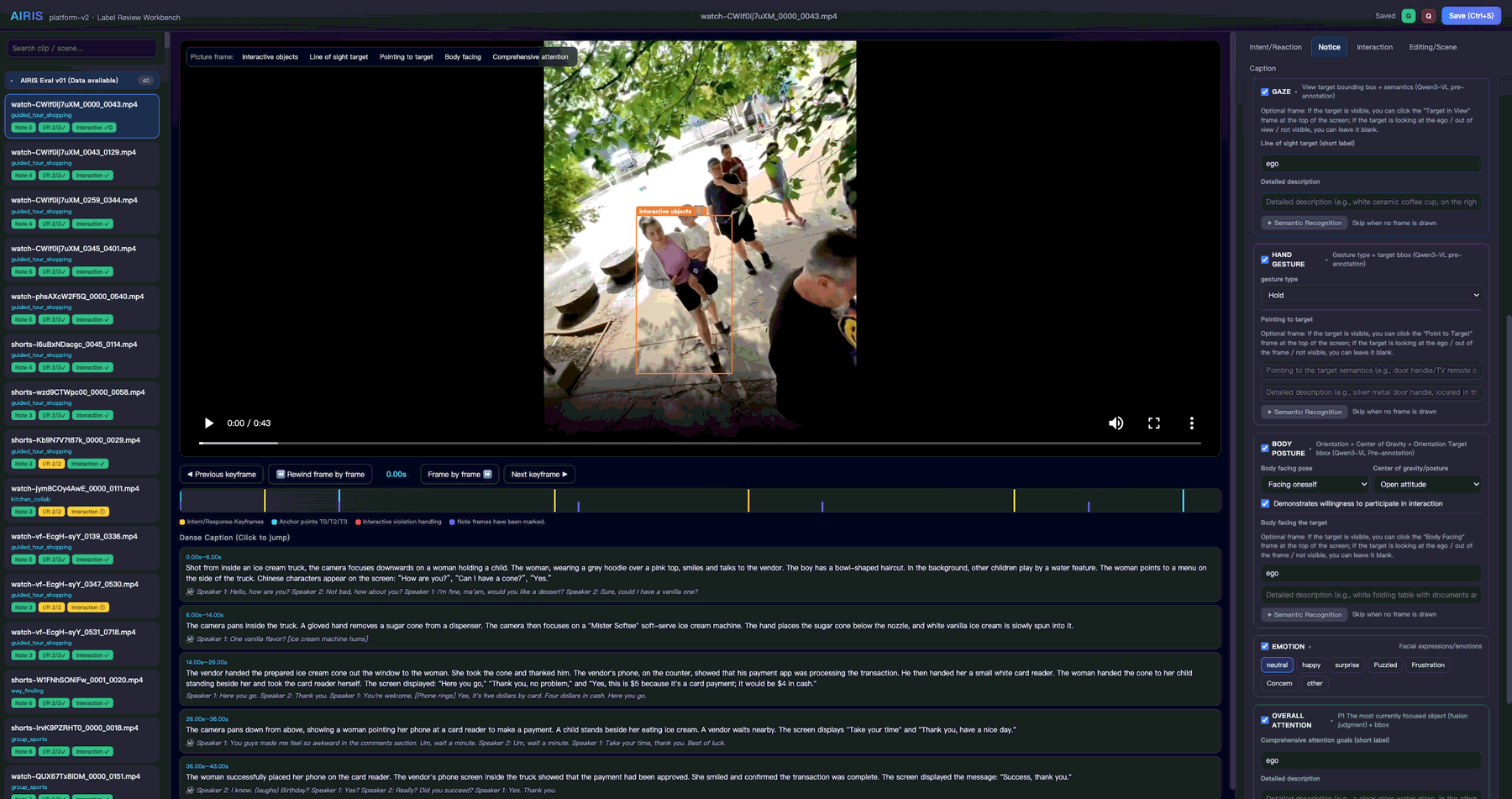}
        \caption{Structured A-I-R-I annotation interface.}
    \end{subfigure}
    \hfill
    \begin{subfigure}[b]{0.48\textwidth}
        \centering
        \includegraphics[width=\textwidth]{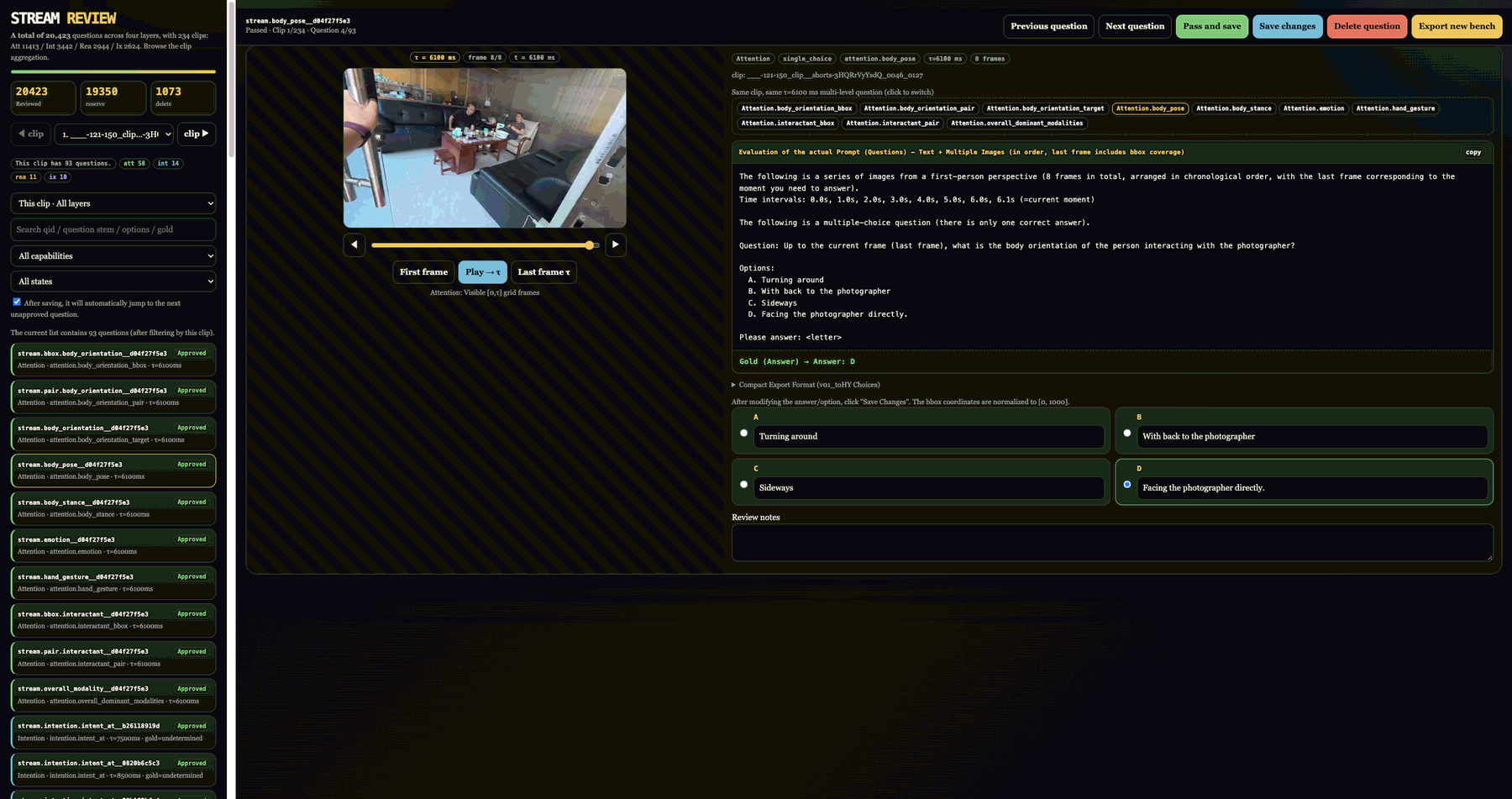}
        \caption{Question review and audit interface.}
    \end{subfigure}
    \caption{AIRIS annotation interfaces used by human experts during the data annotation process.}
    \label{fig:annotation-interfaces}
\end{figure}

As shown in Figure~\ref{fig:annotation-interfaces}, we developed two specialized web interfaces to facilitate the AIRIS annotation and auditing workflow. The first interface (Figure~\ref{fig:annotation-interfaces}a) is designed for structured A-I-R-I annotation: it integrates egocentric video playback, spatial bounding-box annotation, time-aligned dense captions, and structured forms for recording observable attention cues, temporal anchors ($T_0$ to $T_3$), partner intentions, ego-agent reactions, and interaction attributes. The second interface (Figure~\ref{fig:annotation-interfaces}b) is dedicated to question review and auditing, pairing each candidate question and its admissible video prefix with multiple-choice options and review controls for item-level inspection, editing, approval, or removal.

\subsubsection{Expert Annotation and Audit Process}
\label{app:annotation-process}

To ensure the high quality and reliability of the AIRIS benchmark, dataset construction relies on a rigorous human-in-the-loop annotation and verification protocol. A total of 11 human experts, who underwent multiple rounds of training and had their capabilities rigorously verified, participated in the annotation and review process; all participating experts received fair monetary compensation for their work.

Using the two interfaces above, the expert workflow proceeds in two stages. During structured A-I-R-I annotation, annotators review the video stream, identify and refine critical temporal anchors ($T_0$ to $T_3$), manually ground observable attention cues, and adjudicate structured labels for the partner's intention, the ego-agent's reaction, and episode-level interaction compliance strictly based on observable evidence. During question auditing, experts inspect the generated question-answer pairs, distractors, and temporal constraints item by item—retaining valid items, revising correctable ones, and removing flawed candidates—to ensure that every released evaluation item is unambiguous, uniquely answerable from the causal prefix, and free from future-information leakage.

\subsection{Supplementary Experiments and Results}
\label{app:experimental-details}

\subsubsection{Model evaluation pool}
\label{app:evaluated-models}

AIRIS targets a broad model pool spanning proprietary multimodal APIs,
general-purpose open-weight models, streaming video-language models, and
embodied foundation models.  The pool below contains 45 endpoints or
checkpoints; the main result table reports only models with completed,
validated inference records.  Superscript footnotes after each model name give
the official provider documentation for proprietary systems and an official
upstream Hugging Face repository, or the maintained conversion corresponding
to the evaluated checkpoint, for open-weight systems.  All links were checked
on September 25, 2026.  The descriptions distinguish native video or audio
support from image-only interfaces, for which temporally ordered frames are
required.  Model names used by the evaluation harness are retained when they
differ from the canonical public identifier.

\newcommand{\airismodelurl}[1]{\footnote{\url{#1}}}

\paragraph{Proprietary models.}
These models are accessed through provider APIs and do not have public
Hugging Face weights.
\begin{itemize}[leftmargin=*,itemsep=2pt,topsep=3pt]
    \item \textbf{Gemini 3.7 Flash}\airismodelurl{https://ai.google.dev/gemini-api/docs/models/gemini-3.7-flash} \citep{googledeepmind2026gemini37flash} is Google's natively multimodal, efficiency-oriented
    reasoning model.  Its public API accepts text, image, video, audio, and PDF
    inputs with a 1,048,576-token input limit.

    \item \textbf{Gemini 3.1 Pro}\airismodelurl{https://ai.google.dev/gemini-api/docs/models/gemini-3.1-pro-preview} \citep{googledeepmind2026gemini31pro}
    is the higher-compute Gemini endpoint for complex multimodal reasoning over
    text, images, video, audio, and PDFs.  

    \item \textbf{GPT-5.6 Sol}\airismodelurl{https://developers.openai.com/api/docs/models/gpt-5.6-sol} \citep{openai2026gpt56sol} is OpenAI's flagship GPT-5.6 API model, with a 1,050,000-token context window.  Its public API supports text and image input but not audio or video input, so video evaluation requires an external frame representation.

    \item \textbf{Doubao-Seed-2.0-Lite}\airismodelurl{https://seed.bytedance.com/zh/seed2} \citep{bytedanceseed2026seed20lite} is ByteDance's efficiency-oriented proprietary model for high-frequency production workloads.  The updated endpoint natively unifies text, image, video, and audio understanding, including joint audiovisual reasoning and temporal event localization, with a 256K-token context window.
\end{itemize}

\paragraph{General-purpose open-weight models.}
This group covers native vision-language checkpoints as well as image-language
models evaluated through ordered video frames.
\begin{itemize}[leftmargin=*,itemsep=2pt,topsep=3pt]
    \item \textbf{Qwen3.6-35B-A3B}\airismodelurl{https://huggingface.co/Qwen/Qwen3.6-35B-A3B} \citep{qwenteam2026qwen3635ba3b}
    is a native image-and-video model with 35B total parameters and 3B activated
    parameters per token.  It combines a vision encoder with a hybrid
    Gated-DeltaNet and attention MoE language backbone.

    \item \textbf{Gemma-4-31B-it}\airismodelurl{https://huggingface.co/google/gemma-4-31B-it} \citep{gemmateam2026gemma4}
    is Google's 30.7B dense instruction-tuned model with text and image inputs.
    Its official processor handles video as a sequence of frames, while this
    31B variant does not include the audio encoder available in smaller Gemma 4
    variants.

    \item \textbf{DeepSeek-V4.1-Flash}\airismodelurl{https://huggingface.co/deepseek-ai/DeepSeek-V4.1-Flash} \citep{deepseekai2026deepseekv41flash}
    is a 552B-backbone multimodal MoE with native text-and-image input, no
    documented native video route, and a context window of up to one million
    tokens. 

    \item \textbf{Qwen3.6-27B}\airismodelurl{https://huggingface.co/Qwen/Qwen3.6-27B} \citep{qwenteam2026qwen3627b}
    is the dense 27B member of the Qwen3.6 family.  It natively accepts images
    and videos and provides a 262,144-token context window that can be extended
    beyond one million tokens.

    \item \textbf{Qwen3.5-35B-A3B}\airismodelurl{https://huggingface.co/Qwen/Qwen3.5-35B-A3B} \citep{qwenteam2026qwen35}
    is a 35B-total, 3B-active MoE trained as a unified vision-language model.
    Its early-fusion interface supports both image and video inputs.

    \item \textbf{Qwen3-VL-32B-Instruct}\airismodelurl{https://huggingface.co/Qwen/Qwen3-VL-32B-Instruct} \citep{bai2025qwen3vl}
    is an approximately 33B instruction-tuned VLM designed for spatial and
    long-video reasoning.  It uses a native 256K context and timestamp-aware
    video positional encoding.

    \item \textbf{Qwen3.5-27B}\airismodelurl{https://huggingface.co/Qwen/Qwen3.5-27B} \citep{qwenteam2026qwen35}
    is the dense 27B Qwen3.5 checkpoint with a vision encoder.  It shares the
    family's unified image-and-video interface and long-context support.

    \item \textbf{Qwen3.8-27B}\airismodelurl{https://huggingface.co/Qwen/Qwen3.8-27B} \citep{qwenteam2026qwen3827b}
    is a dense 27B native vision-language model derived from the Qwen3.5
    architecture.  Its public checkpoint supports images, videos, and
    controllable reasoning.

    \item \textbf{Qwen2.5-VL-7B-Instruct}\airismodelurl{https://huggingface.co/Qwen/Qwen2.5-VL-7B-Instruct} \citep{bai2025qwen25vl}
    is an instruction-tuned VLM with dynamic image resolution and dynamic
    frame-rate sampling.  Absolute-time alignment in its multimodal positional
    encoding supports temporal localization in video.

    \item \textbf{InternVL3-8B}\airismodelurl{https://huggingface.co/OpenGVLab/InternVL3-8B} \citep{zhu2025internvl3}
    is an 8B-class model trained with interleaved text, image, and video data.
    Its video interface samples frames and packs them as a multi-image sequence
    with frame identifiers.

    \item \textbf{InternVL3.5-30B-A3B}\airismodelurl{https://huggingface.co/OpenGVLab/InternVL3_5-30B-A3B} \citep{wang2025internvl35}
    combines a 0.3B vision component with a 30.5B MoE language component, for
    30.8B total parameters and roughly 3B active parameters.  Video is processed
    as a sampled sequence of dynamically tiled frames.

    \item \textbf{InternVL3.5-8B}\airismodelurl{https://huggingface.co/OpenGVLab/InternVL3_5-8B} \citep{wang2025internvl35}
    combines a 0.3B vision encoder with an 8.2B language model.  It supports
    image, multi-image, and sampled-frame video conversations.

    \item \textbf{Qwen2.5-VL-3B-Instruct}\airismodelurl{https://huggingface.co/Qwen/Qwen2.5-VL-3B-Instruct} \citep{bai2025qwen25vl}
    is the smaller Qwen2.5-VL checkpoint, retaining dynamic resolution, visual
    grounding, and variable-frame-rate video input.  It provides a
    resource-efficient comparison to the 7B variant.

    \item \textbf{LLaVA-OneVision-Qwen2-7B}\airismodelurl{https://huggingface.co/llava-hf/llava-onevision-qwen2-7b-ov-hf} \citep{li2025llavaonevision}
    couples a Qwen2 language model with the SigLIP SO400M vision encoder.  It is
    trained for single-image, multi-image, and video scenarios and exposes all
    three through one checkpoint.

    \item \textbf{Qwen3.5-4B}\airismodelurl{https://huggingface.co/Qwen/Qwen3.5-4B} \citep{qwenteam2026qwen35}
    is a native vision-language model with a 4B language backbone and image and
    video input.  It offers the same hybrid linear-attention and full-attention
    design as larger Qwen3.5 variants at a smaller scale.

    \item \textbf{Qwen3.5-2B}\airismodelurl{https://huggingface.co/Qwen/Qwen3.5-2B} \citep{qwenteam2026qwen35}
    has a 2B language backbone and preserves the Qwen3.5 family's unified
    image-and-video interface for low-compute evaluation.

    \item \textbf{Llama-3.2-11B-Vision-Instruct}\airismodelurl{https://huggingface.co/meta-llama/Llama-3.2-11B-Vision-Instruct} \citep{meta2024llama32vision}
    is Meta's 10.6B-parameter instruction-tuned image-reasoning model with
    text-and-image input and text output.  It has no native video interface, so
    videos must be represented through extracted images.

    \item \textbf{Qwen3.5-0.8B}\airismodelurl{https://huggingface.co/Qwen/Qwen3.5-0.8B} \citep{qwenteam2026qwen35}
    is the smallest Qwen3.5 checkpoint in the pool, with a 0.8B language
    backbone.  It retains the family's native image and video input interface.

    \item \textbf{GLM-5.3-Flash}\airismodelurl{https://huggingface.co/zai-org/GLM-5.3-Flash} \citep{zai2026glm53flash}
    is a native multimodal MoE with 320B total parameters and 18B activated
    parameters.  Its hybrid sparse- and linear-attention architecture supports
    text, image, video, and file inputs with text output; the provider does not
    document native audio input.

\end{itemize}

\paragraph{Streaming video-language models.}
These checkpoints are designed for online or long-running video rather than
only offline clip-level question answering.
\begin{itemize}[leftmargin=*,itemsep=2pt,topsep=3pt]
    \item \textbf{JoyAI-VL-Interaction-Preview}\airismodelurl{https://huggingface.co/jdopensource/JoyAI-VL-Interaction-Preview} \citep{yao2026joyaivlinteraction}
    is an 8B-scale, vision-first interaction model that observes a continuing
    video stream.  At each second it learns to speak, remain silent, or delegate
    a difficult request to a background model.

    \item \textbf{AURA-8B}\airismodelurl{https://huggingface.co/aurateam/AURA} \citep{lu2026aura}
    is a Qwen3-VL-8B-based model for always-on video understanding and proactive
    assistance.  The released system processes a continuous 2-FPS stream and
    combines the VLM with pluggable ASR and streaming TTS.

    \item \textbf{StreamingVLM}\airismodelurl{https://huggingface.co/mit-han-lab/StreamingVLM} \citep{xu2026streamingvlm}
    is an 8B-class model fine-tuned from Qwen2.5-VL for effectively unbounded
    video streams.  It bounds inference cost by retaining attention sinks,
    recent visual tokens, and a longer window of text tokens in a compact KV
    cache.

    \item \textbf{Flash-VStream-Qwen-7B}\airismodelurl{https://huggingface.co/zhang9302002/Flash-VStream-Qwen-7b} \citep{zhang2025flashvstream}
    is a Qwen-based video-language checkpoint for real-time understanding and
    question answering over very long streams.  Its Flash Memory mechanism
    compresses historical visual information while new frames arrive.

    \item \textbf{VITA-1.5}\airismodelurl{https://huggingface.co/VITA-MLLM/VITA-1.5} \citep{fu2025vita15}
    is an 8B video-language model developed for real-time vision and speech
    interaction.  The public release includes the main checkpoint together with
    audio-encoder and speech-generation components.
\end{itemize}

\paragraph{Omni-modal models.}
These checkpoints expose native audio-visual interfaces or otherwise combine
language, vision, and speech in a single model family.
\begin{itemize}[leftmargin=*,itemsep=2pt,topsep=3pt]
    \item \textbf{Qwen2.5-Omni-7B}\airismodelurl{https://huggingface.co/Qwen/Qwen2.5-Omni-7B} \citep{xu2025qwen25omni}
    is an end-to-end omni-modal model that accepts text, images, audio, and video
    and can generate text or natural speech.  Its time-aligned multimodal
    positional encoding synchronizes video and audio streams.

    \item \textbf{Qwen3-Omni-30B-A3B-Instruct}\airismodelurl{https://huggingface.co/Qwen/Qwen3-Omni-30B-A3B-Instruct} \citep{xu2025qwen3omni}
    combines a 30B-total, 3B-active MoE Thinker with a 3B-total, 0.3B-active
    MoE Talker.  It accepts text, images, audio, and video and produces text and
    streaming natural-speech outputs.

    \item \textbf{OmniVinci}\airismodelurl{https://huggingface.co/nvidia/omnivinci} \citep{ye2026omnivinci}
    is NVIDIA's omni-modal model for joint visual, audio, and language
    understanding.  Official inference code accepts images, standalone audio,
    and video with its audio track and decodes textual responses.

    \item \textbf{Qwen2.5-Omni-3B}\airismodelurl{https://huggingface.co/Qwen/Qwen2.5-Omni-3B} \citep{xu2025qwen25omni}
    is the compact Qwen2.5-Omni checkpoint.  It retains text, image, audio, and
    video perception together with text and speech generation.

    \item \textbf{Phi-4-Multimodal-Instruct}\airismodelurl{https://huggingface.co/microsoft/Phi-4-multimodal-instruct} \citep{microsoft2025phi4multimodal}
    is Microsoft's 5.6B model for text, image, and audio input with text output.
    Its documentation treats video as a sequence of images rather than a native
    modality.
\end{itemize}

\paragraph{Embodied foundation models.}
This group specializes in physical-world perception, spatial grounding,
planning, affordance reasoning, or robot-oriented interaction.
\begin{itemize}[leftmargin=*,itemsep=2pt,topsep=3pt]
    \item \textbf{HY-Embodied-VLM-1.0}\airismodelurl{https://huggingface.co/tencent/Hy-Embodied-VLM-1.0} \citep{tencent2026hyembodiedvlm10}
    is Tencent's approximately 30B-total MoE for physical-world agents, with
    roughly 3B parameters activated per token.  It combines a HY3-A3B language
    backbone with the HY-ViT2 vision encoder for embodied perception and
    action-centric reasoning.

    \item \textbf{RynnBrain1.1-9B}\airismodelurl{https://huggingface.co/Alibaba-DAMO-Academy/RynnBrain1.1-9B} \citep{li2026rynnbrain11}
    is the dense 9B member of the RynnBrain 1.1 family for embodied perception,
    localization, reasoning, and planning.  This generation adds native 3D-box
    grounding and instruction-conditioned contact-point prediction.

    \item \textbf{RynnBrain-30B-A3B}\airismodelurl{https://huggingface.co/Alibaba-DAMO-Academy/RynnBrain-30B-A3B} \citep{dang2026rynnbrain}
    is a Qwen3-VL-based embodied MoE whose model designation denotes a
    30B-class backbone with approximately 3B active parameters.  It targets
    egocentric understanding, spatial-temporal localization, physically
    grounded reasoning, and planning.

    \item \textbf{VeBrain}\airismodelurl{https://huggingface.co/OpenGVLab/VeBrain} \citep{luo2025vebrain}
    is the post-training checkpoint associated with the
    Visual Embodied Brain\airismodelurl{https://huggingface.co/papers/2506.00123}
    framework.  The framework unifies multimodal understanding, visual-spatial
    reasoning, and text-form control signals for downstream robot policies.

    \item \textbf{Pelican1.0-VL-7B}\airismodelurl{https://huggingface.co/X-Humanoid/Pelican1.0-VL-7B} \citep{zhang2025pelicanvl}
    is a 7B embodied multimodal foundation model trained for visual-text
    reasoning and spatiotemporal understanding.  Its stated use cases include
    grasping, navigation, action planning, and human-robot interaction.

    \item \textbf{MiMo-Embodied-7B}\airismodelurl{https://huggingface.co/XiaomiMiMo/MiMo-Embodied-7B} \citep{hao2025mimoembodied}
    is Xiaomi's cross-embodied vision-language model spanning embodied AI and
    autonomous-driving tasks.  Its released checkpoint targets task planning,
    affordance prediction, spatial understanding, environmental perception,
    state prediction, and driving planning.

    \item \textbf{RynnBrain1.1-2B}\airismodelurl{https://huggingface.co/Alibaba-DAMO-Academy/RynnBrain1.1-2B} \citep{li2026rynnbrain11}
    is the compact 2B RynnBrain 1.1 checkpoint.  It retains the family's 2D and
    3D grounding, contact-point prediction, embodied reasoning, and planning
    tasks.

    \item \textbf{RoboBrain2.0-7B}\airismodelurl{https://huggingface.co/BAAI/RoboBrain2.0-7B} \citep{baaiteam2025robobrain2}
    is BAAI's 7B embodied VLM that unifies perception, reasoning, and planning.
    It supports spatial outputs such as points, boxes, affordances, and
    trajectories together with temporal decision tasks.

    \item \textbf{RoboBrain2.0-32B}\airismodelurl{https://huggingface.co/BAAI/RoboBrain2.0-32B} \citep{baaiteam2025robobrain2}
    is the full-scale 32B RoboBrain 2.0 checkpoint.  It uses the same embodied
    interface as the 7B model while allocating greater capacity to spatial
    reasoning, scene memory, and long-horizon planning.

    \item \textbf{Embodied-R1-3B-v1}\airismodelurl{https://huggingface.co/IffYuan/Embodied-R1-3B-v1} \citep{yuan2026embodiedr1}
    is a 3B VLM for general robotic manipulation.  It combines a
    representation based on visual pointing with reinforced fine-tuning to
    connect perception, grounding, and action.

    \item \textbf{HY-Embodied-0.5}\airismodelurl{https://huggingface.co/tencent/HY-Embodied-0.5} \citep{tencent2026hyembodied05}
    is Tencent's embodied foundation-model suite for spatial-temporal perception
    and prediction, interaction, and planning.  The released MoT-2B checkpoint
    contains about 4B total parameters while activating approximately 2.2B at
    inference.

    \item \textbf{HY-Embodied-0.5-X}\airismodelurl{https://huggingface.co/tencent/HY-Embodied-0.5-X} \citep{tencent2026hyembodied05x}
    is an embodied-focused enhancement of the HY-Embodied-0.5 MoT-2B
    architecture.  It emphasizes fine-grained manipulation, spatial grounding,
    risk assessment, failure reflection, and long-horizon task planning.
\end{itemize}

\clearpage
\subsubsection{Complete quantitative results}
\label{app:extended-results}

Tables~\ref{tab:supp-attention-results}-\ref{tab:supp-interaction-results} report the stage-wise diagnostics for all 45 evaluated models. Models are grouped as in the main-paper table and ordered by global Overall within each group; this shared order keeps rows aligned across tables. 
Bold marks the best value in each column, including ties; all metrics are higher-is-better except FCR.

\begingroup
\scriptsize
\setlength{\tabcolsep}{2.0pt}
\renewcommand{\arraystretch}{1.08}
\setlength{\LTpre}{0pt}
\setlength{\LTpost}{3pt}
\begin{longtable}{@{}>{\raggedright\arraybackslash}p{0.30\textwidth}>{\columncolor{tableattention}}cccccc@{}}
\caption{Complete Attention results.}\label{tab:supp-attention-results}\\
\toprule
\textbf{Model} & \textbf{All} & \makecell{Nonverbal\\state} & \makecell{Semantic\\target} & \makecell{Spatial\\localization} & \makecell{Joint ID-\\localization} & \makecell{Multimodal\\attribution} \\
\midrule
\endfirsthead
\multicolumn{7}{l}{\small\itshape Table~\ref{tab:supp-attention-results} continued.}\\
\toprule
\textbf{Model} & \textbf{All} & \makecell{Nonverbal\\state} & \makecell{Semantic\\target} & \makecell{Spatial\\localization} & \makecell{Joint ID-\\localization} & \makecell{Multimodal\\attribution} \\
\midrule
\endhead
\midrule
\multicolumn{7}{r}{\small\itshape Continued on next page.}\\
\endfoot
\bottomrule
\endlastfoot
    \multicolumn{7}{l}{\cellcolor{black!4}\textit{Closed-source models}} \\
    Gemini-3.7-Flash & \textbf{0.936} & 0.901 & \textbf{0.951} & \textbf{0.911} & \textbf{0.985} & 0.873 \\
    Gemini-3.1-Pro & 0.851 & 0.859 & 0.897 & 0.718 & 0.921 & \textbf{0.912} \\
    GPT-5.6-Sol & 0.906 & \textbf{0.926} & 0.925 & 0.870 & 0.899 & 0.859 \\
    Doubao-seed-2.0-lite & 0.759 & 0.707 & 0.859 & 0.742 & 0.742 & 0.799 \\
    \midrule
    \multicolumn{7}{l}{\cellcolor{black!4}\textit{Open-source general-purpose models}} \\
    Qwen3.6-35B-A3B & 0.723 & 0.714 & 0.891 & 0.560 & 0.732 & 0.875 \\
    GLM-5.3-Flash & 0.801 & 0.710 & 0.882 & 0.834 & 0.770 & 0.824 \\
    Gemma-4-31B-IT & 0.733 & 0.781 & 0.899 & 0.468 & 0.771 & 0.844 \\
    DeepSeek-V4.1-Flash & 0.707 & 0.718 & 0.829 & 0.675 & 0.580 & 0.817 \\
    Qwen3.6-27B & 0.649 & 0.709 & 0.836 & 0.517 & 0.539 & 0.873 \\
    Qwen3.5-35B-A3B & 0.608 & 0.573 & 0.829 & 0.457 & 0.552 & 0.815 \\
    Qwen3-VL-32B-Instruct & 0.682 & 0.667 & 0.886 & 0.584 & 0.611 & 0.829 \\
    Qwen3.5-27B & 0.668 & 0.692 & 0.827 & 0.511 & 0.675 & 0.703 \\
    Qwen3.8-27B & 0.685 & 0.705 & 0.835 & 0.498 & 0.734 & 0.659 \\
    Qwen2.5-VL-7B-Instruct & 0.507 & 0.610 & 0.769 & 0.296 & 0.369 & 0.753 \\
    InternVL3-8B & 0.583 & 0.539 & 0.778 & 0.426 & 0.600 & 0.803 \\
    InternVL3.5-30B-A3B & 0.598 & 0.455 & 0.727 & 0.575 & 0.553 & 0.850 \\
    InternVL3.5-8B & 0.574 & 0.524 & 0.768 & 0.557 & 0.422 & 0.766 \\
    Qwen2.5-VL-3B-Instruct & 0.495 & 0.484 & 0.758 & 0.254 & 0.490 & 0.822 \\
    LLaVA-OneVision-7B & 0.500 & 0.524 & 0.757 & 0.257 & 0.436 & 0.706 \\
    Qwen3.5-4B & 0.534 & 0.551 & 0.715 & 0.398 & 0.467 & 0.631 \\
    Qwen3.5-2B & 0.477 & 0.539 & 0.728 & 0.343 & 0.375 & 0.606 \\
    Llama-3.2-11B-Vision-Instruct & 0.391 & 0.394 & 0.620 & 0.136 & 0.387 & 0.714 \\
    Qwen3.5-0.8B & 0.381 & 0.421 & 0.537 & 0.292 & 0.265 & 0.551 \\
    \midrule
    \multicolumn{7}{l}{\cellcolor{black!4}\textit{Open-source embodied models}} \\
    HY-Embodied-VLM-1.0 & 0.596 & 0.577 & 0.834 & 0.532 & 0.453 & 0.806 \\
    RynnBrain1.1-9B & 0.593 & 0.646 & 0.819 & 0.400 & 0.539 & 0.619 \\
    RynnBrain-30B-A3B & 0.656 & 0.659 & 0.777 & 0.592 & 0.601 & 0.717 \\
    VeBrain-7B & 0.546 & 0.507 & 0.801 & 0.360 & 0.519 & 0.782 \\
    Pelican1.0-VL-7B & 0.540 & 0.581 & 0.724 & 0.298 & 0.551 & 0.782 \\
    MiMo-Embodied-7B & 0.533 & 0.577 & 0.761 & 0.337 & 0.446 & 0.812 \\
    RynnBrain1.1-2B & 0.602 & 0.529 & 0.824 & 0.335 & 0.791 & 0.574 \\
    RoboBrain2.0-7B & 0.511 & 0.545 & 0.809 & 0.307 & 0.391 & 0.829 \\
    RoboBrain2.0-32B & 0.486 & 0.566 & 0.781 & 0.271 & 0.358 & 0.731 \\
    Embodied-R1-3B-V1 & 0.475 & 0.423 & 0.703 & 0.241 & 0.509 & 0.830 \\
    HY-Embodied-0.5 & 0.482 & 0.354 & 0.694 & 0.311 & 0.618 & 0.462 \\
    HY-Embodied-0.5-X & 0.461 & 0.305 & 0.677 & 0.257 & 0.629 & 0.471 \\
    \midrule
    \multicolumn{7}{l}{\cellcolor{black!4}\textit{Open-source streaming video models}} \\
    JoyAI-VL-Interaction-Preview & 0.639 & 0.522 & 0.761 & 0.519 & 0.737 & 0.777 \\
    AURA-8B & 0.624 & 0.535 & 0.834 & 0.522 & 0.626 & 0.737 \\
    StreamingVLM & 0.456 & 0.507 & 0.753 & 0.222 & 0.361 & 0.763 \\
    Flash-VStream-Qwen-7B & 0.470 & 0.516 & 0.675 & 0.334 & 0.400 & 0.503 \\
    VITA-1.5 & 0.460 & 0.425 & 0.638 & 0.379 & 0.446 & 0.487 \\
    \midrule
    \multicolumn{7}{l}{\cellcolor{black!4}\textit{Open-source omni-modal models}} \\
    Qwen2.5-Omni-7B & 0.507 & 0.531 & 0.745 & 0.295 & 0.455 & 0.784 \\
    Qwen3-Omni-30B-A3B-Instruct & 0.646 & 0.589 & 0.814 & 0.501 & 0.679 & 0.581 \\
    OmniVinci & 0.546 & 0.592 & 0.733 & 0.346 & 0.565 & 0.557 \\
    Qwen2.5-Omni-3B & 0.477 & 0.465 & 0.762 & 0.262 & 0.414 & 0.805 \\
    Phi-4-Multimodal-Instruct & 0.435 & 0.400 & 0.541 & 0.336 & 0.400 & 0.751 \\
\end{longtable}
\endgroup

\paragraph{Attention.}
Table~\ref{tab:supp-attention-results} reports Attention Overall in the All column and aggregates the 19 Attention capabilities into five diagnostic families. Gemini-3.7-Flash leads Overall (0.936), semantic target recognition (0.951), spatial localization (0.911), and joint identity-localization (0.985), whereas GPT-5.6-Sol is strongest on nonverbal states (0.926) and Gemini-3.1-Pro on multimodal attribution (0.912). The breakdown also exposes a recurring gap between identifying a target and grounding it: for example, Llama-3.2-11B-Vision-Instruct scores 0.620 on semantic target recognition but 0.136 on spatial localization, and Qwen2.5-VL-7B-Instruct scores 0.769 versus 0.296. Attention performance is therefore not captured by semantic recognition alone.

\begingroup
\scriptsize
\setlength{\tabcolsep}{2.0pt}
\renewcommand{\arraystretch}{1.08}
\setlength{\LTpre}{0pt}
\setlength{\LTpost}{3pt}
\begin{longtable}{@{}>{\raggedright\arraybackslash}p{0.30\textwidth}>{\columncolor{tableintention}}cccccc@{}}
\caption{Complete Intention results.}\label{tab:supp-intention-results}\\
\toprule
\textbf{Model} & \textbf{All} & \makecell{Key\\moment} & \textbf{Online} & \textbf{Evidence} & \textbf{FCR}$\downarrow$ & \textbf{IAS} \\
\midrule
\endfirsthead
\multicolumn{7}{l}{\small\itshape Table~\ref{tab:supp-intention-results} continued.}\\
\toprule
\textbf{Model} & \textbf{All} & \makecell{Key\\moment} & \textbf{Online} & \textbf{Evidence} & \textbf{FCR}$\downarrow$ & \textbf{IAS} \\
\midrule
\endhead
\midrule
\multicolumn{7}{r}{\small\itshape Continued on next page.}\\
\endfoot
\bottomrule
\endlastfoot
    \multicolumn{7}{l}{\cellcolor{black!4}\textit{Closed-source models}} \\
    Gemini-3.7-Flash & \textbf{0.569} & \textbf{0.549} & \textbf{0.569} & 0.617 & 0.421 & \textbf{0.237} \\
    Gemini-3.1-Pro & 0.496 & 0.425 & 0.523 & 0.587 & 0.220 & 0.126 \\
    GPT-5.6-Sol & 0.501 & 0.520 & 0.457 & 0.569 & 0.574 & 0.154 \\
    Doubao-seed-2.0-lite & 0.418 & 0.409 & 0.363 & 0.582 & 0.686 & 0.066 \\
    \midrule
    \multicolumn{7}{l}{\cellcolor{black!4}\textit{Open-source general-purpose models}} \\
    Qwen3.6-35B-A3B & 0.485 & 0.501 & 0.439 & 0.568 & 0.634 & 0.122 \\
    GLM-5.3-Flash & 0.501 & 0.513 & 0.455 & 0.590 & 0.644 & 0.128 \\
    Gemma-4-31B-IT & 0.502 & 0.501 & 0.482 & 0.554 & 0.372 & 0.167 \\
    DeepSeek-V4.1-Flash & 0.509 & 0.539 & 0.451 & 0.588 & 0.552 & 0.121 \\
    Qwen3.6-27B & 0.480 & 0.434 & 0.483 & 0.578 & 0.250 & 0.119 \\
    Qwen3.5-35B-A3B & 0.448 & 0.369 & 0.474 & 0.565 & 0.441 & 0.126 \\
    Qwen3-VL-32B-Instruct & 0.472 & 0.409 & 0.481 & 0.596 & 0.397 & 0.133 \\
    Qwen3.5-27B & 0.382 & 0.384 & 0.336 & 0.490 & 0.645 & 0.063 \\
    Qwen3.8-27B & 0.395 & 0.391 & 0.374 & 0.459 & 0.512 & 0.052 \\
    Qwen2.5-VL-7B-Instruct & 0.452 & 0.358 & 0.499 & 0.550 & 0.240 & 0.102 \\
    InternVL3-8B & 0.417 & 0.395 & 0.375 & 0.573 & 0.640 & 0.082 \\
    InternVL3.5-30B-A3B & 0.426 & 0.333 & 0.458 & 0.556 & 0.338 & 0.095 \\
    InternVL3.5-8B & 0.406 & 0.367 & 0.374 & 0.579 & 0.613 & 0.094 \\
    Qwen2.5-VL-3B-Instruct & 0.409 & 0.390 & 0.357 & 0.586 & 0.699 & 0.069 \\
    LLaVA-OneVision-7B & 0.387 & 0.355 & 0.380 & 0.476 & 0.507 & 0.078 \\
    Qwen3.5-4B & 0.389 & 0.287 & 0.442 & 0.486 & 0.475 & 0.123 \\
    Qwen3.5-2B & 0.375 & 0.351 & 0.356 & 0.479 & 0.499 & 0.037 \\
    Llama-3.2-11B-Vision-Instruct & 0.384 & 0.269 & 0.422 & 0.549 & 0.250 & 0.011 \\
    Qwen3.5-0.8B & 0.299 & 0.259 & 0.276 & 0.446 & 0.713 & 0.036 \\
    \midrule
    \multicolumn{7}{l}{\cellcolor{black!4}\textit{Open-source embodied models}} \\
    HY-Embodied-VLM-1.0 & 0.396 & 0.284 & 0.419 & 0.595 & 0.605 & 0.105 \\
    RynnBrain1.1-9B & 0.417 & 0.321 & 0.468 & 0.509 & 0.327 & 0.123 \\
    RynnBrain-30B-A3B & 0.435 & 0.382 & 0.471 & 0.461 & 0.245 & 0.091 \\
    VeBrain-7B & 0.449 & 0.358 & 0.494 & 0.541 & 0.186 & 0.111 \\
    Pelican1.0-VL-7B & 0.462 & 0.372 & 0.503 & 0.562 & 0.142 & 0.099 \\
    MiMo-Embodied-7B & 0.440 & 0.396 & 0.407 & \textbf{0.623} & 0.640 & 0.114 \\
    RynnBrain1.1-2B & 0.396 & 0.346 & 0.430 & 0.424 & 0.333 & 0.073 \\
    RoboBrain2.0-7B & 0.439 & 0.368 & 0.447 & 0.581 & 0.254 & 0.054 \\
    RoboBrain2.0-32B & 0.451 & 0.423 & 0.444 & 0.534 & 0.199 & 0.049 \\
    Embodied-R1-3B-V1 & 0.408 & 0.368 & 0.376 & 0.580 & 0.618 & 0.073 \\
    HY-Embodied-0.5 & 0.368 & 0.315 & 0.388 & 0.436 & 0.328 & 0.022 \\
    HY-Embodied-0.5-X & 0.373 & 0.319 & 0.392 & 0.443 & 0.247 & 0.008 \\
    \midrule
    \multicolumn{7}{l}{\cellcolor{black!4}\textit{Open-source streaming video models}} \\
    JoyAI-VL-Interaction-Preview & 0.436 & 0.407 & 0.408 & 0.570 & 0.529 & 0.113 \\
    AURA-8B & 0.472 & 0.411 & 0.490 & 0.563 & 0.242 & 0.113 \\
    StreamingVLM & 0.400 & 0.258 & 0.449 & 0.598 & 0.149 & 0.029 \\
    Flash-VStream-Qwen-7B & 0.396 & 0.384 & 0.398 & 0.415 & 0.518 & 0.065 \\
    VITA-1.5 & 0.388 & 0.324 & 0.435 & 0.412 & 0.320 & 0.091 \\
    \midrule
    \multicolumn{7}{l}{\cellcolor{black!4}\textit{Open-source omni-modal models}} \\
    Qwen2.5-Omni-7B & 0.453 & 0.378 & 0.485 & 0.543 & 0.137 & 0.070 \\
    Qwen3-Omni-30B-A3B-Instruct & 0.419 & 0.323 & 0.498 & 0.439 & 0.291 & 0.153 \\
    OmniVinci & 0.401 & 0.388 & 0.410 & 0.409 & 0.474 & 0.116 \\
    Qwen2.5-Omni-3B & 0.433 & 0.335 & 0.466 & 0.574 & 0.220 & 0.065 \\
    Phi-4-Multimodal-Instruct & 0.398 & 0.285 & 0.426 & 0.587 & \textbf{0.083} & 0.000 \\
\end{longtable}
\endgroup

\paragraph{Intention.}
In Table~\ref{tab:supp-intention-results}, All denotes Intention Overall, FCR is the false-commitment rate (lower is better), and IAS measures timely, stable intent recovery after penalizing premature commitments. Gemini-3.7-Flash obtains the highest Overall score (0.569) and IAS (0.237), but still has an FCR of 0.421. The converse failure is visible for Phi-4-Multimodal-Instruct: it has the lowest FCR (0.083) but an IAS of 0.000. Conservative prediction is thus insufficient by itself; a model must also recover the correct intent early and maintain it.

\begingroup
\scriptsize
\setlength{\tabcolsep}{2.0pt}
\renewcommand{\arraystretch}{1.08}
\setlength{\LTpre}{0pt}
\setlength{\LTpost}{3pt}
\begin{longtable}{@{}>{\raggedright\arraybackslash}p{0.255\textwidth}>{\columncolor{tablereaction}}ccccccccc@{}}
\caption{Complete Reaction results with action-domain diagnostics.}\label{tab:supp-reaction-results}\\
\toprule
\multirow{2}{*}{\textbf{Model}} & \multicolumn{5}{c}{\cellcolor{tablereaction}\textbf{Core Reaction metrics}} & \multicolumn{4}{c}{\cellcolor{tablesetting}\textbf{Subaction + Speak by action domain}} \\
\cmidrule(lr){2-6}\cmidrule(lr){7-10}
& \textbf{All} & \textbf{Engage} & \makecell{Action\\domain} & \makecell{Subaction\\+ Speak} & \makecell{Joint\\EM} & \textbf{Comm.} & \textbf{Nav.} & \textbf{Manip.} & \makecell{Refrain/\\Idle} \\
\midrule
\endfirsthead
\multicolumn{10}{l}{\small\itshape Table~\ref{tab:supp-reaction-results} continued.}\\
\toprule
\multirow{2}{*}{\textbf{Model}} & \multicolumn{5}{c}{\cellcolor{tablereaction}\textbf{Core Reaction metrics}} & \multicolumn{4}{c}{\cellcolor{tablesetting}\textbf{Subaction + Speak by action domain}} \\
\cmidrule(lr){2-6}\cmidrule(lr){7-10}
& \textbf{All} & \textbf{Engage} & \makecell{Action\\domain} & \makecell{Subaction\\+ Speak} & \makecell{Joint\\EM} & \textbf{Comm.} & \textbf{Nav.} & \textbf{Manip.} & \makecell{Refrain/\\Idle} \\
\midrule
\endhead
\midrule
\multicolumn{10}{r}{\small\itshape Continued on next page.}\\
\endfoot
\bottomrule
\endlastfoot
    \multicolumn{10}{l}{\cellcolor{black!4}\textit{Closed-source models}} \\
    Gemini-3.7-Flash & 0.638 & 0.546 & 0.831 & 0.486 & 0.083 & 0.477 & 0.417 & 0.484 & 0.414 \\
    Gemini-3.1-Pro & 0.620 & 0.530 & 0.838 & 0.487 & 0.111 & 0.487 & 0.498 & 0.494 & 0.443 \\
    GPT-5.6-Sol & 0.489 & 0.261 & 0.829 & 0.426 & 0.128 & 0.438 & 0.484 & 0.446 & 0.466 \\
    Doubao-seed-2.0-lite & 0.619 & 0.564 & 0.792 & 0.435 & 0.092 & 0.483 & 0.526 & 0.504 & 0.573 \\
    \midrule
    \multicolumn{10}{l}{\cellcolor{black!4}\textit{Open-source general-purpose models}} \\
    Qwen3.6-35B-A3B & 0.655 & 0.605 & 0.818 & 0.446 & 0.064 & 0.466 & 0.463 & 0.451 & 0.530 \\
    GLM-5.3-Flash & 0.590 & 0.531 & 0.810 & 0.379 & 0.067 & 0.403 & 0.463 & 0.428 & 0.384 \\
    Gemma-4-31B-IT & 0.535 & 0.394 & 0.810 & 0.471 & 0.123 & 0.452 & 0.416 & 0.390 & 0.370 \\
    DeepSeek-V4.1-Flash & 0.548 & 0.456 & 0.815 & 0.401 & 0.108 & 0.420 & 0.470 & 0.391 & 0.313 \\
    Qwen3.6-27B & 0.545 & 0.383 & 0.824 & 0.453 & 0.128 & 0.488 & 0.457 & 0.472 & 0.448 \\
    Qwen3.5-35B-A3B & 0.609 & 0.563 & \textbf{0.848} & 0.428 & 0.100 & 0.421 & 0.445 & 0.384 & 0.447 \\
    Qwen3-VL-32B-Instruct & 0.479 & 0.217 & 0.826 & 0.425 & 0.144 & 0.452 & 0.489 & 0.452 & 0.457 \\
    Qwen3.5-27B & 0.539 & 0.446 & 0.684 & 0.481 & 0.123 & 0.504 & 0.506 & 0.474 & 0.513 \\
    Qwen3.8-27B & 0.475 & 0.251 & 0.651 & 0.491 & 0.137 & 0.541 & 0.605 & 0.597 & 0.567 \\
    Qwen2.5-VL-7B-Instruct & 0.466 & 0.195 & 0.758 & 0.461 & 0.144 & 0.489 & 0.498 & 0.459 & 0.416 \\
    InternVL3-8B & 0.518 & 0.271 & 0.781 & 0.499 & 0.123 & 0.534 & 0.590 & 0.484 & 0.574 \\
    InternVL3.5-30B-A3B & 0.442 & 0.134 & 0.816 & 0.426 & 0.144 & 0.441 & 0.457 & 0.414 & 0.503 \\
    InternVL3.5-8B & 0.458 & 0.309 & 0.752 & 0.394 & 0.119 & 0.388 & 0.315 & 0.352 & 0.406 \\
    Qwen2.5-VL-3B-Instruct & 0.483 & 0.142 & 0.789 & 0.475 & 0.144 & 0.548 & 0.583 & 0.589 & 0.651 \\
    LLaVA-OneVision-7B & 0.379 & 0.132 & 0.704 & 0.338 & 0.144 & 0.350 & 0.328 & 0.340 & 0.438 \\
    Qwen3.5-4B & 0.481 & 0.453 & 0.605 & 0.413 & 0.129 & 0.412 & 0.395 & 0.343 & 0.365 \\
    Qwen3.5-2B & 0.530 & 0.489 & 0.540 & 0.385 & 0.036 & 0.436 & 0.446 & 0.454 & 0.433 \\
    Llama-3.2-11B-Vision-Instruct & 0.463 & 0.371 & 0.638 & 0.364 & 0.128 & 0.426 & 0.407 & 0.492 & 0.519 \\
    Qwen3.5-0.8B & 0.506 & 0.517 & 0.460 & 0.343 & 0.044 & 0.427 & 0.482 & 0.493 & 0.489 \\
    \midrule
    \multicolumn{10}{l}{\cellcolor{black!4}\textit{Open-source embodied models}} \\
    HY-Embodied-VLM-1.0 & \textbf{0.688} & 0.485 & 0.749 & 0.522 & 0.028 & 0.558 & 0.575 & 0.502 & 0.549 \\
    RynnBrain1.1-9B & 0.557 & 0.605 & 0.760 & 0.315 & 0.082 & 0.300 & 0.345 & 0.249 & 0.244 \\
    RynnBrain-30B-A3B & 0.410 & 0.245 & 0.789 & 0.329 & \textbf{0.147} & 0.325 & 0.260 & 0.222 & 0.241 \\
    VeBrain-7B & 0.470 & 0.154 & 0.800 & 0.440 & 0.141 & 0.501 & 0.550 & 0.517 & 0.547 \\
    Pelican1.0-VL-7B & 0.440 & 0.128 & 0.734 & 0.441 & 0.144 & 0.500 & 0.546 & 0.535 & 0.478 \\
    MiMo-Embodied-7B & 0.643 & 0.558 & 0.785 & 0.463 & 0.056 & 0.512 & 0.550 & 0.536 & 0.622 \\
    RynnBrain1.1-2B & 0.467 & 0.426 & 0.719 & 0.334 & 0.090 & 0.299 & 0.257 & 0.235 & 0.349 \\
    RoboBrain2.0-7B & 0.471 & 0.184 & 0.782 & 0.448 & 0.146 & 0.488 & 0.420 & 0.527 & 0.516 \\
    RoboBrain2.0-32B & 0.441 & 0.229 & 0.709 & 0.407 & 0.136 & 0.427 & 0.354 & 0.403 & 0.462 \\
    Embodied-R1-3B-V1 & 0.468 & 0.126 & 0.784 & 0.451 & 0.144 & 0.518 & 0.552 & 0.576 & \textbf{0.655} \\
    HY-Embodied-0.5 & 0.563 & \textbf{0.606} & 0.692 & 0.295 & 0.065 & 0.343 & 0.356 & 0.308 & 0.359 \\
    HY-Embodied-0.5-X & 0.480 & 0.515 & 0.671 & 0.290 & 0.126 & 0.333 & 0.337 & 0.315 & 0.363 \\
    \midrule
    \multicolumn{10}{l}{\cellcolor{black!4}\textit{Open-source streaming video models}} \\
    JoyAI-VL-Interaction-Preview & 0.610 & 0.508 & 0.752 & 0.421 & 0.029 & 0.434 & 0.455 & 0.372 & 0.529 \\
    AURA-8B & 0.434 & 0.146 & 0.812 & 0.391 & 0.144 & 0.389 & 0.414 & 0.331 & 0.404 \\
    StreamingVLM & 0.536 & 0.302 & 0.733 & \textbf{0.534} & 0.141 & \textbf{0.600} & \textbf{0.609} & \textbf{0.650} & 0.622 \\
    Flash-VStream-Qwen-7B & 0.376 & 0.131 & 0.711 & 0.341 & 0.142 & 0.366 & 0.356 & 0.298 & 0.383 \\
    VITA-1.5 & 0.420 & 0.344 & 0.655 & 0.342 & 0.103 & 0.313 & 0.300 & 0.251 & 0.330 \\
    \midrule
    \multicolumn{10}{l}{\cellcolor{black!4}\textit{Open-source omni-modal models}} \\
    Qwen2.5-Omni-7B & 0.449 & 0.126 & 0.796 & 0.411 & 0.144 & 0.453 & 0.439 & 0.479 & 0.485 \\
    Qwen3-Omni-30B-A3B-Instruct & 0.390 & 0.138 & 0.757 & 0.343 & 0.141 & 0.356 & 0.369 & 0.310 & 0.338 \\
    OmniVinci & 0.340 & 0.126 & 0.713 & 0.256 & 0.144 & 0.265 & 0.286 & 0.221 & 0.263 \\
    Qwen2.5-Omni-3B & 0.467 & 0.126 & 0.769 & 0.465 & 0.144 & 0.524 & 0.559 & 0.523 & 0.594 \\
    Phi-4-Multimodal-Instruct & 0.435 & 0.133 & 0.648 & 0.435 & 0.137 & 0.524 & 0.502 & 0.599 & 0.647 \\
\end{longtable}
\endgroup

\paragraph{Reaction.}
Table~\ref{tab:supp-reaction-results} reports the Overall Reaction performance across all events. For individual metrics, Engage uses an active-versus-non-intervention Macro-F1, action domain and subaction use multi-label F1, and the Speak subtype uses accuracy. For Communication events with a Speak annotation, the domain score averages the subaction and Speak-subtype scores; note that the four domain columns are non-exclusive. HY-Embodied-VLM-1.0 achieves the highest Overall score (0.688) but yields only 0.028 in Joint-EM, whereas RynnBrain-30B-A3B attains the highest Joint-EM (0.147) despite a lower Overall score of 0.410. StreamingVLM exhibits a complementary profile: it leads in Subaction + Speak (0.534) and the Communication, Navigation, and Manipulation breakdowns (0.600, 0.609, and 0.650), yet its Engage score remains low at 0.302. This indicates that strong conditional action specification does not guarantee correct engagement decisions or the successful composition of full reaction chains.

\begingroup
\scriptsize
\setlength{\tabcolsep}{2.0pt}
\renewcommand{\arraystretch}{1.08}
\setlength{\LTpre}{0pt}
\setlength{\LTpost}{3pt}
\begin{longtable}{@{}>{\raggedright\arraybackslash}p{0.32\textwidth}>{\columncolor{tableinteraction}}ccccc@{}}
\caption{Complete Interaction results.}\label{tab:supp-interaction-results}\\
\toprule
\textbf{Model} & \textbf{All} & \textbf{State} & \makecell{Task\\outcome} & \textbf{Compliance} & \textbf{Violation} \\
\midrule
\endfirsthead
\multicolumn{6}{l}{\small\itshape Table~\ref{tab:supp-interaction-results} continued.}\\
\toprule
\textbf{Model} & \textbf{All} & \textbf{State} & \makecell{Task\\outcome} & \textbf{Compliance} & \textbf{Violation} \\
\midrule
\endhead
\midrule
\multicolumn{6}{r}{\small\itshape Continued on next page.}\\
\endfoot
\bottomrule
\endlastfoot
    \multicolumn{6}{l}{\cellcolor{black!4}\textit{Closed-source models}} \\
    Gemini-3.7-Flash & \textbf{0.760} & \textbf{0.441} & 0.885 & 0.903 & 0.525 \\
    Gemini-3.1-Pro & \textbf{0.760} & 0.368 & 0.903 & 0.902 & 0.495 \\
    GPT-5.6-Sol & 0.729 & 0.400 & 0.880 & 0.906 & 0.494 \\
    Doubao-seed-2.0-lite & 0.669 & 0.345 & 0.870 & 0.726 & 0.572 \\
    \midrule
    \multicolumn{6}{l}{\cellcolor{black!4}\textit{Open-source general-purpose models}} \\
    Qwen3.6-35B-A3B & 0.738 & 0.293 & 0.927 & 0.881 & 0.577 \\
    GLM-5.3-Flash & 0.638 & 0.386 & 0.823 & 0.706 & 0.470 \\
    Gemma-4-31B-IT & 0.728 & 0.281 & 0.812 & 0.907 & 0.495 \\
    DeepSeek-V4.1-Flash & 0.657 & 0.339 & 0.828 & 0.737 & 0.508 \\
    Qwen3.6-27B & 0.733 & 0.235 & 0.853 & 0.905 & 0.501 \\
    Qwen3.5-35B-A3B & 0.715 & 0.258 & 0.900 & 0.877 & 0.461 \\
    Qwen3-VL-32B-Instruct & 0.723 & 0.299 & 0.690 & \textbf{0.923} & \textbf{0.589} \\
    Qwen3.5-27B & 0.656 & 0.264 & 0.870 & 0.768 & 0.478 \\
    Qwen3.8-27B & 0.632 & 0.339 & 0.860 & 0.717 & 0.341 \\
    Qwen2.5-VL-7B-Instruct & 0.713 & 0.310 & 0.870 & 0.863 & 0.483 \\
    InternVL3-8B & 0.603 & 0.212 & 0.925 & 0.724 & 0.318 \\
    InternVL3.5-30B-A3B & 0.639 & 0.226 & 0.730 & 0.754 & 0.511 \\
    InternVL3.5-8B & 0.562 & 0.270 & 0.728 & 0.689 & 0.351 \\
    Qwen2.5-VL-3B-Instruct & 0.515 & 0.200 & 0.903 & 0.538 & 0.385 \\
    LLaVA-OneVision-7B & 0.602 & 0.232 & 0.887 & 0.657 & 0.453 \\
    Qwen3.5-4B & 0.425 & 0.241 & 0.417 & 0.523 & 0.322 \\
    Qwen3.5-2B & 0.396 & 0.238 & 0.657 & 0.320 & 0.347 \\
    Llama-3.2-11B-Vision-Instruct & 0.455 & 0.249 & 0.608 & 0.498 & 0.381 \\
    Qwen3.5-0.8B & 0.398 & 0.281 & 0.740 & 0.353 & 0.329 \\
    \midrule
    \multicolumn{6}{l}{\cellcolor{black!4}\textit{Open-source embodied models}} \\
    HY-Embodied-VLM-1.0 & 0.716 & 0.209 & 0.943 & 0.873 & 0.481 \\
    RynnBrain1.1-9B & 0.736 & 0.275 & 0.818 & 0.918 & 0.583 \\
    RynnBrain-30B-A3B & 0.674 & 0.252 & 0.927 & 0.723 & 0.508 \\
    VeBrain-7B & 0.701 & 0.270 & 0.905 & 0.828 & 0.502 \\
    Pelican1.0-VL-7B & 0.700 & 0.206 & 0.877 & 0.875 & 0.470 \\
    MiMo-Embodied-7B & 0.487 & 0.322 & 0.215 & 0.669 & 0.466 \\
    RynnBrain1.1-2B & 0.567 & 0.238 & 0.807 & 0.530 & 0.490 \\
    RoboBrain2.0-7B & 0.597 & 0.229 & 0.943 & 0.696 & 0.379 \\
    RoboBrain2.0-32B & 0.571 & 0.220 & 0.920 & 0.639 & 0.367 \\
    Embodied-R1-3B-V1 & 0.569 & 0.197 & 0.828 & 0.635 & 0.412 \\
    HY-Embodied-0.5 & 0.429 & 0.238 & \textbf{0.960} & 0.353 & 0.318 \\
    HY-Embodied-0.5-X & 0.415 & 0.290 & 0.905 & 0.306 & 0.338 \\
    \midrule
    \multicolumn{6}{l}{\cellcolor{black!4}\textit{Open-source streaming video models}} \\
    JoyAI-VL-Interaction-Preview & 0.586 & 0.287 & 0.943 & 0.635 & 0.417 \\
    AURA-8B & 0.601 & 0.342 & 0.468 & 0.766 & 0.510 \\
    StreamingVLM & 0.655 & 0.342 & 0.925 & 0.785 & 0.413 \\
    Flash-VStream-Qwen-7B & 0.631 & 0.232 & 0.940 & 0.682 & 0.436 \\
    VITA-1.5 & 0.477 & 0.243 & 0.542 & 0.523 & 0.468 \\
    \midrule
    \multicolumn{6}{l}{\cellcolor{black!4}\textit{Open-source omni-modal models}} \\
    Qwen2.5-Omni-7B & 0.705 & 0.267 & 0.820 & 0.855 & 0.510 \\
    Qwen3-Omni-30B-A3B-Instruct & 0.647 & 0.357 & 0.848 & 0.746 & 0.415 \\
    OmniVinci & 0.651 & 0.316 & 0.785 & 0.740 & 0.454 \\
    Qwen2.5-Omni-3B & 0.520 & 0.258 & 0.830 & 0.534 & 0.200 \\
    Phi-4-Multimodal-Instruct & 0.356 & 0.217 & 0.917 & 0.172 & 0.301 \\
\end{longtable}
\endgroup

\paragraph{Interaction.}
Table~\ref{tab:supp-interaction-results} reports Interaction Overall in All. Compliance equally averages spatial, physical, communication, and overall compliance, while Violation equally averages violation-presence F1, online violation type, and segment-level violation types. Gemini-3.7-Flash and Gemini-3.1-Pro tie for the best Overall score (0.760), whereas Qwen3-VL-32B-Instruct leads both Compliance (0.923) and Violation (0.589). Component scores can nevertheless diverge sharply: HY-Embodied-0.5 attains the highest Task Outcome score (0.960) but only 0.429 Overall, with 0.238 on State and 0.353 on Compliance. Accurate resolved-outcome prediction therefore does not imply reliable tracking or assessment of the full interaction.

\clearpage

\subsection{Benchmark Visualizations and Qualitative Analysis}
\label{app:qualitative}

\newcommand{\airiscasegallery}{}
\newcounter{airiscaseonpage}
\newlength{\airiscaserowheight}
\setlength{\airiscaserowheight}{0.3333\textheight}

\ExplSyntaxOn
\box_new:N \l_airis_case_image_box
\dim_new:N \l_airis_case_image_width_dim
\cs_new_protected:Npn \airisuniformcaseimage #1
  {
    \hbox_set:Nn \l_airis_case_image_box { \includegraphics{#1} }
    \dim_set:Nn \l_airis_case_image_width_dim
      {
        \fp_to_dim:n
          {
            sqrt(47000 * \dim_to_fp:n { \box_wd:N \l_airis_case_image_box }
              / (\dim_to_fp:n { \box_ht:N \l_airis_case_image_box }
                 + \dim_to_fp:n { \box_dp:N \l_airis_case_image_box }))
          }
      }
    \resizebox{\l_airis_case_image_width_dim}{!}
      {\box_use:N \l_airis_case_image_box}
  }
\ExplSyntaxOff

\newcommand{\airisrendercase}[4]{%
  \ifnum\value{airiscaseonpage}=3\relax
    \newpage
    \setcounter{airiscaseonpage}{0}%
  \fi
  \par\nointerlineskip
  \vbox to \airiscaserowheight{%
    \hsize=\textwidth
    \vfil
    \hbox to \textwidth{\hfil\airisuniformcaseimage{#1}\hfil}%
    \captionsetup{font=small,skip=2pt,hypcap=false}%
    \captionof{figure}{Representative #4 case for #2.}%
    \label{#3}%
    \vfil
  }%
  \stepcounter{airiscaseonpage}%
}
\newcommand{\airisqueuecase}[4]{%
  \gappto\airiscasegallery{\airisrendercase{#1}{#2}{#3}{#4}}%
}

\subsubsection{Attention cases}
\label{app:attention-cases}

\newcommand{\airisattentioncase}[3]{%
  \airisqueuecase{#1}{#2}{#3}{Attention}}

Figures~\ref{fig:attn-body-orientation-bbox}-\ref{fig:attn-overall-target-pair}
present representative streaming examples for all 19 Attention capabilities.
To balance visual consistency with scenario diversity, adjacent capabilities are
grouped within four expertly verified episodes: a professional craft workshop,
a retail checkout, an outdoor farmers' market, and a card shop checkout.
Each panel displays only the causal observation prefix available at the designated
query time, followed by the original options and the verified gold response.
For tasks requiring spatial or combined semantic and spatial reasoning, candidate
bounding boxes are additionally overlaid on the final visible frame.

\airisattentioncase{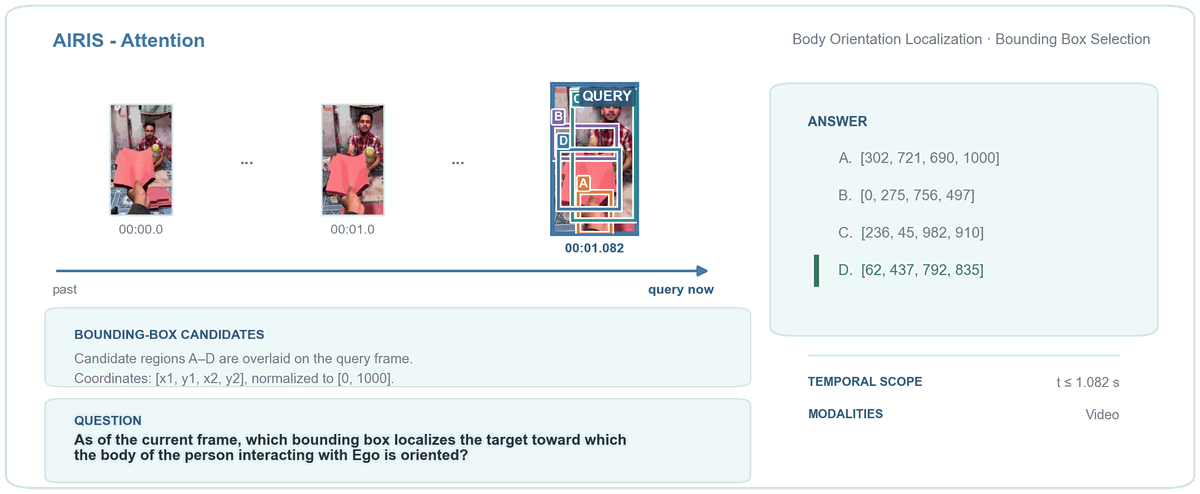}{body-orientation target localization}{fig:attn-body-orientation-bbox}
\airisattentioncase{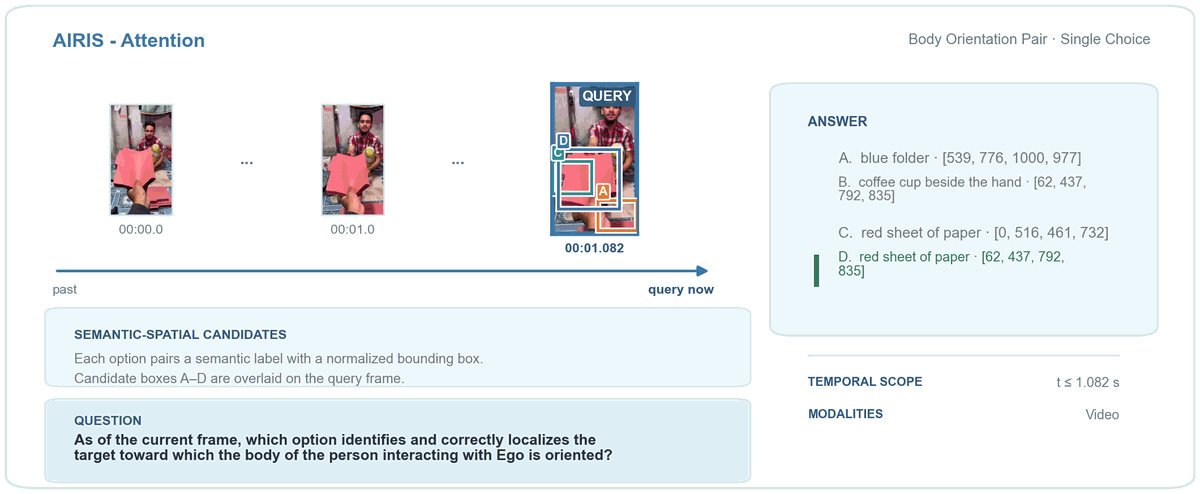}{joint body-orientation target identification and localization}{fig:attn-body-orientation-pair}
\airisattentioncase{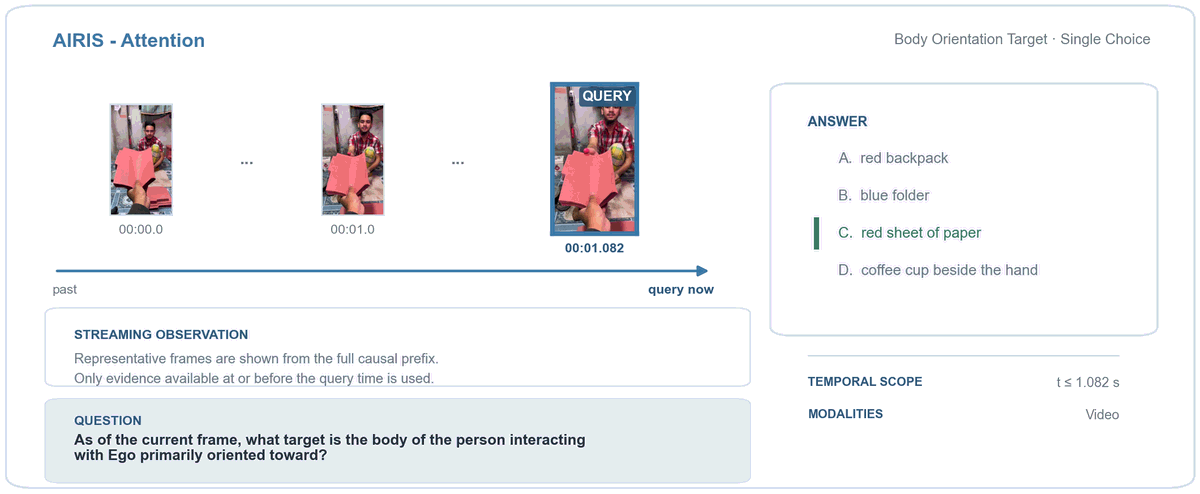}{body-orientation target recognition}{fig:attn-body-orientation-target}
\airisattentioncase{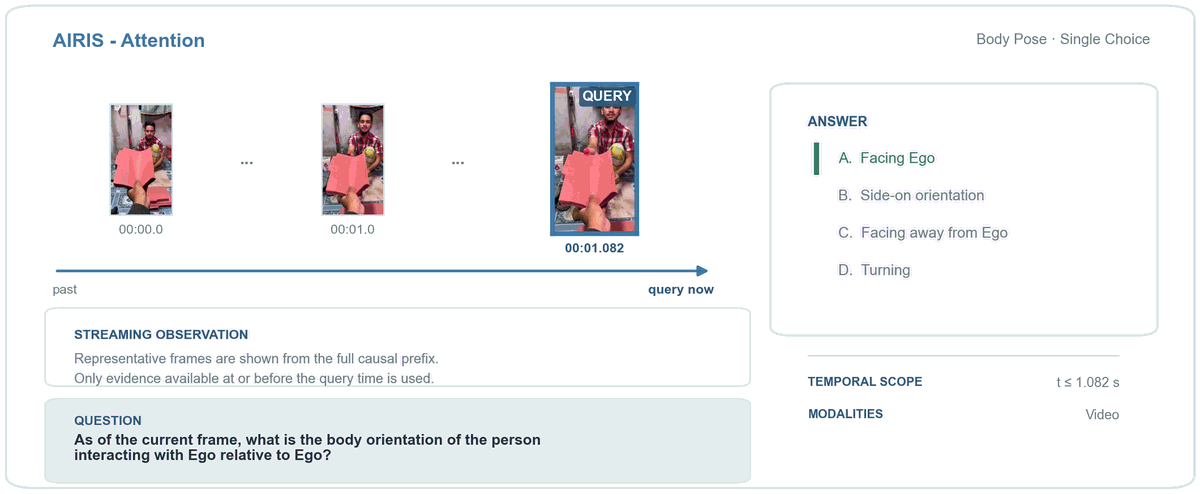}{body-pose recognition}{fig:attn-body-pose}
\airisattentioncase{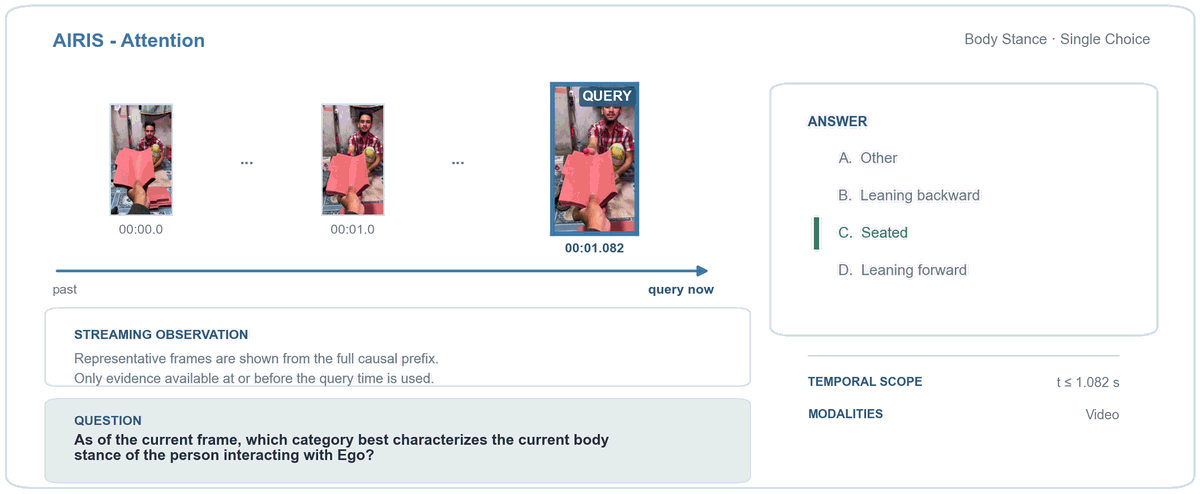}{body-stance recognition}{fig:attn-body-stance}
\airisattentioncase{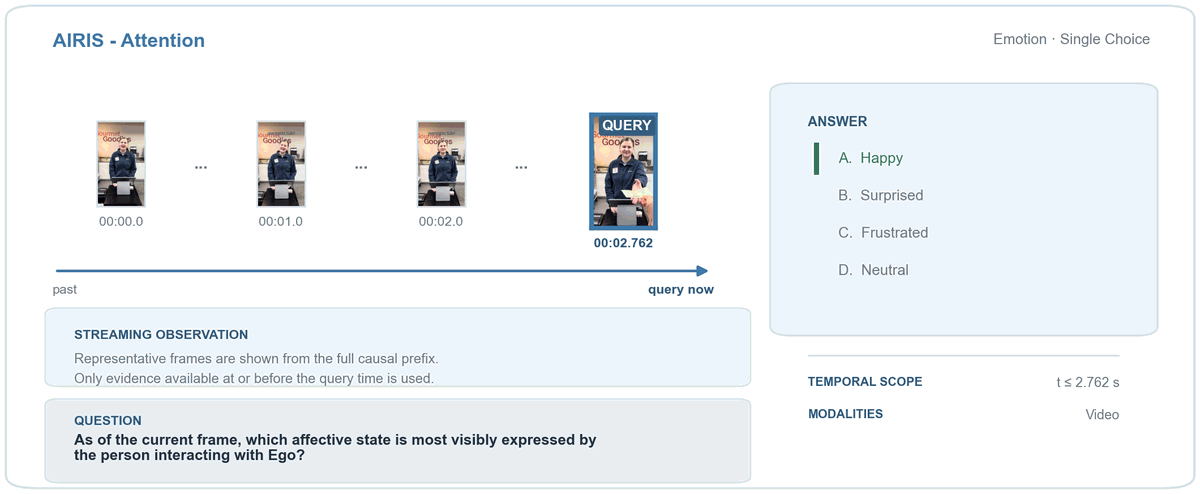}{emotion recognition}{fig:attn-emotion}
\airisattentioncase{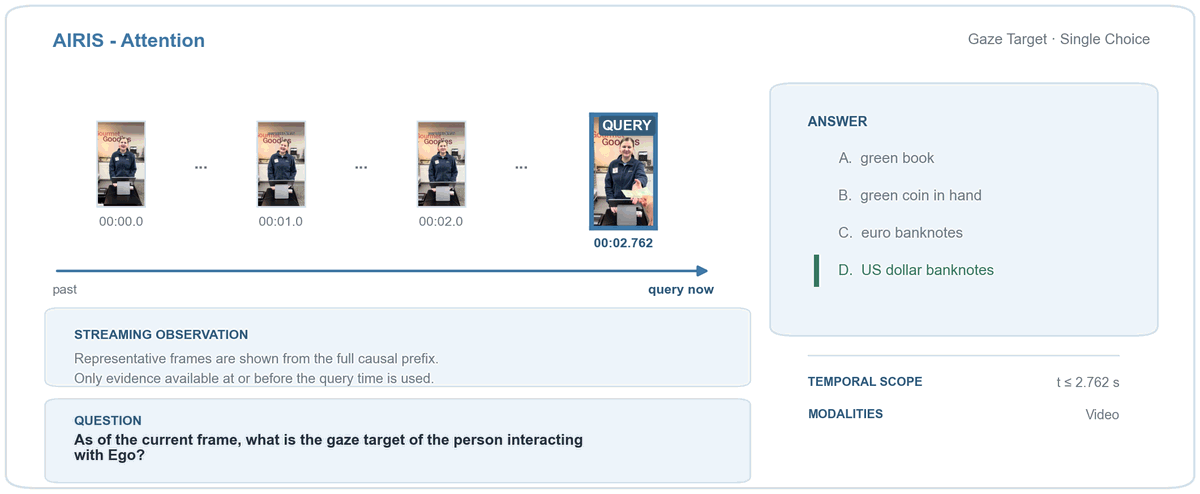}{gaze-target recognition}{fig:attn-gaze-target}
\airisattentioncase{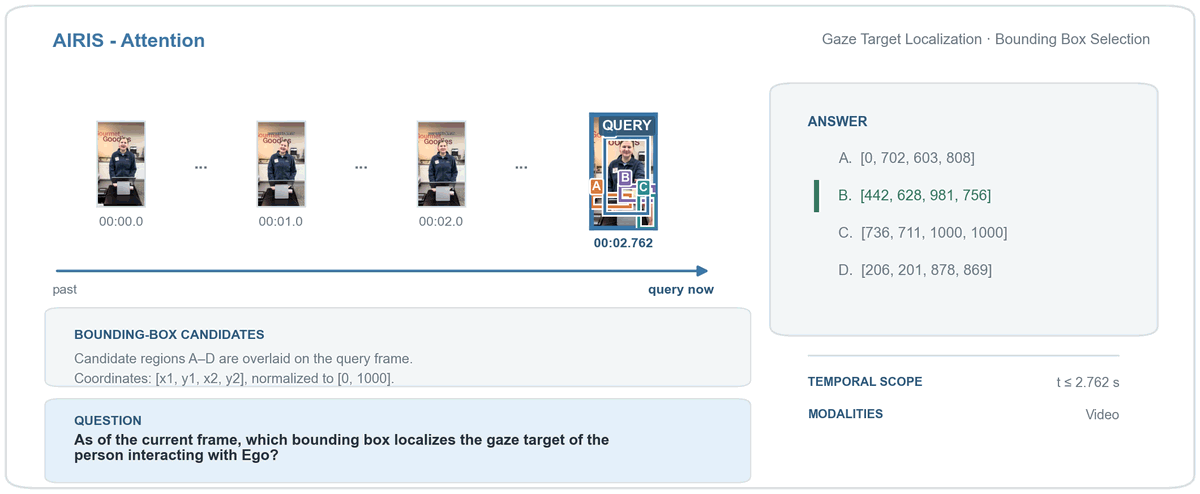}{gaze-target localization}{fig:attn-gaze-target-bbox}
\airisattentioncase{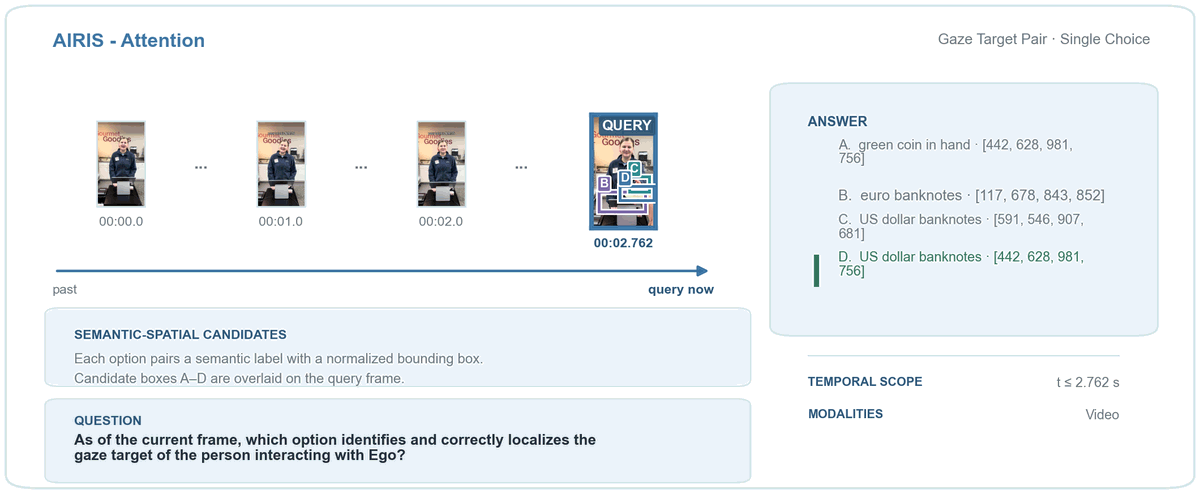}{joint gaze-target identification and localization}{fig:attn-gaze-target-pair}
\airisattentioncase{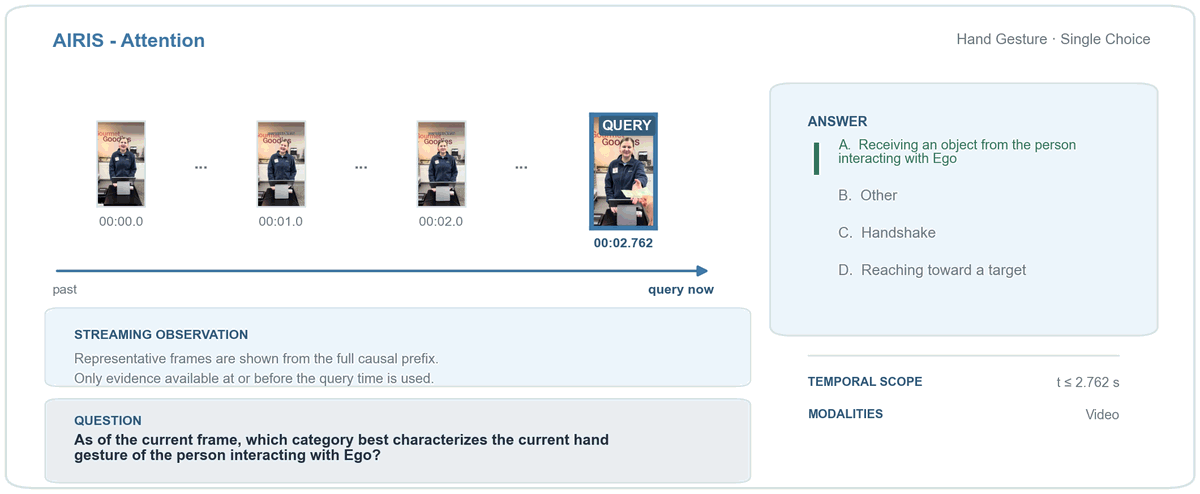}{hand-gesture recognition}{fig:attn-hand-gesture}
\airisattentioncase{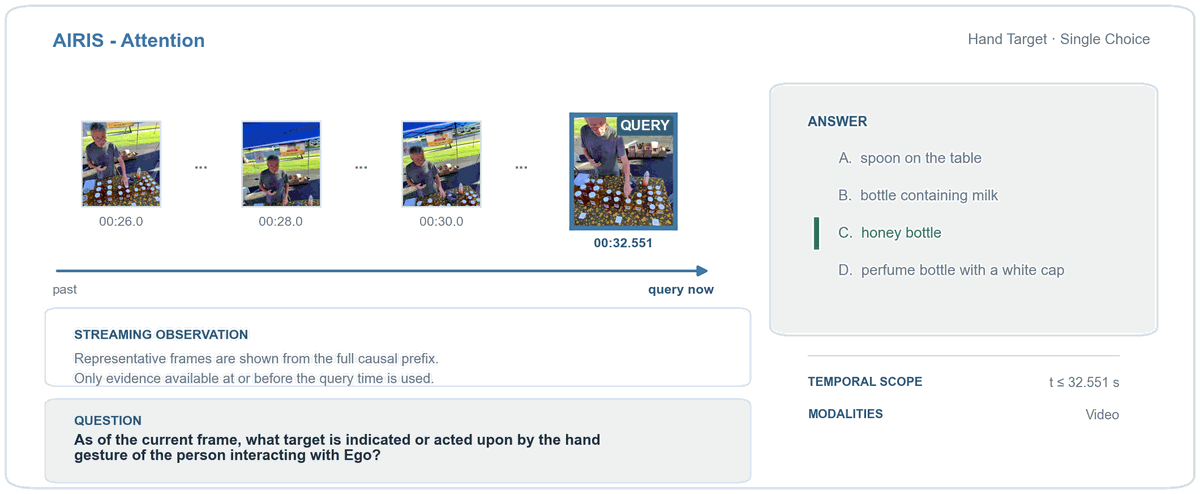}{hand-target recognition}{fig:attn-hand-target}
\airisattentioncase{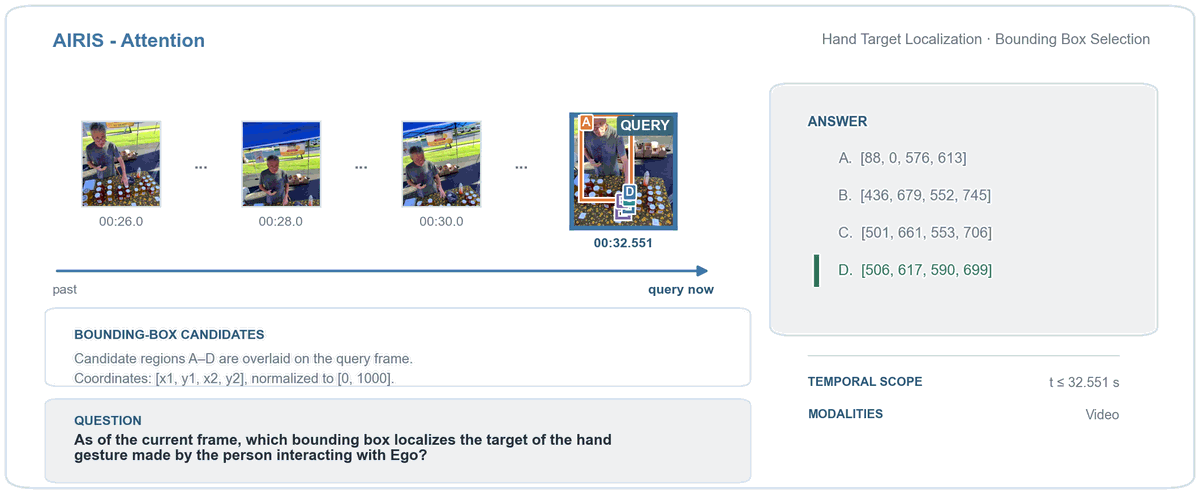}{hand-target localization}{fig:attn-hand-target-bbox}
\airisattentioncase{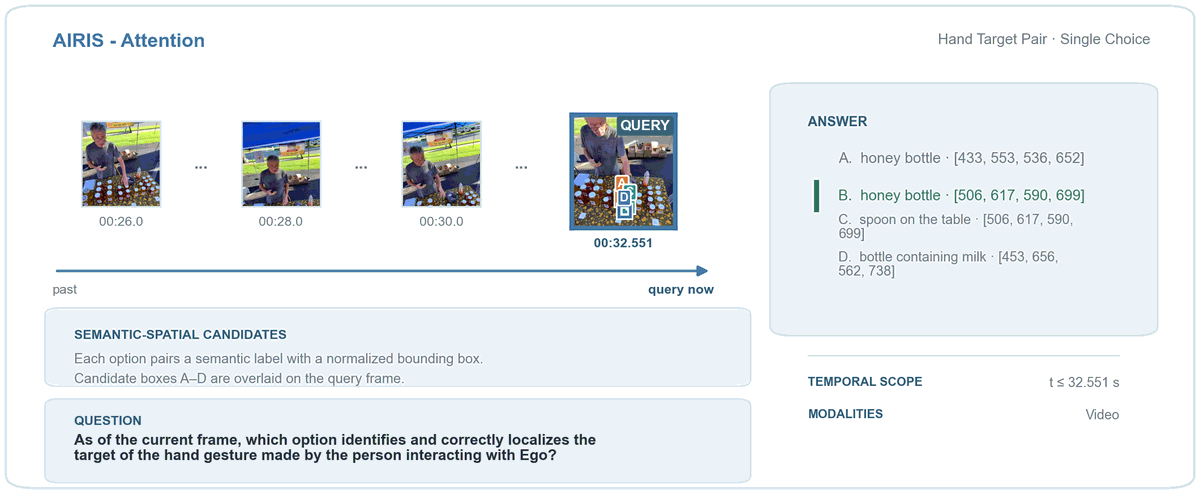}{joint hand-target identification and localization}{fig:attn-hand-target-pair}
\airisattentioncase{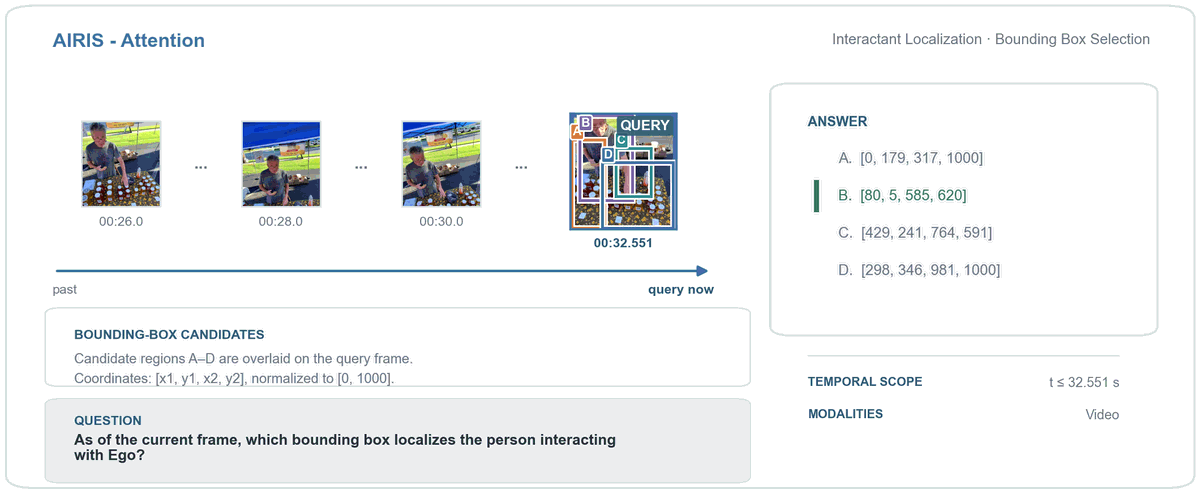}{interactant localization}{fig:attn-interactant-bbox}
\airisattentioncase{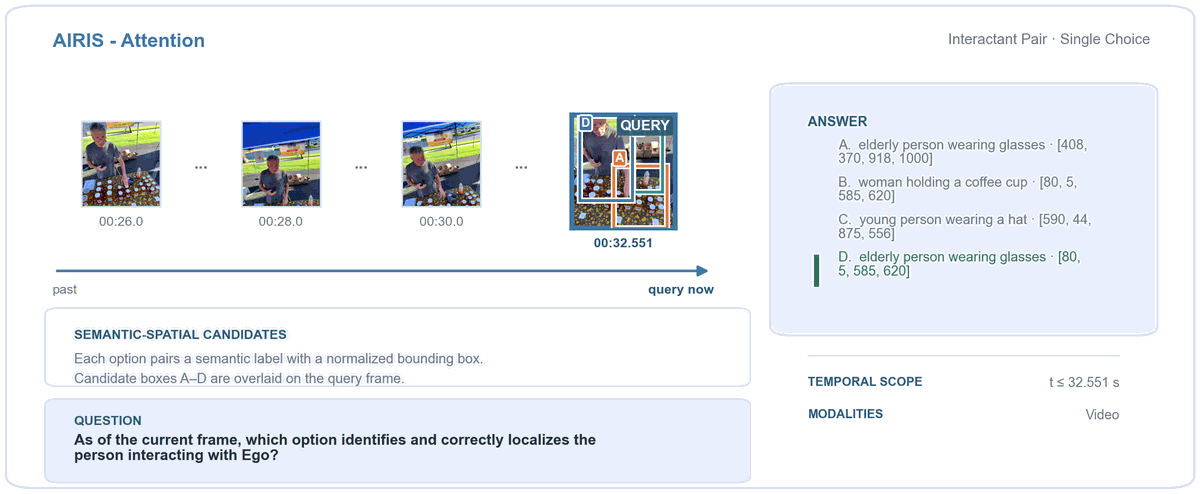}{joint interactant identification and localization}{fig:attn-interactant-pair}
\airisattentioncase{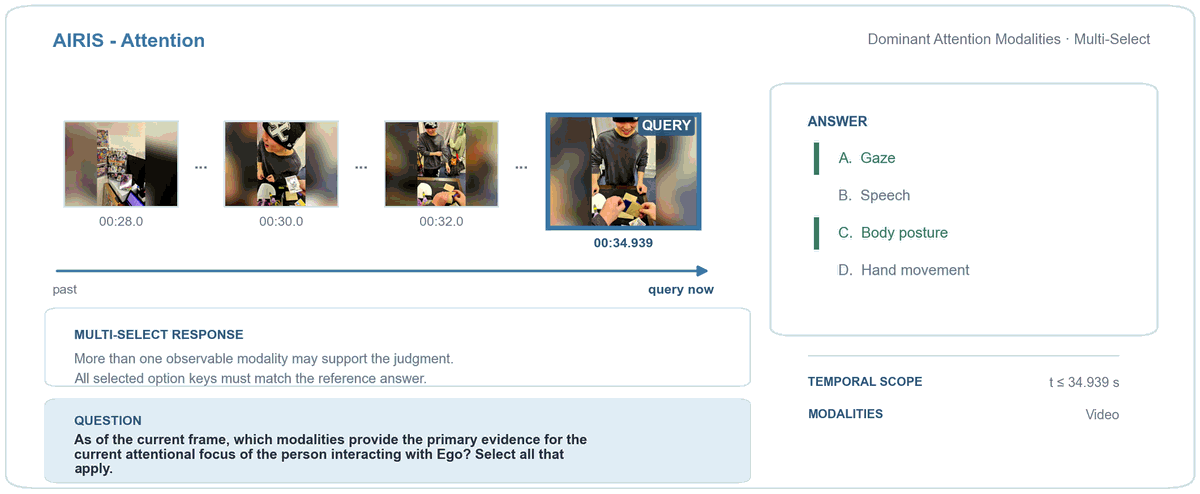}{dominant-attention modality recognition}{fig:attn-overall-modalities}
\airisattentioncase{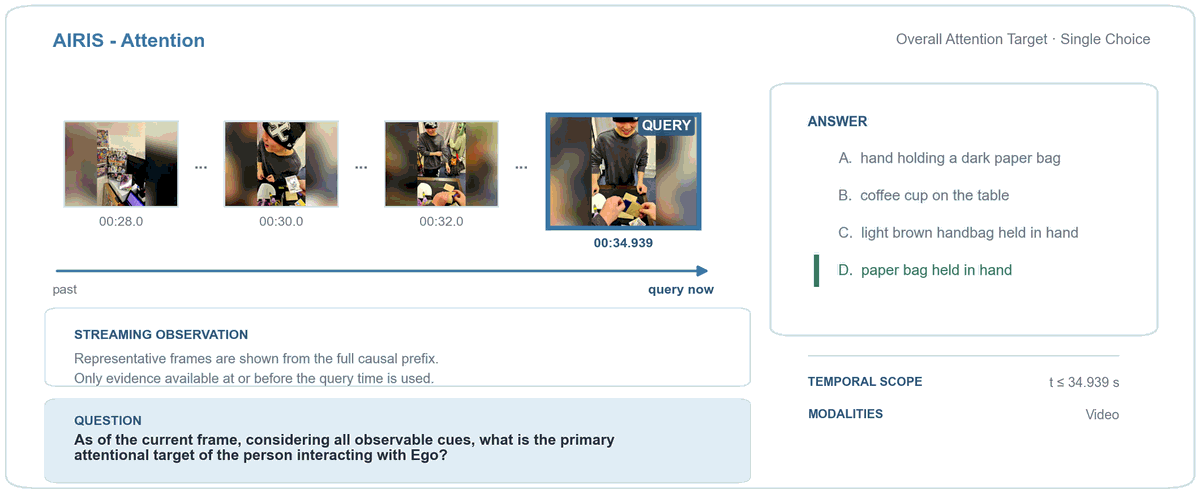}{overall attention-target recognition}{fig:attn-overall-target}
\airisattentioncase{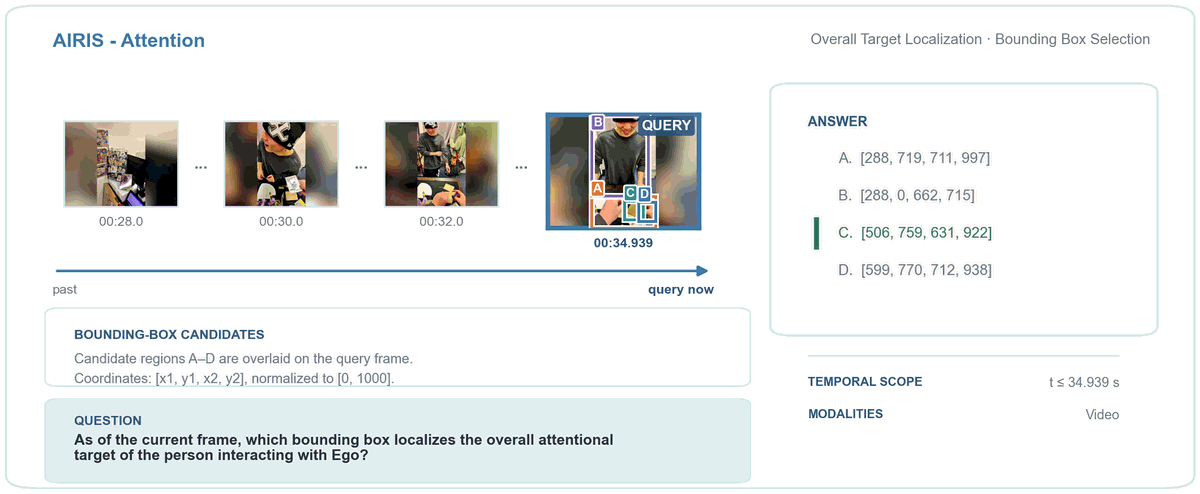}{overall attention-target localization}{fig:attn-overall-target-bbox}
\airisattentioncase{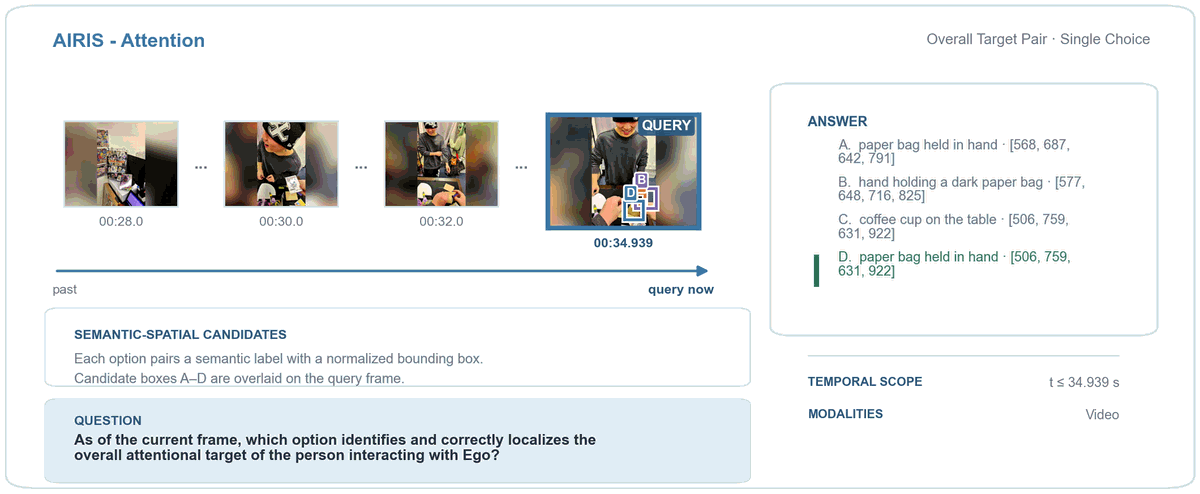}{joint overall attention-target identification and localization}{fig:attn-overall-target-pair}

\subsubsection{Intention cases}
\label{app:intention-cases}

\newcommand{\airisintentioncase}[3]{%
  \airisqueuecase{#1}{#2}{#3}{Intention}}

Figures~\ref{fig:intent-kf-social-conversation}-\ref{fig:intent-evidence-other}
provide two representative examples for each of the three Intention
capabilities: key-moment intention recognition, temporal intention recognition,
and intention evidence grounding.  The paired cases cover distinct social
contexts and evidence conditions, including both supportable intentions and an
early prefix for which the calibrated response is insufficient evidence.  Each
panel shows only the causal video prefix and the ASR available by the marked
query time, together with the original answer choices and reviewed gold
response.

\airisintentioncase{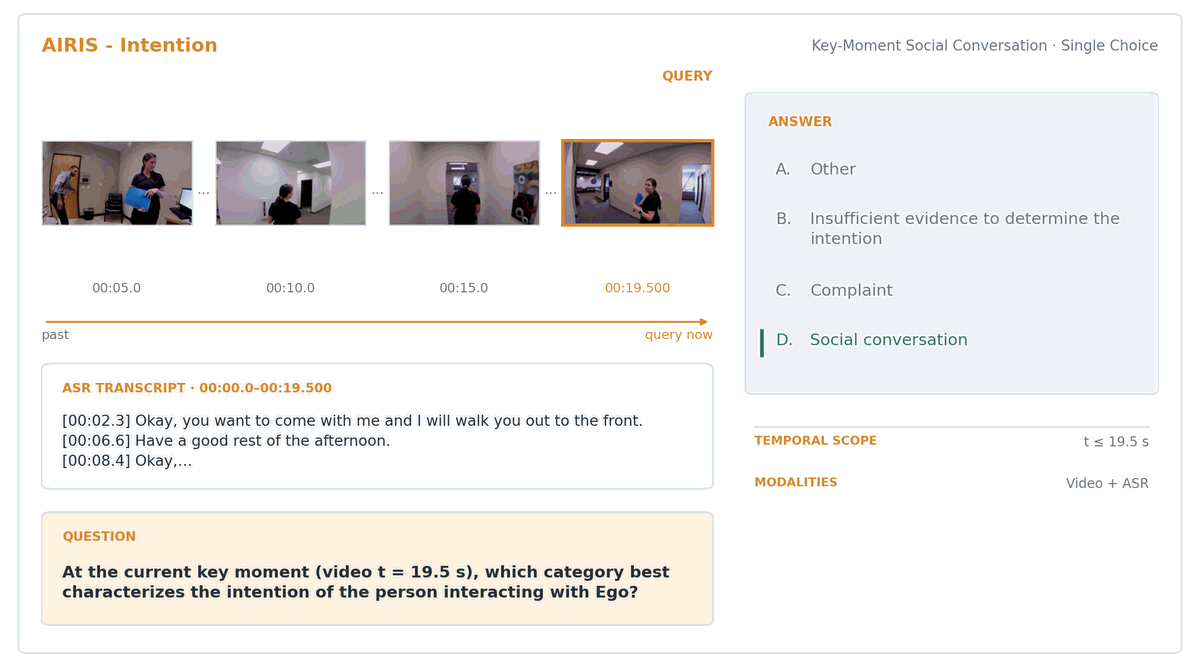}{key-moment social-conversation recognition}{fig:intent-kf-social-conversation}
\airisintentioncase{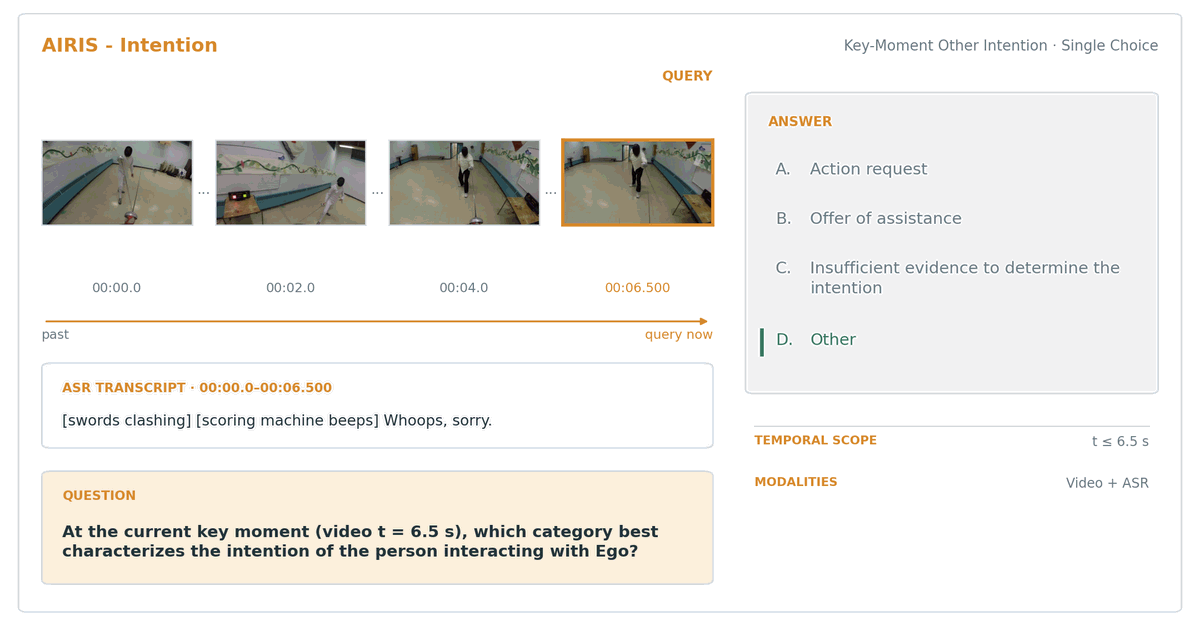}{key-moment other-intention recognition}{fig:intent-kf-other}
\airisintentioncase{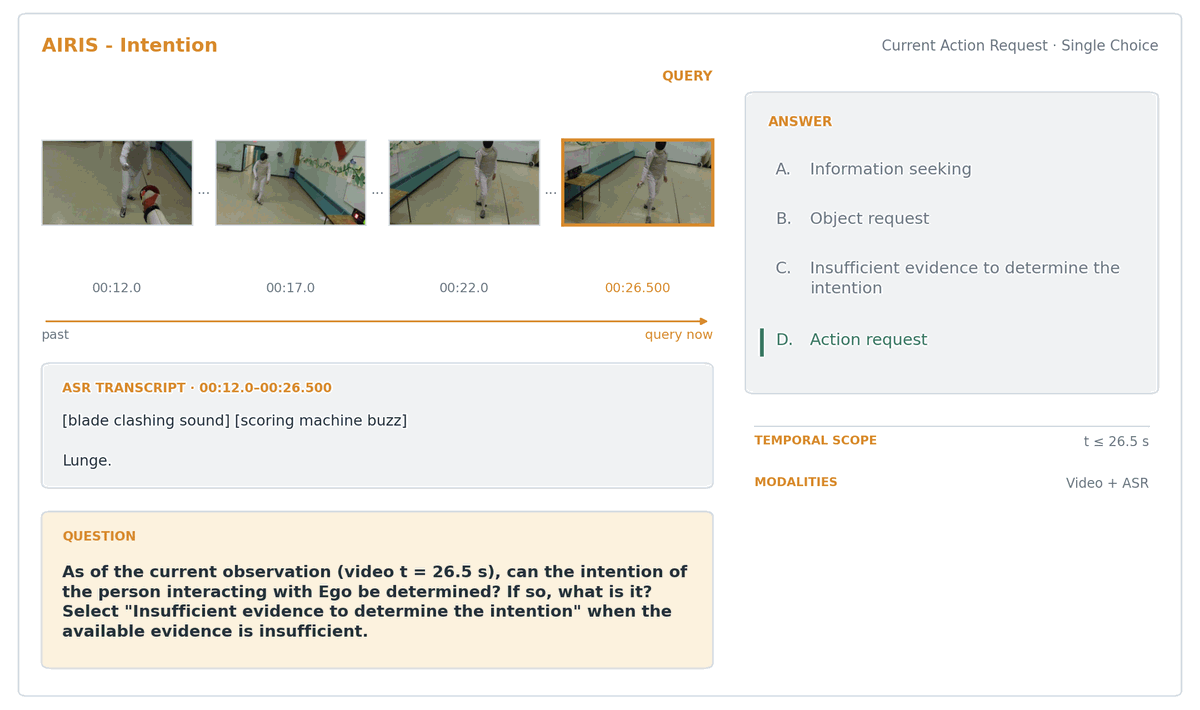}{temporal action-request recognition}{fig:intent-at-action-request}
\airisintentioncase{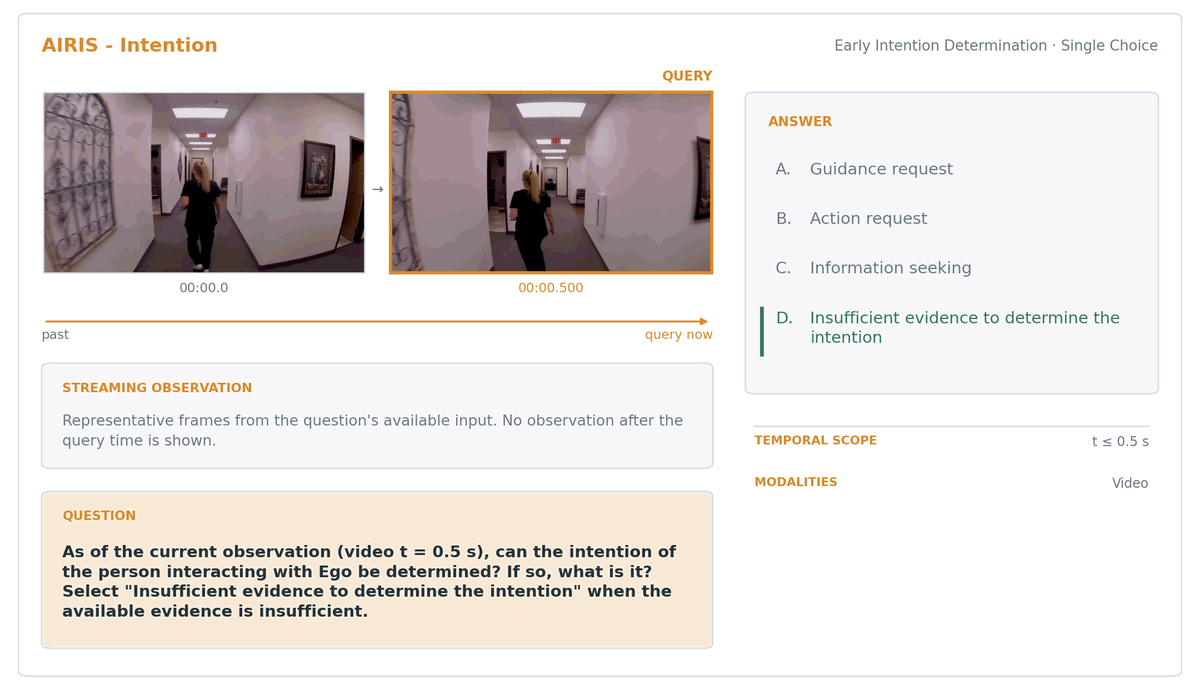}{early intention determination under insufficient evidence}{fig:intent-at-insufficient-evidence}
\airisintentioncase{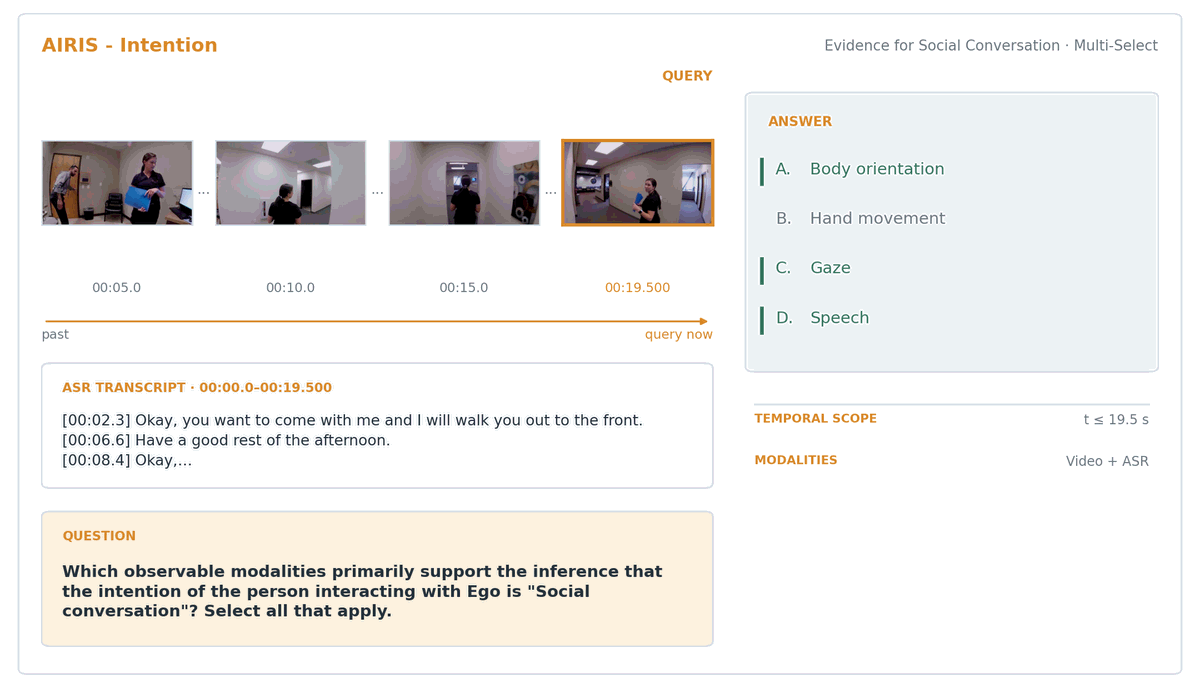}{social-conversation evidence grounding}{fig:intent-evidence-social-conversation}
\airisintentioncase{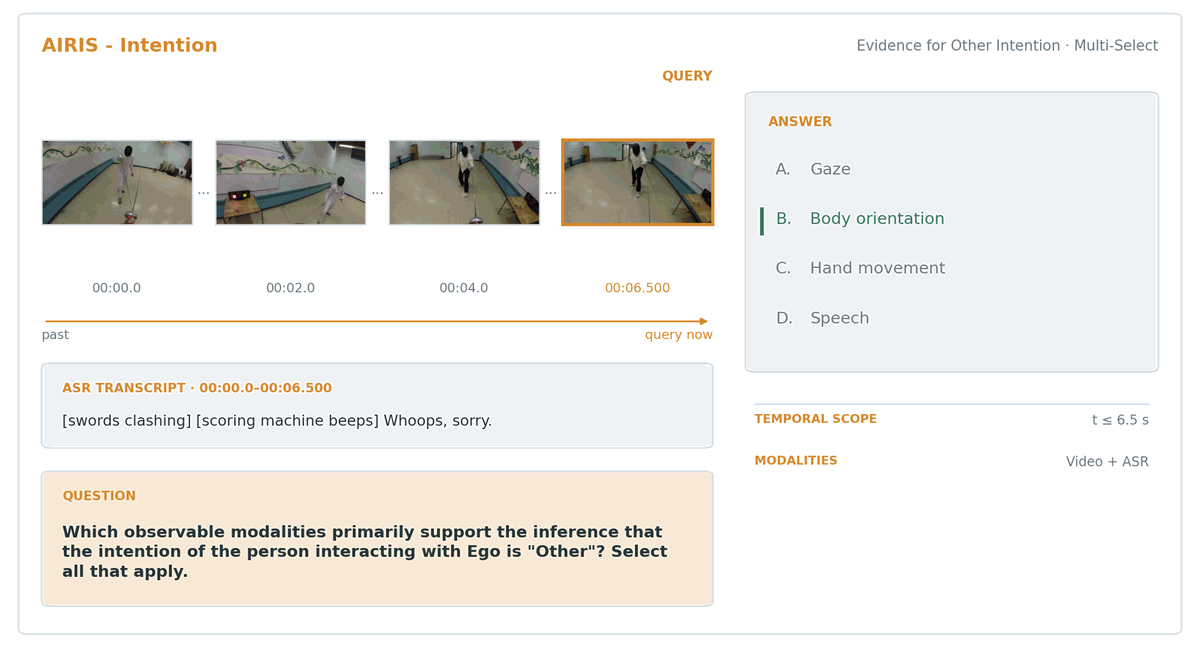}{other-intention evidence grounding}{fig:intent-evidence-other}

\subsubsection{Reaction cases}
\label{app:reaction-cases}

\newcommand{\airisreactioncase}[3]{%
  \airisqueuecase{#1}{#2}{#3}{Reaction}}

Figures~\ref{fig:reaction-engagement-checkout}-\ref{fig:reaction-speech-service-window}
provide representative examples spanning the four Reaction
capabilities: engagement decision, action-domain selection, subaction
selection, and speech-function selection.  The paired checkout and service-window
scenes illustrate how AIRIS progresses from deciding whether Ego should respond
to specifying the appropriate response modality, concrete action, and, when
speech is warranted, communicative function.  Each panel uses only the causal
video prefix and any ASR available by the marked query time, and retains the
original answer choices and reviewed gold response.

\airisreactioncase{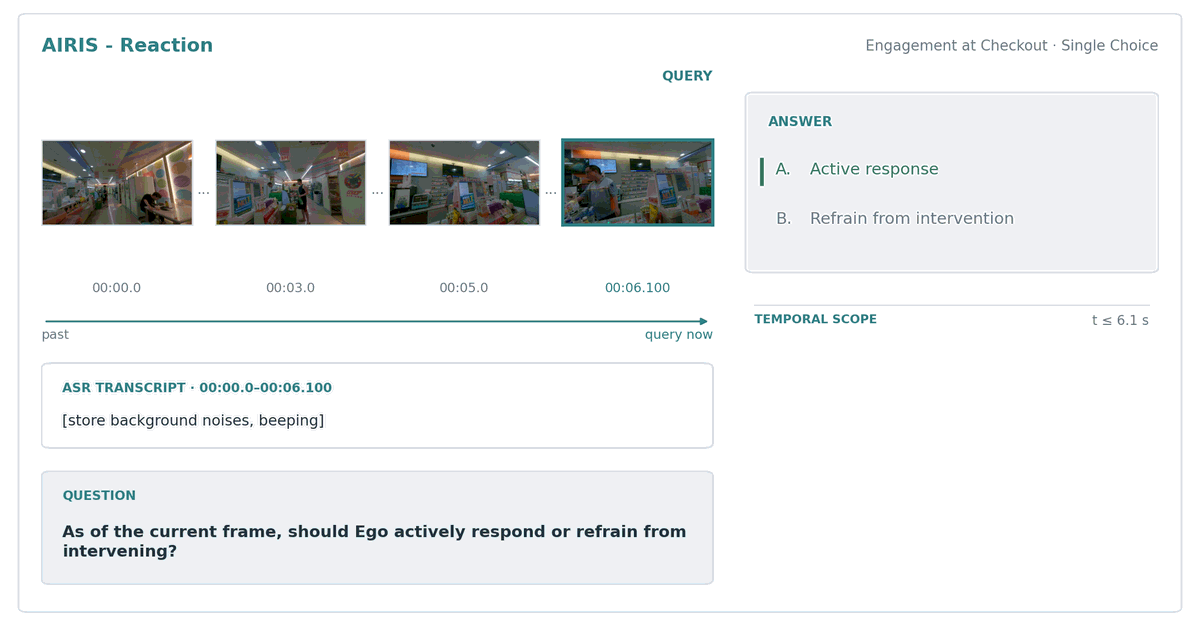}{engagement decision at checkout}{fig:reaction-engagement-checkout}
\airisreactioncase{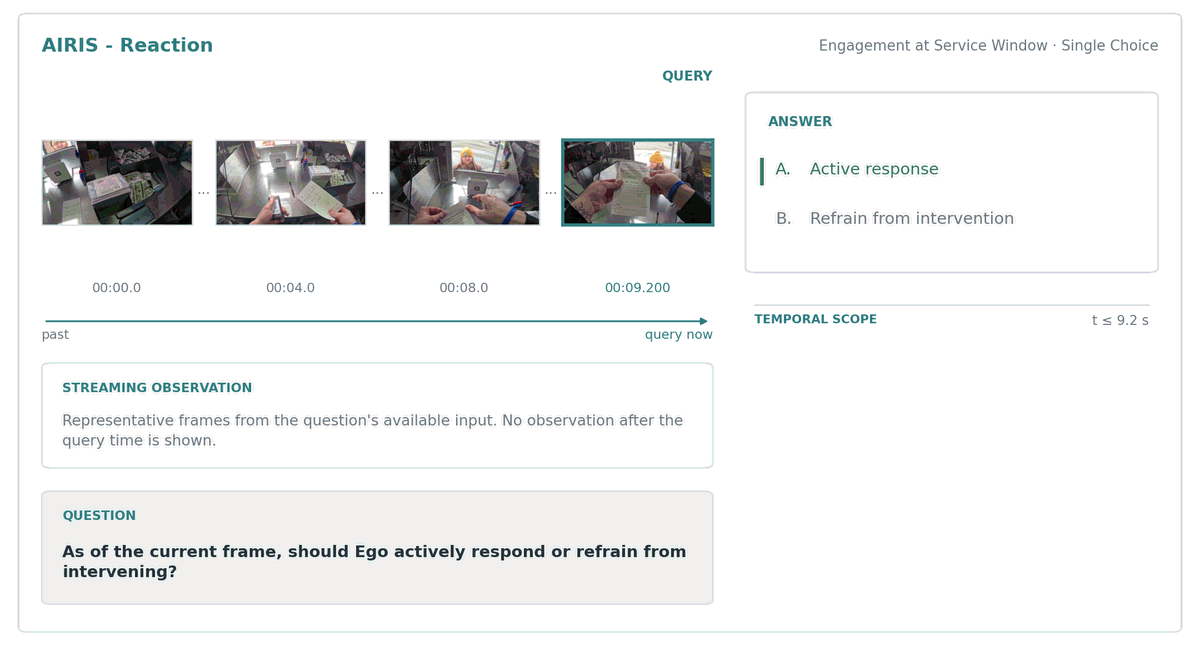}{engagement decision at a service window}{fig:reaction-engagement-service-window}
\airisreactioncase{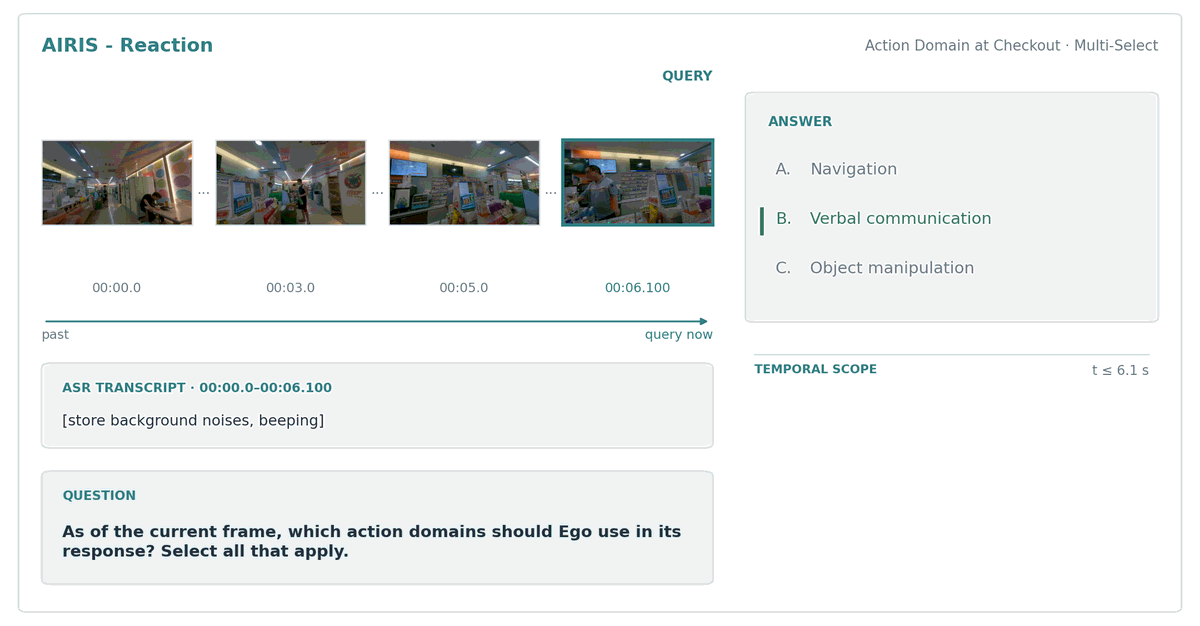}{action-domain selection at checkout}{fig:reaction-domain-checkout}
\airisreactioncase{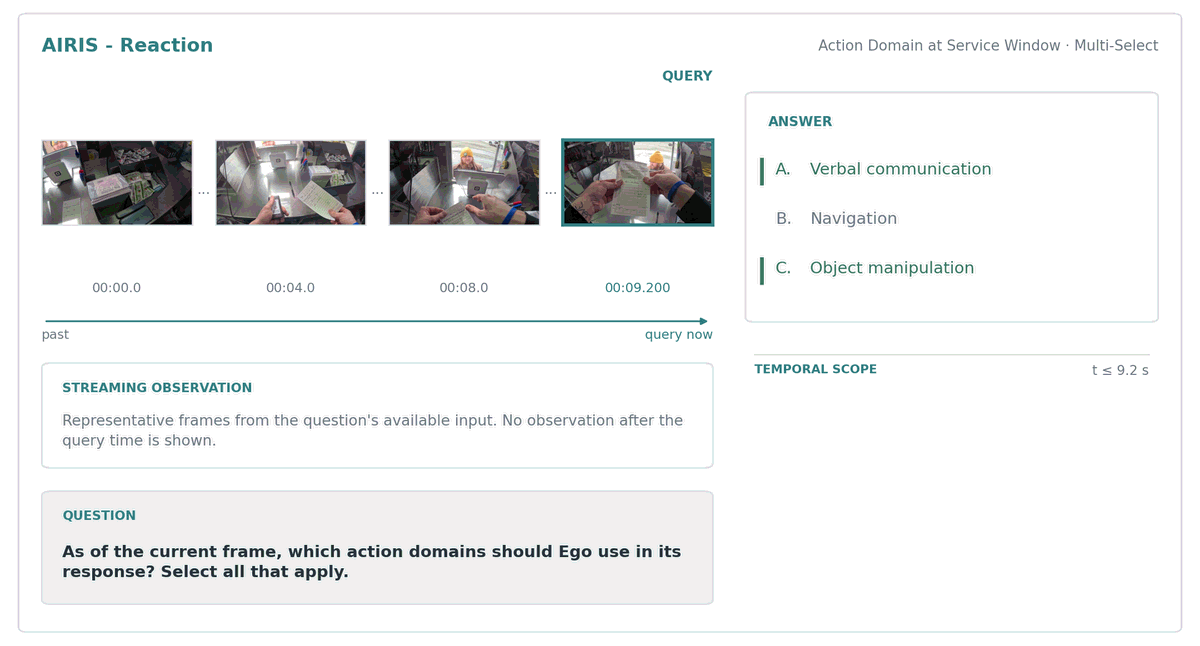}{action-domain selection at a service window}{fig:reaction-domain-service-window}
\airisreactioncase{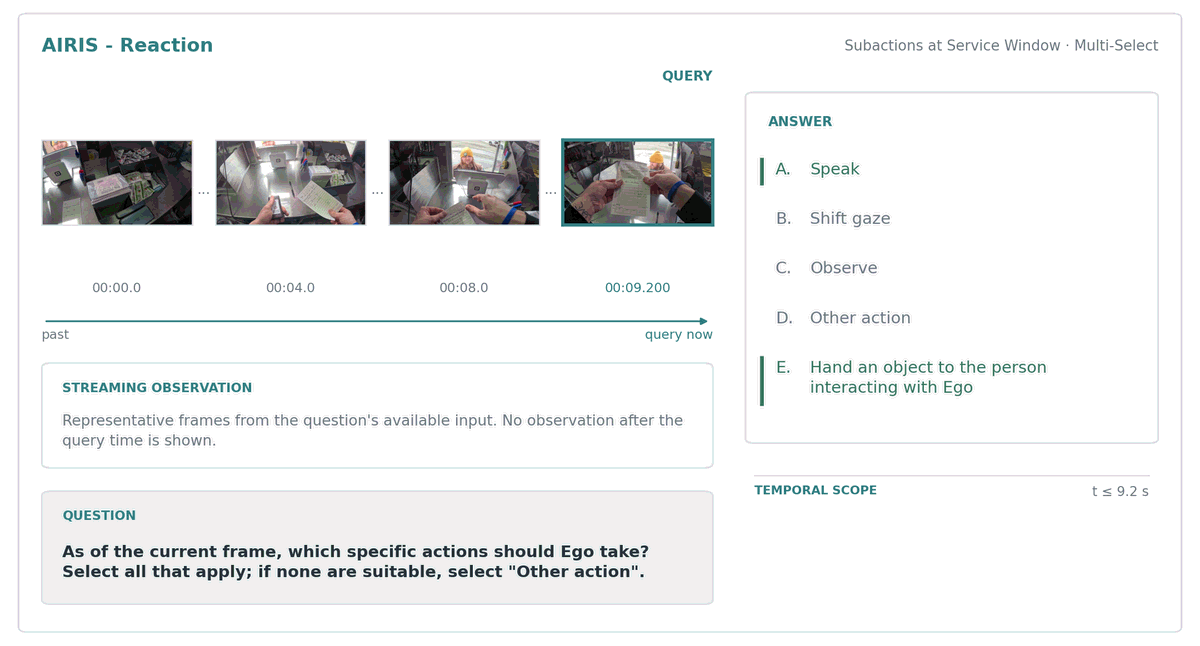}{subaction selection at a service window}{fig:reaction-subaction-service-window}
\airisreactioncase{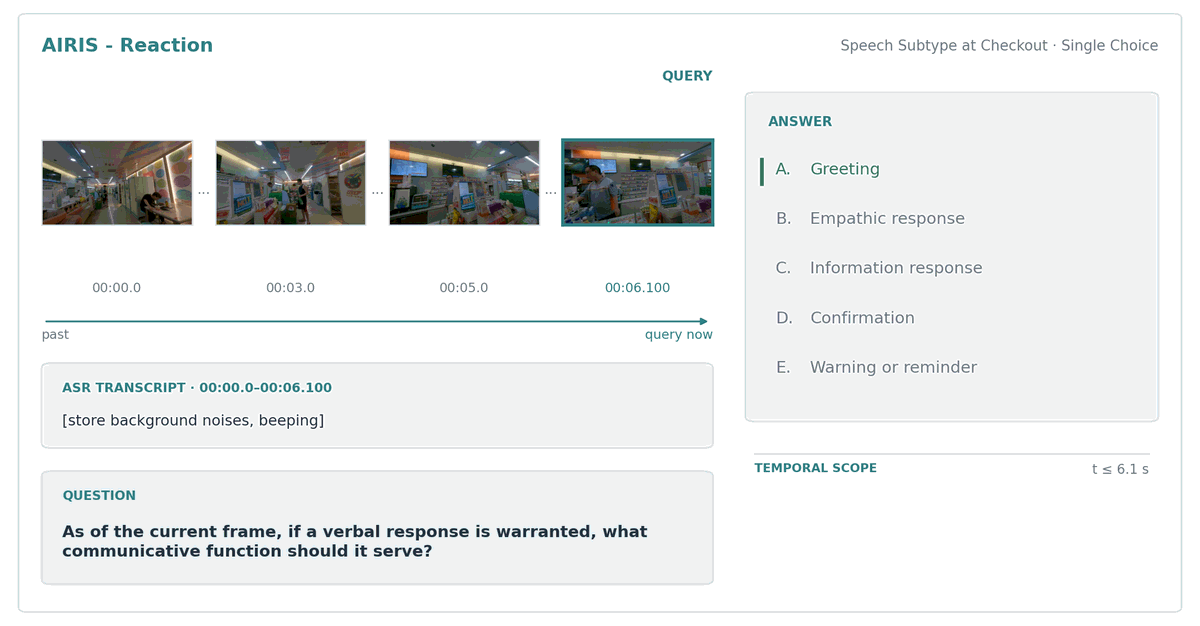}{speech-function selection at checkout}{fig:reaction-speech-checkout}
\airisreactioncase{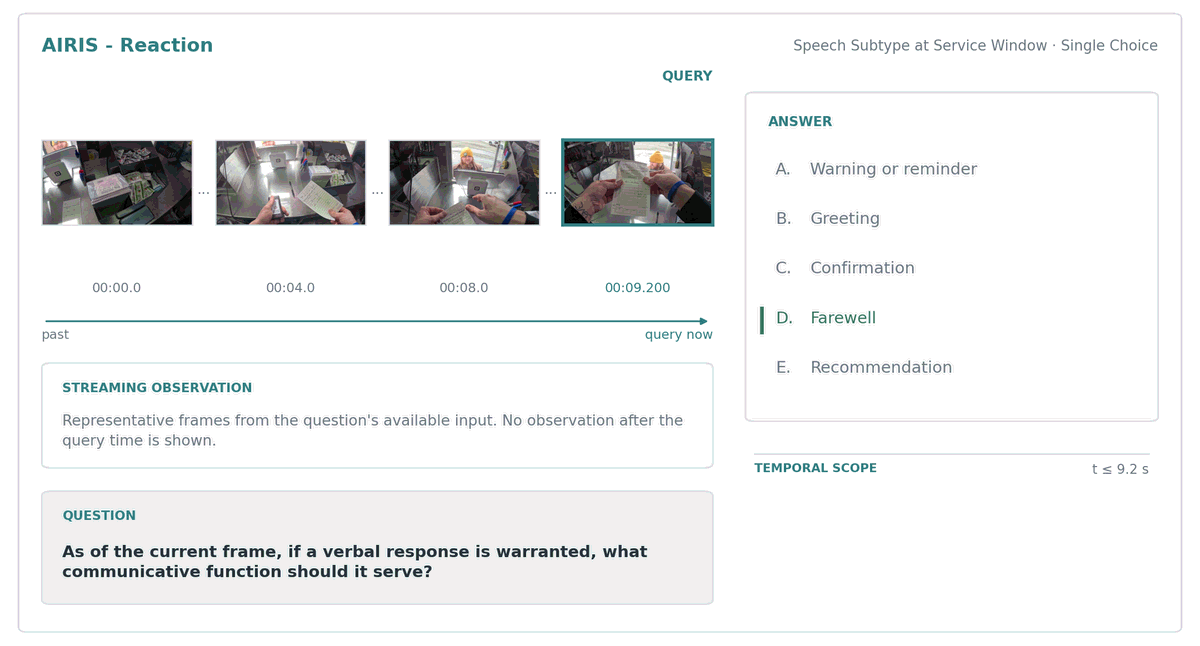}{speech-function selection at a service window}{fig:reaction-speech-service-window}

\subsubsection{Interaction cases}
\label{app:interaction-cases}

\newcommand{\airisinteractioncase}[3]{%
  \airisqueuecase{#1}{#2}{#3}{Interaction}}

Figures~\ref{fig:interaction-state-sales-floor}-\ref{fig:interaction-violations-segment-exam}
provide one representative example for each of the ten Interaction capabilities.
The query-time cases assess the evolving interaction state and whether a social
violation is present at the current prefix, whereas the segment-level cases
assess completion, mutual understanding, spatial, physical, and communication
compliance, overall compliance category, and the violation types observed over
the episode.  Together, the examples cover both compliant and violating
interactions across retail-counter, reception, sales-floor, and simulated skills exam
settings.  Each panel retains the original answer choices and reviewed gold
response, and explicitly distinguishes a causal query time from post-hoc review
of the supplied interaction segment.

\airisinteractioncase{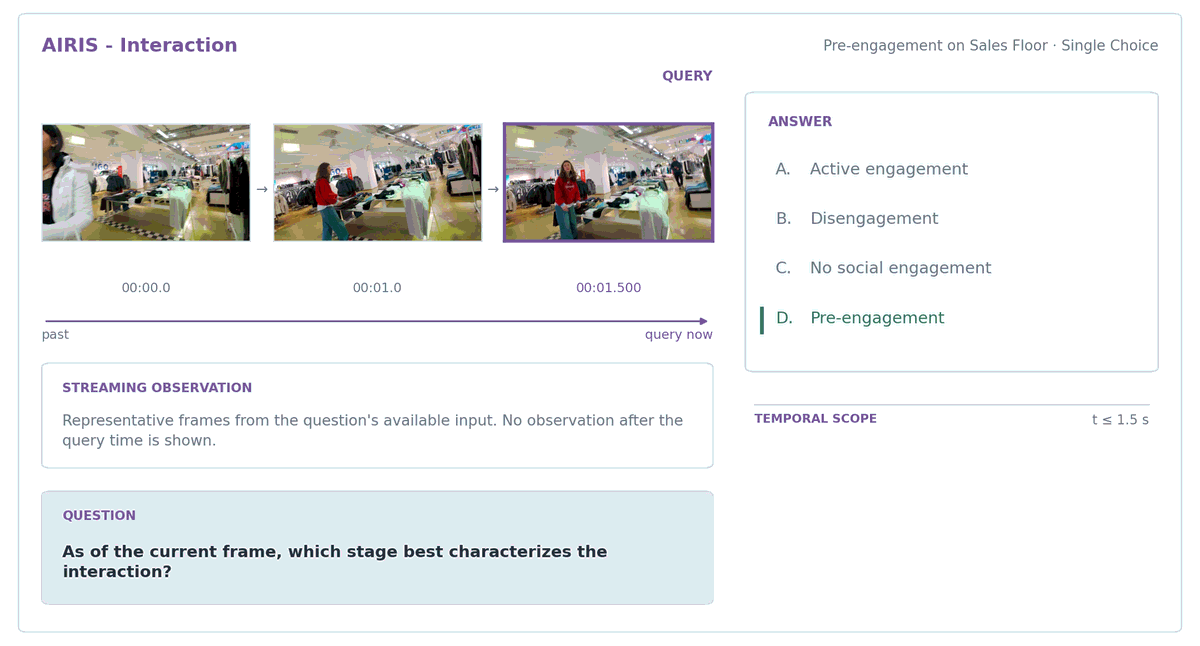}{pre-engagement-state recognition on a retail sales floor}{fig:interaction-state-sales-floor}
\airisinteractioncase{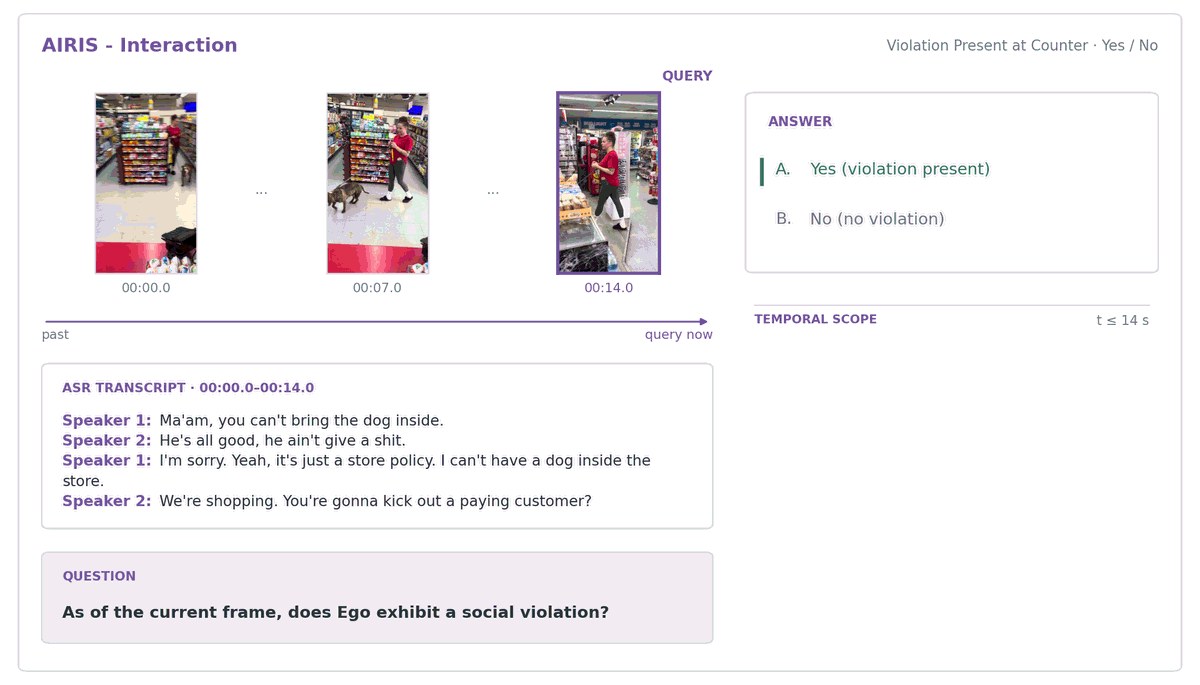}{query-time violation detection at a store counter}{fig:interaction-violation-presence-counter}
\airisinteractioncase{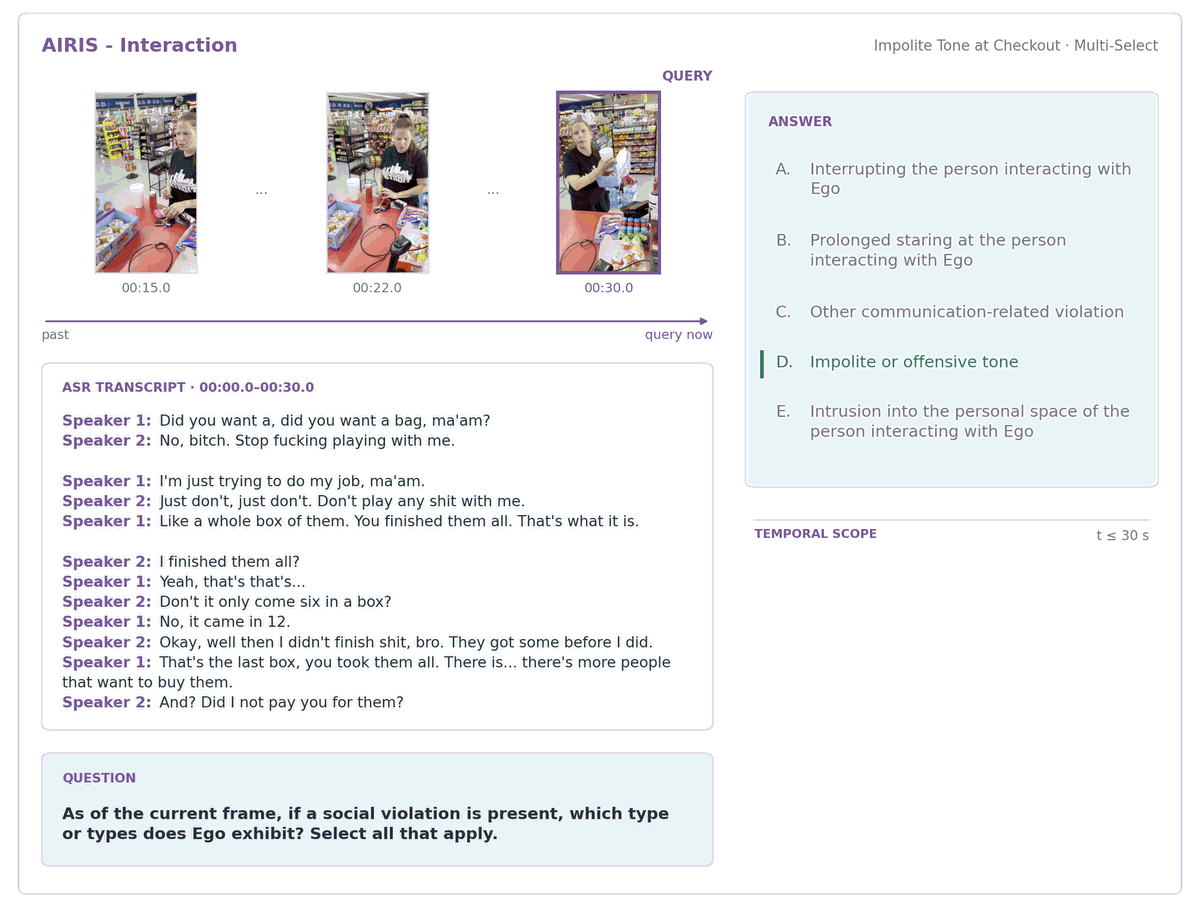}{query-time communication-violation typing at checkout}{fig:interaction-violation-type-checkout}
\airisinteractioncase{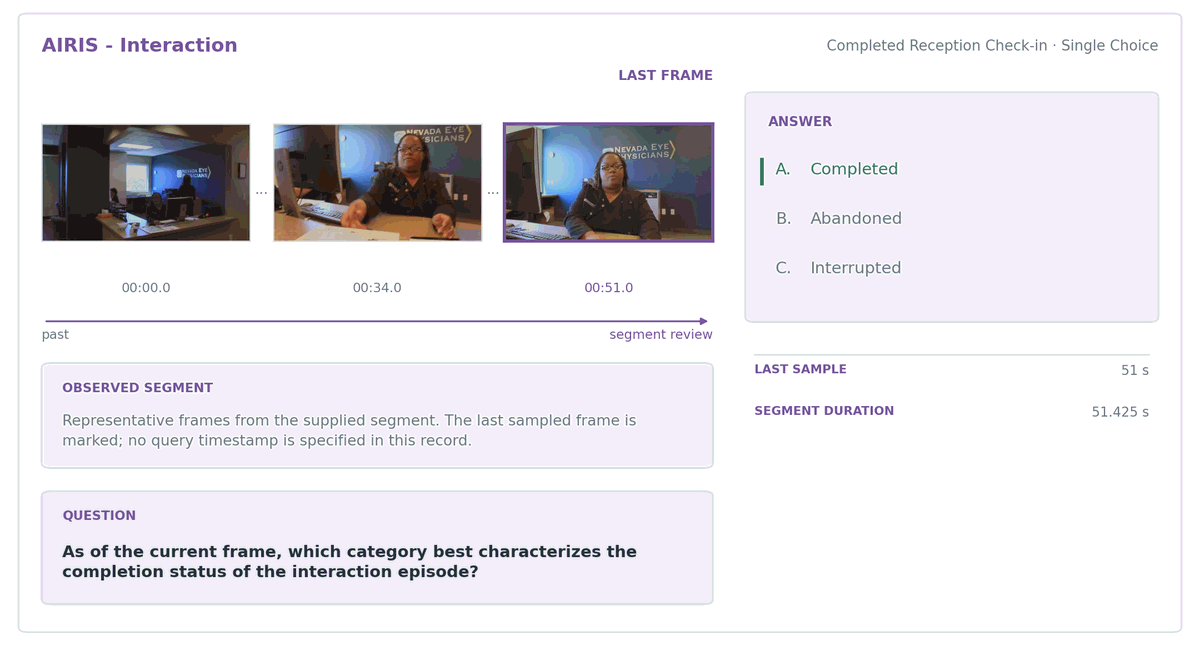}{episode-completion assessment at reception}{fig:interaction-completeness-reception}
\airisinteractioncase{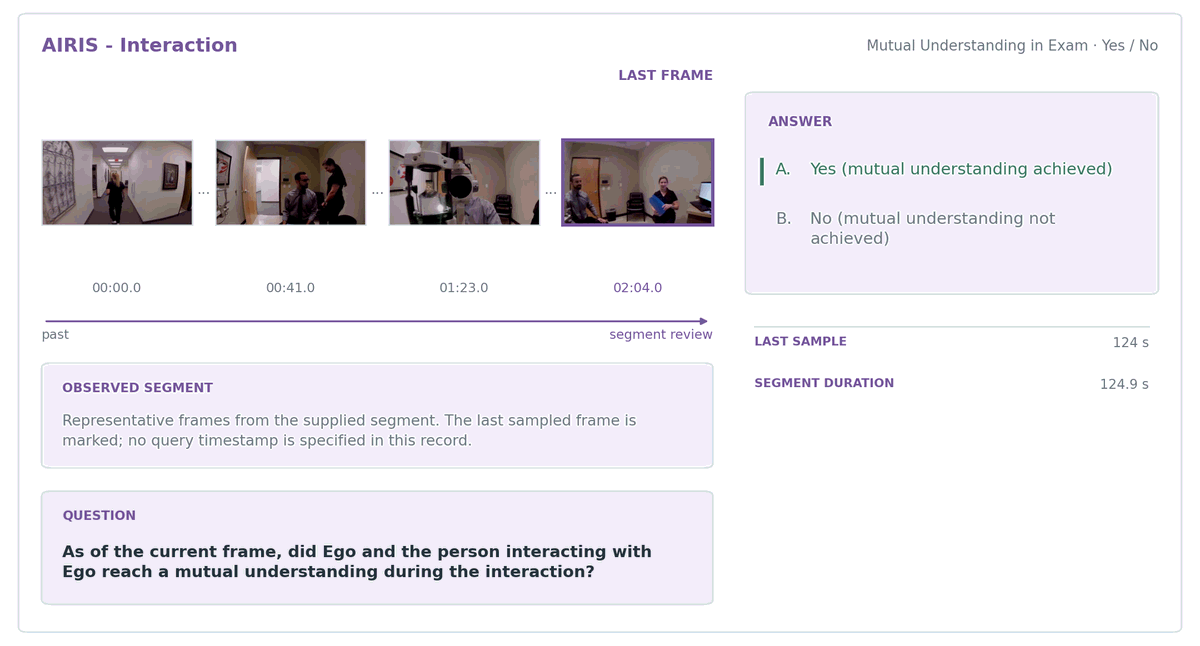}{mutual-understanding assessment in a simulated skills exam}{fig:interaction-mutual-exam}
\airisinteractioncase{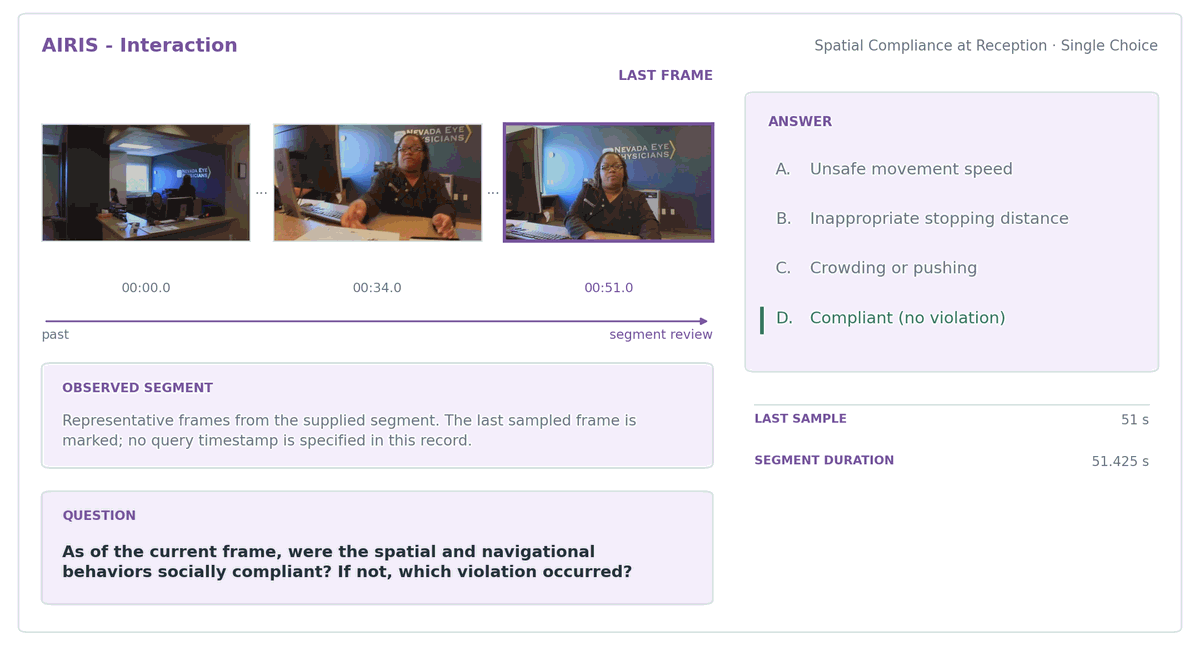}{spatial-compliance assessment at reception}{fig:interaction-spatial-reception}
\airisinteractioncase{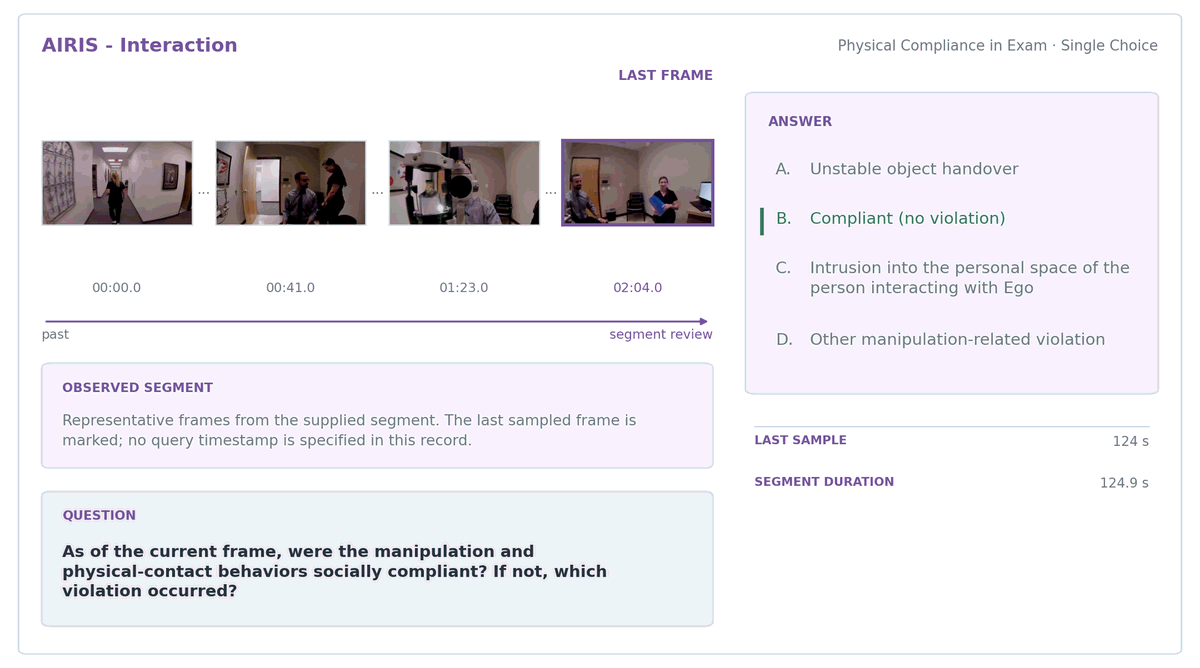}{physical-compliance assessment in a simulated skills exam}{fig:interaction-physical-exam}
\airisinteractioncase{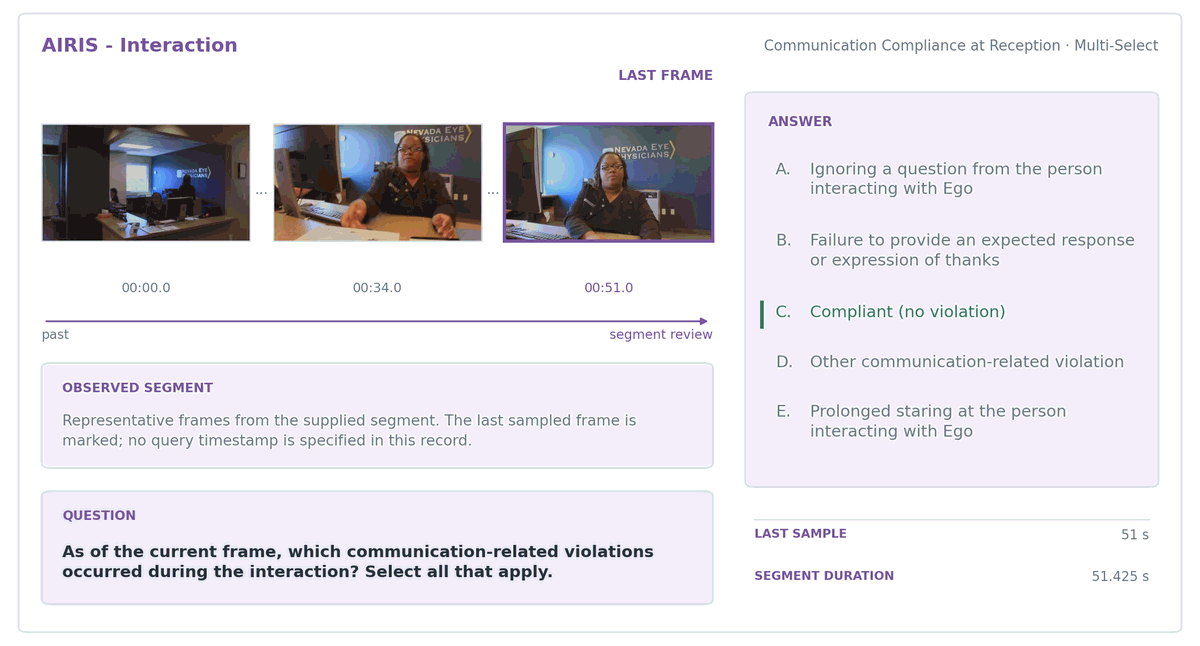}{communication-compliance assessment at reception}{fig:interaction-communication-reception}
\airisinteractioncase{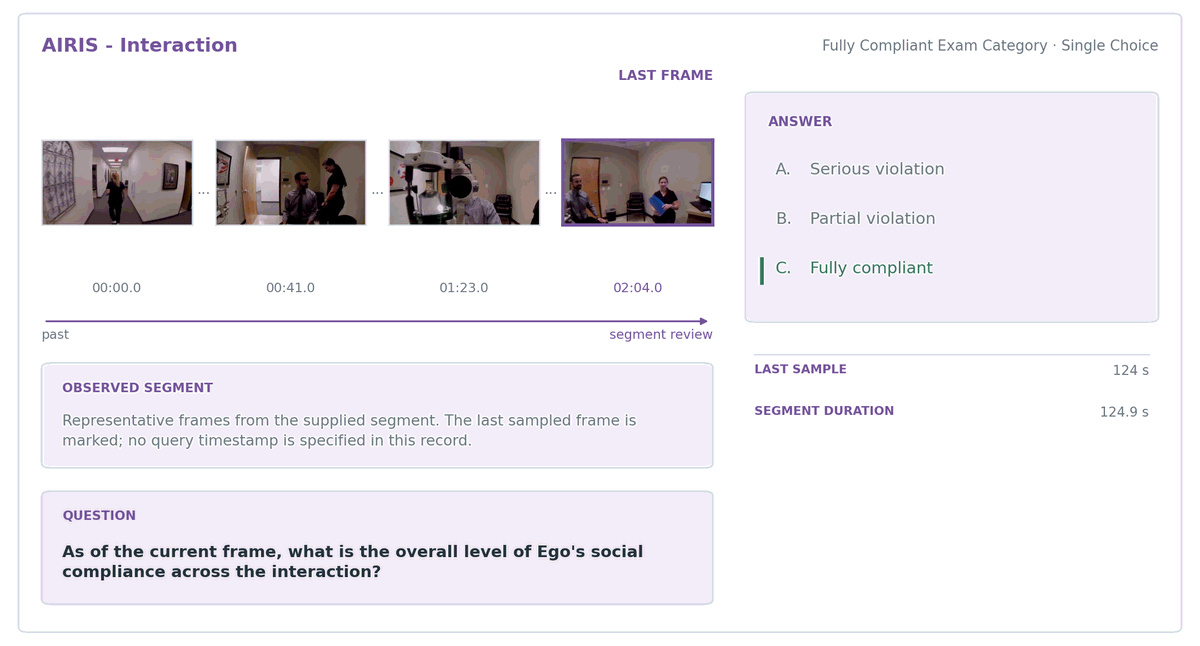}{overall social-compliance categorization in a simulated skills exam}{fig:interaction-category-exam}
\airisinteractioncase{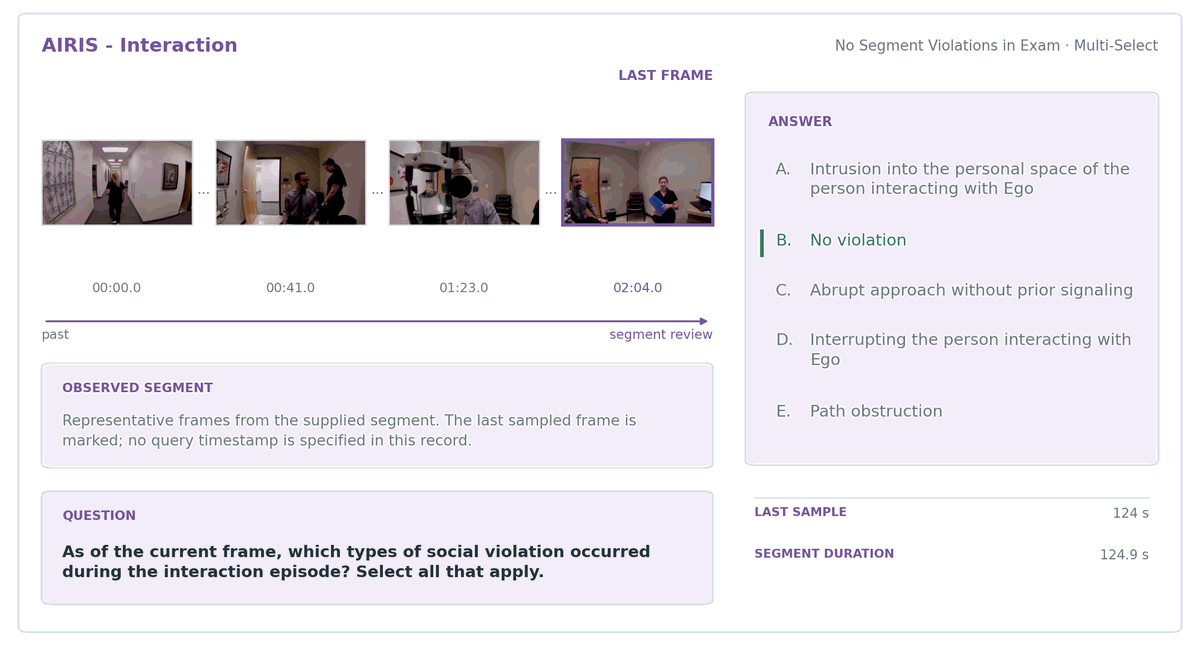}{segment-level violation-type assessment in a simulated skills exam}{fig:interaction-violations-segment-exam}

\clearpage
\begingroup
\setlength{\parskip}{0pt}
\setlength{\parindent}{0pt}
\airiscasegallery
\par
\endgroup

\end{document}